\documentclass[10pt,twocolumn,letterpaper]{article}

\usepackage{iccv}              %
\usepackage[accsupp]{axessibility}  %

\usepackage[accsupp]{axessibility}  %

\usepackage{multirow}
\usepackage{arydshln}
\usepackage{amssymb}
\usepackage{pifont}
\usepackage{nicematrix}

\usepackage{lipsum}

\newcommand{\perf}[1]{%
    {\scriptsize \textcolor{%
        \ifdim #1pt<0pt red\else  ForestGreen\fi%
    }{(#1)}}%
}

\definecolor{mygray}{gray}{.95}
\newcommand{\revperf}[1]{%
    {\scriptsize \textcolor{%
        \ifdim #1pt>0pt red\else  ForestGreen\fi%
    }{(#1)}}%
}

\definecolor{softred}{RGB}{200, 50, 50}
\definecolor{softgreen}{RGB}{50, 150, 50}
\definecolor{softorange}{RGB}{200, 120, 50}
\definecolor{softblue}{RGB}{90, 140, 200}

\usepackage{booktabs}
\usepackage{algorithm,algpseudocode}
\usepackage{colortbl}

\definecolor{iccvblue}{rgb}{0.21,0.49,0.74}
\usepackage[pagebackref,breaklinks,colorlinks,allcolors=iccvblue]{hyperref}

\def\paperID{6074} %
\def\confName{ICCV}
\def\confYear{2025}

\title{Task Vector Quantization for Memory-Efficient Model Merging }

\author{
Youngeun Kim\textsuperscript{1}\thanks{Equal contribution.} \quad
Seunghwan Lee\textsuperscript{2}\footnotemark[1] \quad
Aecheon Jung\textsuperscript{2}\footnotemark[1] \quad
Bogon Ryu\textsuperscript{2} \quad
Sungeun Hong\textsuperscript{2}\thanks{Corresponding author.} \\
\textsuperscript{1}Yale University \quad
\textsuperscript{2}Sungkyunkwan University
\\
{\tt\small youngeun.kim@yale.edu \qquad \{simon2, 
kasurashan, bogon.ryu, csehong\}@skku.edu}}

\begin{document}
\maketitle
\begin{abstract}

Model merging enables efficient multi-task models by combining task-specific fine-tuned checkpoints. However, storing multiple task-specific checkpoints requires significant memory, limiting scalability and restricting model merging to larger models and diverse tasks. In this paper, we propose quantizing task vectors (i.e., the difference between pre-trained and fine-tuned checkpoints) instead of quantizing fine-tuned checkpoints. We observe that task vectors exhibit a narrow weight range, enabling low-precision quantization ($\le 4$ bit) within existing task vector merging frameworks. To further mitigate quantization errors within ultra-low bit precision (e.g., $2$ bit), we introduce Residual Task Vector Quantization, which decomposes the task vector into a base vector and offset component. We allocate bits based on quantization sensitivity, ensuring precision while minimizing error within a memory budget. Experiments on image classification and dense prediction show our method maintains or improves model merging performance while using only 8\% of the memory required for full-precision checkpoints. Our code is available at \url{https://aim-skku.github.io/TVQ/}.

\end{abstract}
    
\section{Introduction}
\label{sec:intro}

{Model merging aims to combine multiple well-trained models into a single set of parameters, preserving their capabilities while using less memory than ensembles, which require multiple models for inference.
Prior studies~\cite{WiseFT_2022_CVPR, ModelSoups_2022_ICML, ModelStock_2024_ECCV} have shown that interpolating weights from single-task models improves generalization, even in out-of-distribution settings.
Building on these, recent work has explored more challenging scenarios, such as multi-task learning~\cite{MTL_1997,MTL_2021_TPAMI}, by unifying single-task models trained for distinct tasks into a single model.
}

\begin{figure}[t]
\begin{center}
\centering
\def\arraystretch{0.5}
\begin{tabular}{@{\hskip 0.01\linewidth}c@{\hskip 0.03\linewidth}c@{}c}
\includegraphics[width=1.0\linewidth]{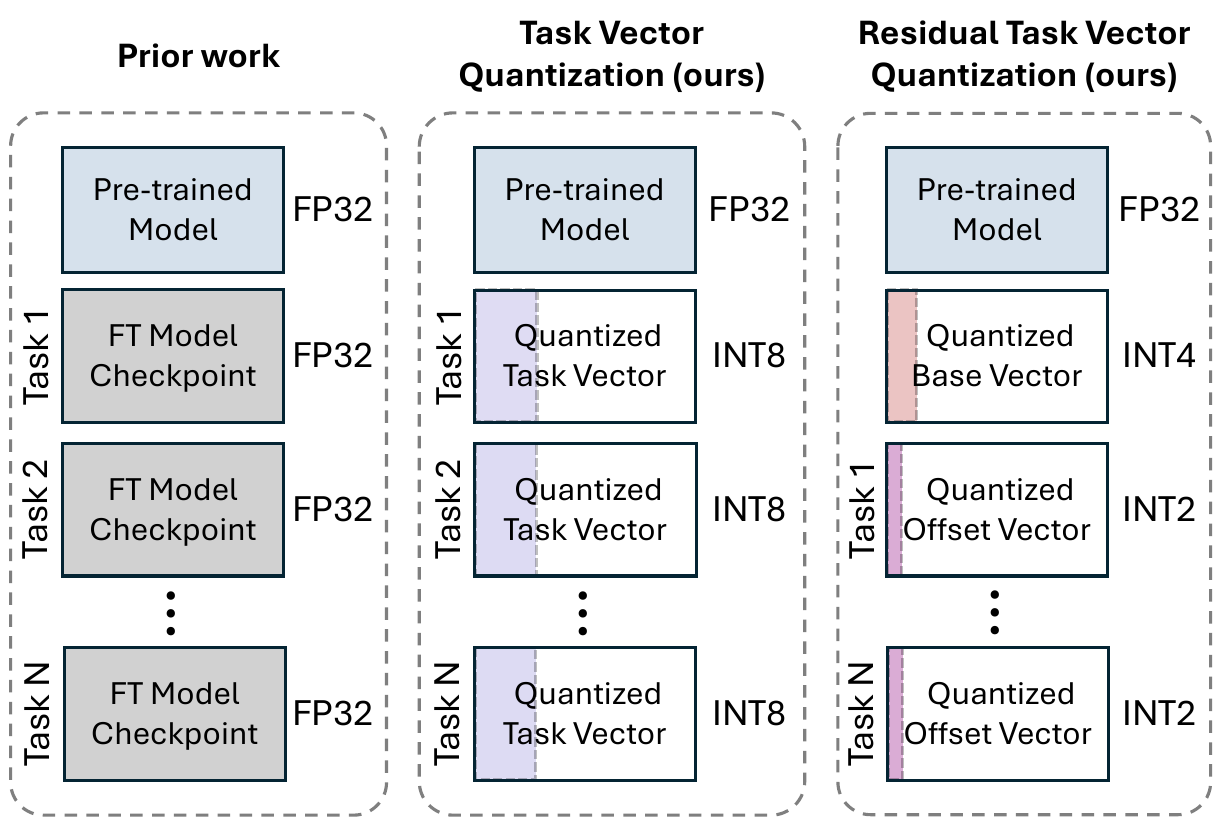} 
\end{tabular}
\end{center}
\vspace{-5mm}
\caption{ 
Prior work stores multiple fine-tuned checkpoints in full precision, causing high storage overhead. We introduce \textbf{Task Vector Quantization} to reduce storage by quantizing task vectors. Furthermore, we propose \textbf{Residual Task Vector Quantization}, which decomposes task vectors into a base and offset, both quantized at ultra-low precision to minimize error. Partially shaded boxes indicate reduced bit storage.}
\label{fig:intro:concept}
\end{figure}

{Task Arithmetic~\cite{TaskArithmetic_2023_ICLR}  has emerged as a promising method for model merging by representing each task as the weight difference between pre-trained and fine-tuned models. This approach brings the advantage of efficiently capturing task-specific adaptations without retraining from scratch.}
{In particular, it provides a flexible and scalable framework for creating multi-task merged models: simply leveraging stored fine-tuned model checkpoints (\ie, the trained model parameters) allows seamless integration of newly added tasks. Based on this concept, following studies~\cite{EMRMerging_2024,TwinMerging_2024,Surgery_2024_ICML} have achieved performance comparable to joint multi-task training.}
However, storing these checkpoints requires significant memory overhead. For instance, a ViT-L/14 model needs 1.14 GB per fine-tuned checkpoint, totaling 22.8 GB for 20 tasks. In resource-constrained environments like an NVIDIA Jetson Nano \cite{kurniawan2021introduction} with 16 GB storage, such high memory demands for checkpoints hinder scaling to larger models and more tasks.

In this paper, we introduce Task Vector Quantization (TVQ), a quantization method that enhances memory efficiency when storing multiple checkpoints as illustrated in Fig.~\ref{fig:intro:concept}. Our approach is based on the observation that task vectors (\ie, the difference between fine-tuned and pre-trained checkpoints) exhibit a weight range that is an order of magnitude narrower than that of fine-tuned weights (refer to Section \ref{sec:method_observation}). Since the upper bound of quantization is determined by the range of weight value, quantizing task vectors results in a smaller quantization error compared to quantizing full fine-tuned checkpoints. Based on this, we propose to quantize task vectors  rather than the entire set of fine-tuned weights. This allows for effective low bit precision quantization (\eg, $4$ bit) while maintaining performance within existing task vector merging frameworks.

\begin{figure}[t]
\begin{center}
\centering
\def\arraystretch{0.5}
\begin{tabular}{@{\hskip 0.01\linewidth}c@{\hskip 0.03\linewidth}c@{}c}
\includegraphics[width=1.0\linewidth]{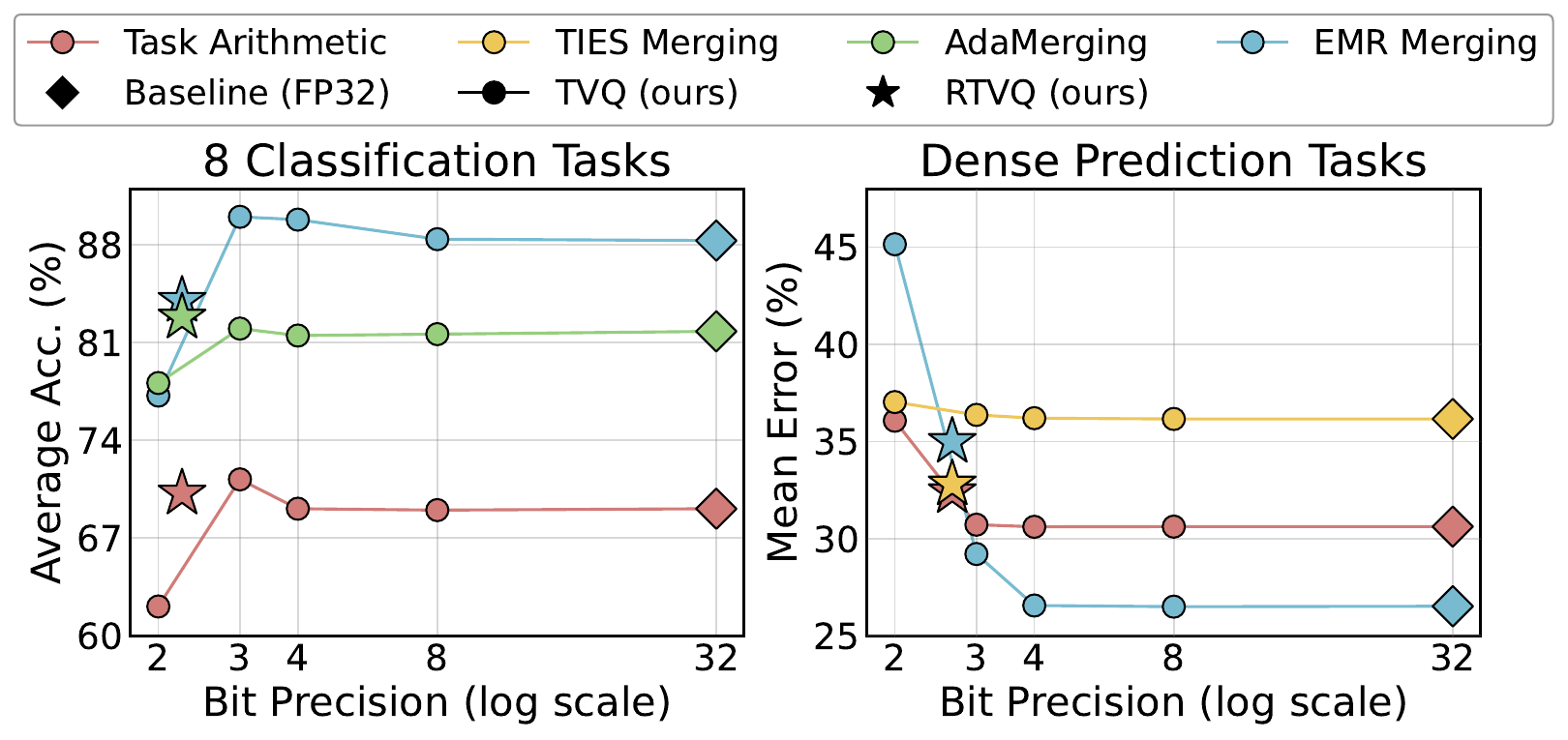} 
\end{tabular}
\end{center}
\vspace{-5mm}
\caption{ 
Effect of our quantization methods (TVQ and RTVQ) on model merging. (Left) merging 8 classification tasks; (Right) merging dense prediction tasks, evaluated on normal estimation. TVQ remains competitive at 3 bit but sharply declines at 2 bit, while RTVQ retains greater robustness under low-bit conditions.}
 \vspace{-2mm}
\label{fig:intro:performance}
\end{figure}

Although the proposed TVQ method remains robust up to 4-bit precision, our experiments show a sharp performance drop at 2-bit, which may limit deployment in highly memory-constrained settings. To further reduce quantization error in low-precision settings, we propose Residual Task Vector Quantization (RTVQ).
This approach represents multiple task vectors with a shared base vector and several small offsets. By assigning higher bit precision to the base vector, it reduces quantization error while keeping memory overhead minimal, as only one base vector is stored across all tasks. We found that even with a 3-bit base vector and 2-bit offset vectors, performance is maintained for most model merging methods.

Our comprehensive experiments demonstrate that our method maintains or even improves performance with state-of-the-art merging techniques for image classification and dense prediction (see Fig.~\ref{fig:intro:performance}). More importantly, our quantization method modifies only the checkpoints, meaning that it can be integrated into existing model merging frameworks without any modifications. Using just 8\% of the original storage, our quantized checkpoints remain effective for both vision tasks.
Especially, dense prediction tasks~\cite{choi2023intra,zhang2025memory}\footnote{These include segmentation, depth estimation, and normal estimation.} are crucial for vision applications requiring spatially precise reasoning but has been largely overlooked in prior model merging studies. To address this, we first reproduce existing merging approaches in dense prediction settings and evaluate our quantization method, demonstrating its strong performance on these complex tasks.

The main contributions of our work are as follows:
\begin{itemize}
\item We introduce a novel quantization method for task vectors, leveraging their significantly narrower weight range to achieve effective low-precision quantization.

\item To further reduce quantization error, we propose Residual Task Vector Quantization that decomposes task vectors into base and offset components. 

\item Our method modifies only checkpoints, ensuring seamless integration into existing model merging frameworks. Extensive experiments across multiple domains, including dense prediction and {natural language processing}, validate its effectiveness.

\end{itemize}

\section{Related Work}
\label{sec:related_work}

\subsection{Model Merging}
\label{sec:related_work}

Early model merging studies~\cite{ModelSoups_2022_ICML, ModelStock_2024_ECCV, WiseFT_2022_CVPR} explore combining models trained on the same task in weight space, demonstrating robustness to distribution shifts and improved performance on unseen data.
Recent studies have extended this paradigm to multi-task learning~\cite{MTL_1997,MTL_2021_TPAMI,zhang2022survey_MTL}, where merging multiple models trained on different tasks enables efficient knowledge sharing across domains.

Task Arithmetic~\cite{TaskArithmetic_2023_ICLR} construct multi-task models through additive task vectors, which are computed by subtracting the pre-trained model's parameter from those of the fine-tuned model. Unlike direct interpolation, task vectors enable modular composition of specialized models, making them more flexible for multi-task settings.
Sparsification-based merging~\cite{TiesMerging_2023_NeurIPS,Dare_2024_ICML}, where only the most significant parameters are retained to reduce interference, often complemented by rescaling strategies to enhance stability before integration.
However, these methods require manual tuning of task vector coefficients, which increases computational cost and limits scalability. To mitigate this issue, AdaMerging~\cite{Adamerging_2024_ICLR} determines the coefficients adaptively through test-time training.
LiNeS~\cite{LiNeS_2025_ICLR} simplifies coefficient selection by increasing them linearly with layer depth.
To move beyond fixed coefficients for all inputs, router-based approaches~\cite{TwinMerging_2024, WeMoE_2024_ICML} dynamically compute instance-wise coefficients. Representation Surgery~\cite{Surgery_2024_ICML} further learns task-specific parameters at test time to align the merged model’s output with each individual task model.

As the number of tasks increases, storing multiple fine-tuned checkpoints becomes a major bottleneck. 
Despite efforts to address storage issues, existing methods like TALL Mask~\cite{Tall_Mask_2024_ICML} and TSV-C~\cite{gargiulo2025task} respectively rely on task-specific binary masks and perform SVD on the task matrix.
In contrast, we introduce a more universally applicable strategy that quantizes each task vector, significantly reducing storage overhead while preserving performance. Crucially, our approach integrates seamlessly with existing task vector-based methods, requiring only minimal modifications.

\subsection{Quantization}

Quantization methods for compressing pre-trained deep neural networks generally fall into two categories: Post-Training Quantization (PTQ) and Quantization-Aware Training (QAT). PTQ calibrates a pre-trained model by quantizing its weights and activations to low-bit representations while preserving accuracy~\cite{nahshan2021loss, fang2020post, liu2021post,banner2019post, finkelstein2019fighting, zhao2019improving, nagel2020up, li2021brecq, wei2022qdrop, hubara2020improving}.
QAT, on the other hand, integrates quantization into the training process via fine-tuning (often with the Straight-Through Estimator~\cite{yin2019understanding}), but it is more complex and computationally demanding~\cite{hubara2017quantized, choi2018pact, esser2019learned, zhou2016dorefa, zhang2018lq, yamamoto2021learnable}.
Our approach aligns with PTQ because we do not have access to training data for quantization in the model merging scenario. While prior PTQ methods typically focus on quantizing full model weights, we instead quantize the difference between pre-trained and fine-tuned checkpoints. These task vectors have a narrower weight range, making them well-suited for low-precision quantization and reducing quantization error more effectively than full-model quantization.

\section{Preliminary}
\label{sec:prelim}

\begin{figure}[t]
\begin{center}
\centering
\def\arraystretch{0.5}
\begin{tabular}{@{\hskip 0.01\linewidth}c@{\hskip 0.03\linewidth}c@{}c}
\includegraphics[width=1.00\linewidth]{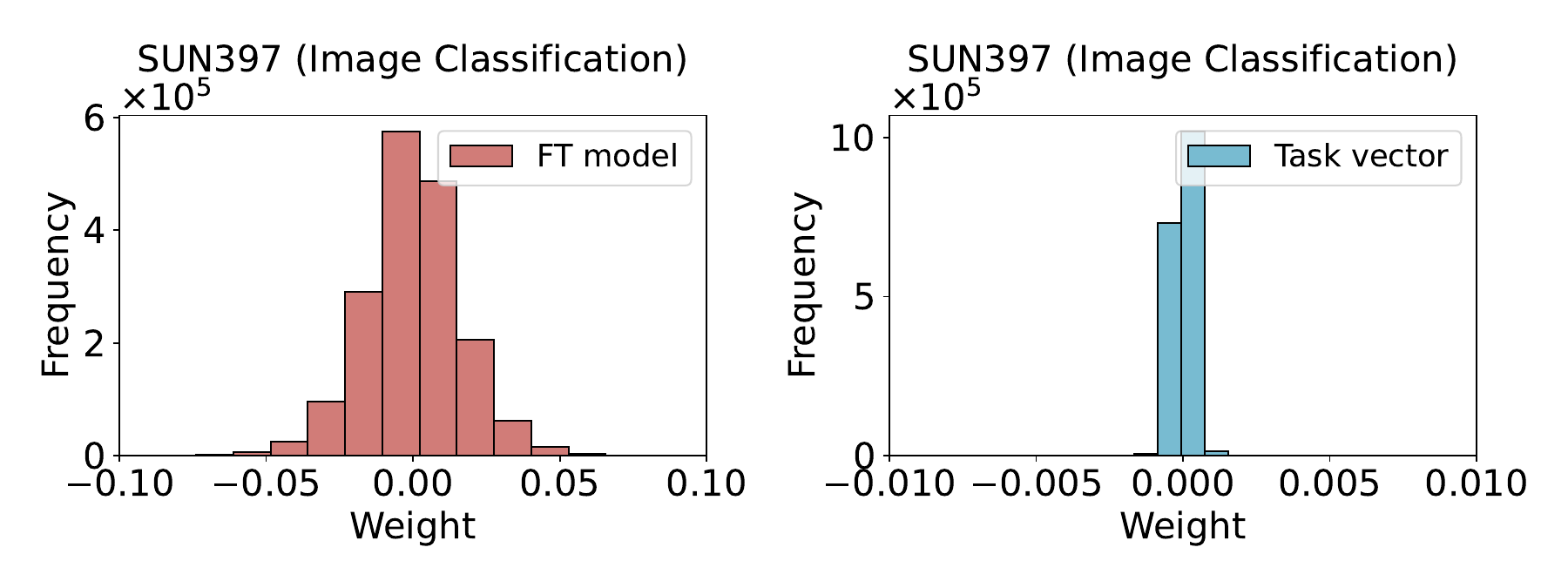}\\
\includegraphics[width=1.00\linewidth]{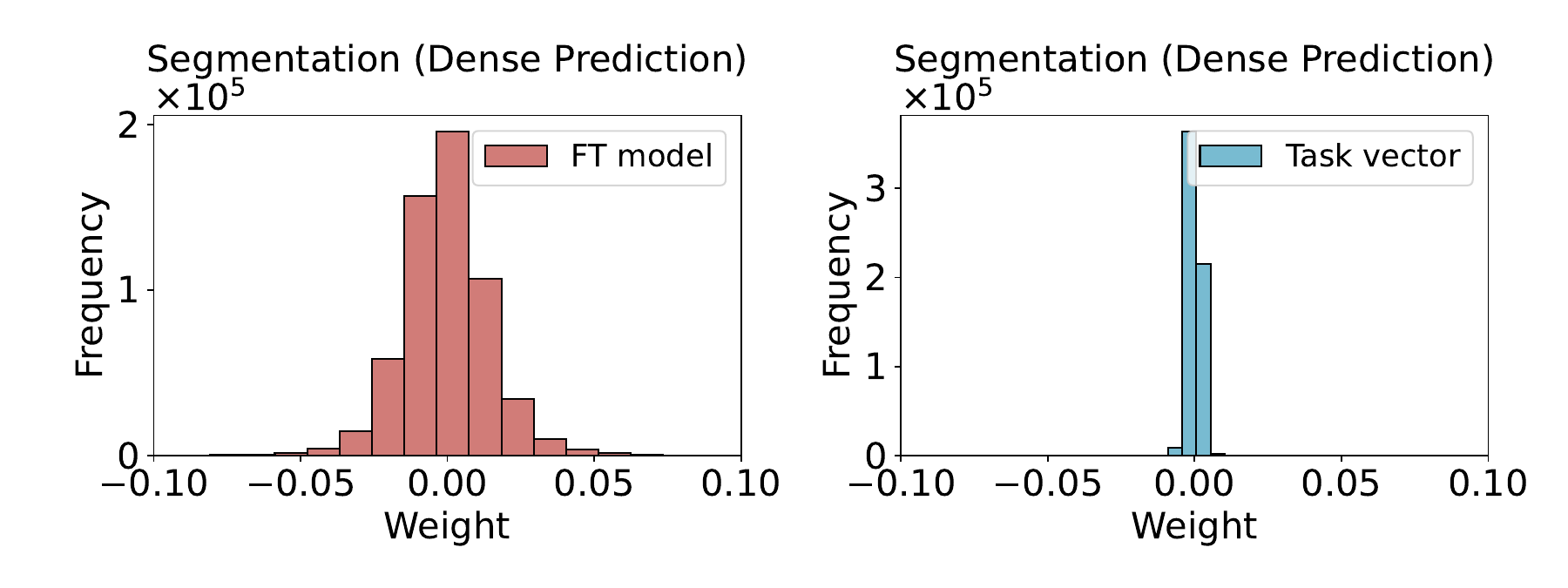} 
\end{tabular}
\end{center}
\vspace{-7mm}
\caption{Weight distribution comparison between the fine-tuned model (red) and the task vector (blue) for image classification (SUN397) and dense prediction (Segmentation). The task vector consistently shows a smaller weight range.}

 \vspace{-2mm}
\label{fig:method:observation}
\end{figure}

\subsection{Model Merging}
Let $\theta_{pre}$ be the parameters of a pre-trained model. Fine-tuning $\theta_{pre}$ on a specific task $t$ produces a task-specific model with parameters $\theta^t_{ft}$. The goal of model merging is to construct a unified multi-task model $\theta_{MTL}$ that effectively integrates multiple fine-tuned models $\theta^1_{ft}, \dots, \theta^T_{ft}$. A common approach to achieving this is \textit{Task Arithmetic}~\cite{TaskArithmetic_2023_ICLR}, which defines the task vector $\tau_t$ as the difference between the fine-tuned weights and the pre-trained weights: $\tau_t = \theta^t_{ft} - \theta_{pre}$. By leveraging a linear combination of task vectors, the merged model can be formulated as: $\theta_{MTL} = \theta_{pre} + \lambda \sum_{t=1}^{T} \tau_t$, where the coefficient \(\lambda\) is a scaling factor, shared across all tasks, controlling the relative importance of the task vectors in the merged model.

\subsection{Weight Quantization}

We demonstrate the fundamentals of asymmetric quantization, which is a widely used technique in neural network compression \cite{liu2023llm,you2024parameterize,liu2024kivi}. Asymmetric quantization maps full-precision floating-point values to a lower-bit integer representation by using distinct scaling factors and zero-points for each quantized tensor. This approach helps in better preserving the dynamic range of the original data.

Let $\mathbf{\theta} \in \mathbb{R}$ be a full-precision weight tensor. The quantization process transforms $\mathbf{\theta}$ into an integer vector $\theta^q \in \mathbb{Z}^n$ using the following affine mapping:  
\begin{equation}
    \theta^q = \text{Round}\left(\frac{\theta}{\Delta} \right) + z, \hspace{3ex} \Delta = \frac{\theta_{\max} - \theta_{\min}}{2^b - 1}, 
    \label{eq:quantization}
\end{equation}  
where $b$ is the number of bits used for quantization, $\Delta$ is the scaling factor, and $z = - \text{Round}(\frac{\theta_{\min}}{\Delta})$ is the zero-point. The scaling factor $\Delta$ ensures that the range $[\theta_{\min}, \theta_{\max}]$ is appropriately mapped to the target integer range, while the zero-point $z$ compensates for any asymmetry in the distribution of $\mathbf{\theta}$.  
The original values can be reconstructed from the quantized representation through the dequantization step:  
\begin{equation}
    \hat{\theta} = \Delta  (\theta^q - z).
\end{equation}

\section{Methodology}
\label{sec:method}

\begin{figure}[t]
\begin{center}
\centering
\def\arraystretch{0.5}
\begin{tabular}{@{\hskip 0.0\linewidth}c@{\hskip 0.0\linewidth}c@{}c}
\hspace{-4mm}
\includegraphics[width=0.67\linewidth]{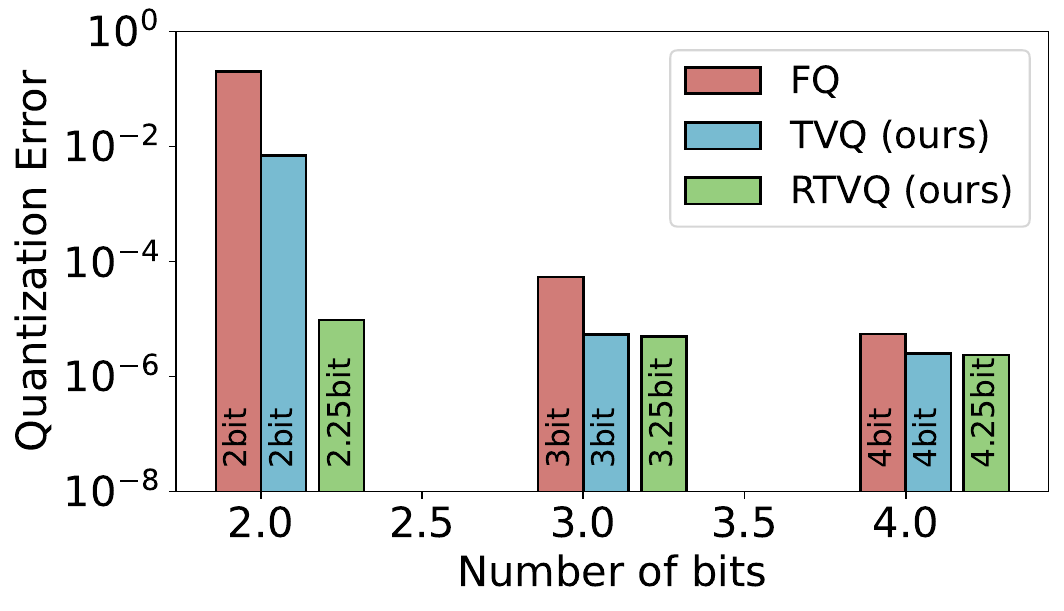}
\end{tabular}
\end{center}
\vspace{-5mm}
\caption{Comparison of quantization error across methods. The y-axis is logarithmic, with errors averaged over eight vision tasks and normalized by model parameters. Numbers inside bars indicate bit precision. FQ denotes fine-tuned checkpoint quantization, while TVQ and RTVQ represent task vector and residual task vector quantization, respectively.} \vspace{-2mm}
\label{fig:method:quant_error}
\end{figure}

\begin{figure*}[t]
\begin{center}
\centering
\def\arraystretch{0.5}
\begin{tabular}{@{\hskip 0.01\linewidth}c@{\hskip 0.03\linewidth}c@{}c}
\includegraphics[width=0.9\linewidth]{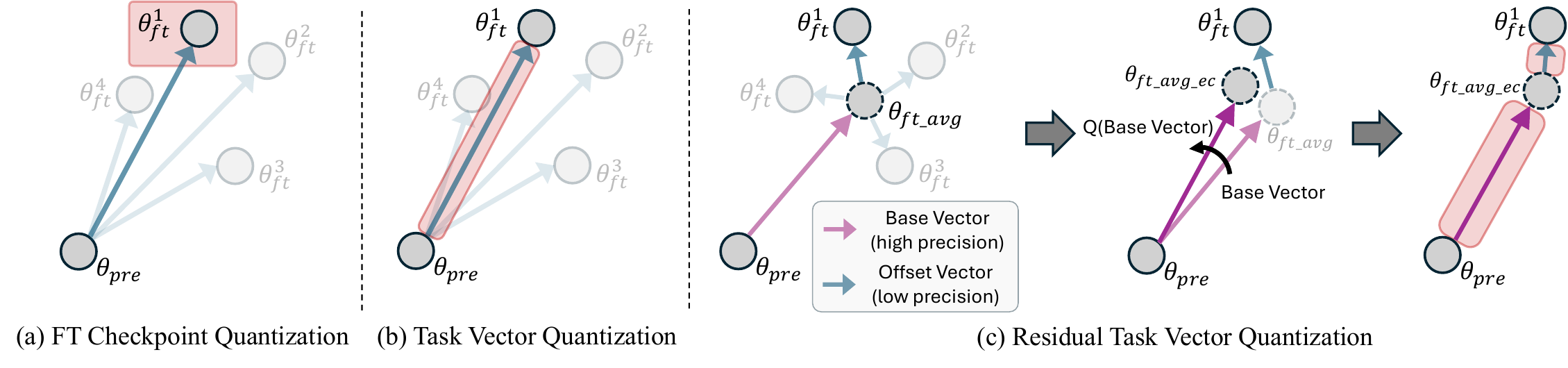} 
\end{tabular}
\end{center}
\vspace{-5mm}
\caption{An illustration of quantization methods for reducing model merging memory costs. Red boxes indicate stored quantized information. (a) Each of the fine-tuned checkpoints provided by prior work~\cite{TaskArithmetic_2023_ICLR} is directly quantized. (b) We propose quantizing the task vector instead, leveraging its smaller weight range for lower quantization error (see Section \ref{sec:method:TVQ}). (c) We further decompose the task vector into a high-precision base vector and multiple low-precision offset vectors. The base vector is computed as the difference between the averaged fine-tuned weights and the pre-trained weights, ensuring that the base vector is close to every fine-tuned model. In the middle figure, we illustrate a quantization error correction step by calibrating the averaged checkpoint before computing the offset vectors. 
 (see Section \ref{sec:advanced_quant}).}
\label{fig:method:vector}
\end{figure*}

\subsection{Observation}
\label{sec:method_observation}

Most previous studies store multiple checkpoints, each fine-tuned for a single task, for model merging \cite{Adamerging_2024_ICLR,TaskArithmetic_2023_ICLR,EMRMerging_2024}. One straightforward approach to reduce memory requirements is to directly quantize the fine-tuned checkpoints. Alternatively, one could quantize the task vector $\tau_t$, which represents the difference between the fine-tuned and pre-trained weights. This raises the question of which strategy provides better memory efficiency while preserving performance. To explore this, we revisit the quantization error introduced by the rounding operation, which is formulated as:
\begin{equation}
    |\epsilon| \leq \frac{\Delta}{2} = \frac{\theta_{\max} - \theta_{\min}}{2(2^b - 1)} ,
    \label{eq:quant_err_bound}
\end{equation}
where $\Delta$ is the scaling factor introduced in Eq. \ref{eq:quantization}. 
The key insight is that a narrower weight range $[\theta_{\min}, \theta_{\max}]$ leads to a smaller quantization error. Therefore, if we can identify a representation with a reduced dynamic range for storing checkpoints, it would result in a lower quantization error.
Based on this, we compare the weight ranges of the fine-tuned checkpoint $\theta^t_{ft}$, and its corresponding task vector $\tau_t$, as illustrated in Fig.~\ref{fig:method:observation}. The figure shows that the task vector has a weight range that is an order of magnitude smaller than that of the fine-tuned checkpoint. This trend is consistent across various datasets and architectures.

\subsection{Task Vector Quantization (TVQ)}
\label{sec:method:TVQ}
Based on this observation, we propose quantizing task vectors instead of fine-tuned checkpoints, leveraging their limited dynamic range for more efficient quantization. To validate this approach, we compute the distance between the full-precision task vector and the task vector obtained from the quantized checkpoint, $\text{Dist}(\tau_t, \hat{\theta}^{\,t}_{ft} - \theta_{pre})$ and compare it with the distance for the quantized task vector, $\text{Dist}(\tau_t, \hat{\tau}_t)$.
Here, we use the $L_2$ norm as our distance metric. As shown in Fig.~\ref{fig:method:quant_error}, quantizing the fine-tuned model yields significantly higher quantization error compared to quantizing the task vector. Moreover, our experiments in 8-task image classification scenario (see Table~\ref{tab:main_vitb32_8_tasks}) indicate that quantizing fine-tuned checkpoints fails in the low precision regime (\ie, below 4 bits), further underscoring the benefits of our proposed approach.
Fig.~\ref{fig:method:vector}(a) and Fig.~\ref{fig:method:vector}(b) illustrate fine-tuned checkpoint quantization and task vector quantization, respectively.

\subsection{Residual Task Vector Quantization (RTVQ)}
\label{sec:advanced_quant}
We observed that TVQ performs well when using more than 3 bits, but its performance degrades at extremely low bit precision (\eg, 2 bits). To further reduce quantization error in this low-bit regime, we extend our approach with Residual Task Vector Quantization (RTVQ).
The key idea behind RTVQ is to represent multiple task vectors using a single common base vector, along with several residual offset vectors that capture the differences between the base vector and each fine-tuned weights as shown in Fig.~\ref{fig:method:vector}(c). We compute the base vector as the average of all fine-tuned weights, ensuring that it is close to every fine-tuned model.
{Since this base vector is shared among all tasks, it needs to be stored only once, resulting in minimal overhead. Moreover, the base vector typically exhibits a larger weight range compared to the offset vectors, making it beneficial to assign higher precision to it.}
{Meanwhile,} the small magnitude of the offset vectors enables their quantization at low precision without causing significant performance degradation.

To be more precise, the task vector is defined as the difference between the fine-tuned and pre-trained checkpoints: $\tau_t = \theta^t_{ft} - \theta_{pre}$.
We decompose this task vector into an offset component and a base component as follows:
\begin{equation}
    \tau_t = \underbrace{(\theta^t_{ft} - \theta_{{ft\_avg}})}_{\text{Offset Vector}} 
    + \underbrace{(\theta_{{ft\_avg}} - \theta_{{pre}})}_{\text{Base Vector}},
    \label{eq:offset_base}
\end{equation}
where $\theta_{{ft\_avg}} = \frac{1}{N_t} \sum_t \theta^t_{ft}$ denotes an average of the fine-tuned checkpoints across $N_t$ tasks. 
As we mentioned, since the base vector is shared across all tasks and is independent of the number of tasks, its storage cost is relatively small compared to that of the offset vectors. For example, consider a scenario with 8 tasks. If the offset vector is quantized with 2 bits per task and the base vector with 4 bits, the effective bit requirement per task is: $2\,\text{bits} + \frac{4\,\text{bits}}{8\,\text{tasks}} = 2.5\,\text{bits}$.

Let $Q_{\text{err}}(\theta, b)$ represent the quantization error when a weight tensor $\theta$ is quantized with $b$ bits. Our key assumption is that there exists an assignment of bits such that quantizing the offset task vector with $b_o$ bits and the base vector with $b_b$ bits satisfies:
\begin{equation}
Q_{\text{err}}(\tau_t, b_t) \ge Q_{\text{err}}(\theta^t_{ft} - \theta_{{ft\_avg}}, b_o) + Q_{\text{err}}(\theta_{{ft\_avg}} - \theta_{{pre}}, b_b),
\end{equation}
where $b_t$ denotes the bitwidth for the task vector quantization, and $b_t \approx b_o + \frac{b_b}{N_t}$ to make sure the decomposition does not bring significant memory overhead compared to the original TVQ method.  Fig.~\ref{fig:method:quant_error} validates this assumption where RTVQ shows less quantization error with a similar number of bits. 
Moreover, the benefit of RTVQ increases significantly as bit precision is reduced.

\noindent \textbf{Quantization error correction.} Both the offset and base vectors in Eq.~\ref{eq:offset_base} can introduce quantization errors. To mitigate these errors, we propose a correction method that accounts for the quantization error of the base vector during the computation of the offset vector, as illustrated in Fig.~\ref{fig:method:vector}(c). 
Specifically, after computing the base vector, we calibrate the averaged fine-tuned model checkpoint from $\theta_{ft\_avg}$ to a corrected version $\theta_{ft\_avg\_ec}$, defined as
\begin{equation}
\theta_{ft\_avg\_ec} = Q\bigl(\theta_{ft\_avg} - \theta_{pre}\bigr) + \theta_{pre}.
\label{eq:err_correction}
\end{equation}
Here, $Q(\cdot)$ denotes the quantization operation.
We then compute the offset vector using the calibrated checkpoint: $\theta^t_{ft} - \theta_{ft\_avg\_ec}.$
This correction step refines the overall quantized representation and reduces the cumulative quantization error. Algorithm~\ref{algorithm: overall} illustrates the overall process of our proposed RTVQ.

\noindent \textbf{Seamless integration with model merging.} A key advantage of our approach is that it operates solely on the checkpoints. As a result, our quantization method can be seamlessly integrated into existing model merging frameworks without any modifications. Extensive experiments on image classification and dense vision tasks demonstrate that our approach maintains or even improves model performance while drastically reducing checkpoint storage requirements.

\begin{algorithm}[t]\small
        \caption{Residual Task Vector Quantization (RTVQ)}
       \textbf{Input}: pre-trained model checkpoint $\theta_{\text{pre}}$, fine-tuned model checkpoints $\{\theta^t_{\text{ft}}\}_{t=1}^{T}$, base vector precision $b_b$, offset vector precision $b_o$, quantization operator $Q(\cdot)$\\
      \textbf{Output}:  quantized task vector $\hat{\tau}_t$
      \begin{algorithmic}[1]
        \State {compute $\theta_{\text{ft\_avg}} = \frac{1}{T} \sum_{t=1}^{T} \theta^t_{\text{ft}}$}  
        \State {$\text{base\_vector} \gets \theta_{\text{ft\_avg}} - \theta_{\text{pre}}$}  
        \State {$\theta_{\text{ft\_avg\_ec}} \gets Q(\text{base\_vector}, b_b) + \theta_{\text{pre}}$}   
        \State {$\text{offset\_vector} \gets \theta^t_{\text{ft}} - \theta_{\text{ft\_avg\_ec}}$}
        \State {$\hat{\tau}_t \gets Q(\text{offset\_vector}, b_o) + Q(\text{base\_vector}, b_b)$}
      \end{algorithmic}
          \label{algorithm: overall}
\end{algorithm}

\section{Experiments}
\label{sec:exp}

\subsection{Experimental Setup}
\noindent\textbf{Datasets.} 
{We evaluate our method through a series of multi-task model merging experiments.
For image classification, we adopt the 8-task protocol from~\cite{TaskArithmetic_2023_ICLR} and expand to 14 and 20 tasks~\cite{Tall_Mask_2024_ICML} to challenge the scalability of our method against increasing storage requirements.
For dense prediction, we follow~\cite{tang2024fusionbench} with the NYUv2~\cite{silberman2012indoor}. 
For NLP tasks, we use the GLUE benchmark~\cite{GLUE_2018} with the same setup in~\cite{EMRMerging_2024}. Additional details are in the Appendix~\ref{appendix:setup}.}

\noindent\textbf{Models.} 
{In our image classification experiments, we utilize CLIP~\cite{CLIP_2021_ICML}'s visual encoder ViT-B/32 and ViT-L/14. 
For dense prediction tasks, we employ ResNet-50~\cite{ResNet_2016_CVPR} as the backbone and follow the protocol of~\cite{tang2024fusionbench} for each task. 
For NLP tasks, we use RoBERTa-base~\cite{RoBERTa_2019}, following the experimental setup of~\cite{EMRMerging_2024}. See Appendix~\ref{appendix:setup} for details.}

\begin{table*}[t]
    \centering
    \footnotesize 
    \begin{tabular}{l|c|ll|llll|l}
        \toprule
        \multirow{2}{*}{Method} & \multicolumn{1}{c|}{Baseline} & \multicolumn{2}{c|}{FQ} & \multicolumn{4}{c|}{TVQ (ours)} & \multicolumn{1}{c}{\multirow{2}{*}{RTVQ (ours)}} \\
        \cmidrule(lr){2-2} \cmidrule(lr){3-4} \cmidrule(lr){5-8} %
        & \multicolumn{1}{c|}{FP32} & \multicolumn{1}{c}{INT8} & \multicolumn{1}{c|}{INT4} & \multicolumn{1}{c}{INT8} & \multicolumn{1}{c}{INT4} & \multicolumn{1}{c}{INT3} & \multicolumn{1}{c|}{INT2} &  \\
        \midrule
        Individual & 90.5 & 90.4 \perf{-0.1} & 4.2 \perf{-86.3} & 90.5 \perf{0.0} & 90.5 \perf{0.0} & 90.7 \perf{0.2} & 83.5 \perf{-7.0} & \multicolumn{1}{c}{-} \\
        \midrule
        
        Task arithmetic~\cite{TaskArithmetic_2023_ICLR} & 69.2 & 68.1 \perf{-1.1} & 4.2 \perf{-65.0} & 69.0 \perf{-0.2} & 69.1 \perf{-0.1} & 71.2 \perf{2.0} & 62.1 \perf{-7.1} & 70.2 \perf{1.0} \\
        
        Ties merging~\cite{TiesMerging_2023_NeurIPS} & 72.9 & 64.8 \perf{-8.1} & 4.2 \perf{-68.7} & 72.7 \perf{-0.2} & 72.0 \perf{-0.9} & 73.6 \perf{0.7} & 62.6 \perf{-10.3} & 72.7 \perf{-0.2} \\

        LiNeS~\cite{LiNeS_2025_ICLR} & 74.1 & 73.9 \perf{-0.2} & 4.3 \perf{-69.8} & 74.2 \perf{0.1} & 74.2 \perf{0.1} & 74.7 \perf{0.6} & 60.7 \perf{-13.4} & 74.2 \perf{0.1}  \\
        
        Consensus TA~\cite{Tall_Mask_2024_ICML} & 74.9 & 70.6 \perf{-4.3} & 3.7 \perf{-71.2} & 74.9 \perf{0.0} & 74.9 \perf{0.0} & 74.8 \perf{-0.1} & 58.5 \perf{-16.4} & 72.7 \perf{-2.2}\\
        
        AdaMerging~\cite{Adamerging_2024_ICLR} & 81.8 & 81.6 \perf{-0.2} & 4.5 \perf{-77.3} & 81.6 \perf{-0.2} & 81.5 \perf{-0.3} & 82.0 \perf{0.2} & 78.1 \perf{-3.7} & 82.8 \perf{1.0} \\

        EMR-Merging~\cite{EMRMerging_2024} & 88.3 & 88.7 \perf{0.4} & 3.9 \perf{-84.4} & 88.4 \perf{0.1} & 89.8 \perf{1.5} & 90.0 \perf{1.7} & 77.2 \perf{-11.1} & 83.2 \perf{-5.1} \\

        \bottomrule
    \end{tabular}
    \caption{Comparison of quantization methods for merging 8 classification tasks using ViT-B/32. Our primary objective is to improve storage efficiency while minimizing performance degradation relative to the full-precision (FP32) baseline. FQ denotes fine-tuned checkpoint quantization, TVQ represents Task Vector Quantization (Section~\ref{sec:method:TVQ}), and RTVQ applies Residual Task Vector Quantization (Section~\ref{sec:advanced_quant}) with a 3-bit base vector and a 2-bit offset vector (2.375 bits per task). Average accuracy (\%) is reported, with differences from FP32 in parentheses (\textcolor{red}{red} for drops, \textcolor{ForestGreen}{green} for gains).}
    \label{tab:main_vitb32_8_tasks}
\end{table*}

\begin{table*}[t]
    \centering
    \footnotesize 
    \begin{tabular}{l|c|ll|llll|l}
        \toprule
        \multirow{2}{*}{Method} & \multicolumn{1}{c|}{Baseline} & \multicolumn{2}{c|}{FQ} & \multicolumn{4}{c|}{TVQ (ours)} & \multicolumn{1}{c}{\multirow{2}{*}{RTVQ (ours)}}  \\
        \cmidrule(lr){2-2} \cmidrule(lr){3-4} \cmidrule(lr){5-8} %
        & \multicolumn{1}{c|}{FP32} & \multicolumn{1}{c}{INT8} & \multicolumn{1}{c|}{INT4} & \multicolumn{1}{c}{INT8} & \multicolumn{1}{c}{INT4} & \multicolumn{1}{c}{INT3} & \multicolumn{1}{c|}{INT2} &  \\
        \midrule
        Individual & 94.2 & 94.2 \perf{0.0} & 4.3 \perf{-89.9} & 94.1 \perf{-0.1} & 94.2 \perf{0.0} & 94.2 \perf{0.0} & 90.9 \perf{-3.3} &\multicolumn{1}{c}{-} \\

        \midrule
        Task arithmetic~\cite{TaskArithmetic_2023_ICLR} & 84.3 & 84.1 \perf{-0.2} & 4.5 \perf{-79.8} & 84.3 \perf{0.0} & 84.4 \perf{0.1} & 84.8 \perf{0.5} & 77.9 \perf{-6.4} & 84.8 \perf{0.5}  \\
        
        Ties merging~\cite{TiesMerging_2023_NeurIPS} & 84.5 & 78.3 \perf{-6.2} & 4.2 \perf{-80.3} & 84.5 \perf{0.0} & 84.6 \perf{0.1} & 85.3 \perf{0.8} & 78.0 \perf{-6.5} & 81.6 \perf{-2.9} \\

        LiNeS~\cite{LiNeS_2025_ICLR} & 86.9 & 86.4 \perf{-0.5} & 5.4 \perf{-81.5} & 86.9 \perf{0.0} & 86.9 \perf{0.0} & 87.7 \perf{0.8} & 81.8 \perf{-5.1} & 87.7 \perf{0.8}  \\
        
        Consensus TA~\cite{Tall_Mask_2024_ICML} & 86.6 & 84.6 \perf{-2.0} & 4.2 \perf{-82.4} & 86.6 \perf{0.0} & 86.6 \perf{0.0} & 87.1 \perf{0.5} & 79.0 \perf{-7.6} & 87.3 \perf{0.7} \\
        
        AdaMerging~\cite{Adamerging_2024_ICLR} & 90.8 & 90.8 \perf{0.0} & 4.8 \perf{-86.0} & 90.9 \perf{0.1} & 90.9 \perf{0.1} & 91.0 \perf{0.2} & 89.4 \perf{-1.4} & 90.9 \perf{0.1}\\
        
        EMR-Merging~\cite{EMRMerging_2024} & 93.5 & 92.8 \perf{-0.7} & 4.4 \perf{-89.1} & 93.5 \perf{0.0} & 93.9 \perf{0.4} & 93.9 \perf{0.4} & 87.6 \perf{-5.9} & 90.3 \perf{-3.2} \\
                
        \bottomrule
    \end{tabular}
    
    \caption{Comparison of quantization methods for merging 8 classification tasks using ViT-L/14. For RTVQ, we use a 3-bit base vector and a 2-bit offset vector (equivalent to 2.375 bits per task).}
    \label{tab:main_vitl14_8_tasks}
\end{table*}

\subsection{Experimental Results}

To assess the impact of quantization on model merging, we first quantize the task-specific fine-tuned checkpoints, task vectors, and residual task vectors. Next, we apply these quantized weights to various merging methods that use task vectors. We compare all methods with their full-precision (FP32) counterparts across multiple tasks. Our goal is not to maximize absolute performance but to show that even with highly compact quantization, the model remains effective across multiple tasks.

\noindent\textbf{Merging 8 classification tasks.}
Table~\ref{tab:main_vitb32_8_tasks} and Table~\ref{tab:main_vitl14_8_tasks} show that 8-bit FQ results in a slight but acceptable performance drop.
However, reducing the precision to 4-bit causes significant degradation. This confirms our earlier observation that large quantization errors make model merging much harder.
In contrast, TVQ show much better stability. 
Even at 4-bit and 3-bit precision, model accuracy stays close to FP32.
Surprisingly, at 3-bit, some merging methods surpass their FP32 baselines.
This suggests that quantization reduces overfitting and improves generalization (see Section~\ref{sec:analyses} for details).
However, at 2-bit, performance drops sharply, indicating that excessive compression introduces substantial quantization noise.
RTVQ overcomes these limitations with a robust approach. It decomposes each task vector into a shared base vector (3 bits) and a residual vector (2 bits), requiring about 2.375 bits per task. This vector decomposition reduces the performance drop seen in 2-bit TVQ while preserving most of the accuracy benefits of higher-bit quantization.

\begin{table*}[!t]
    \centering
    \footnotesize 
    \setlength{\tabcolsep}{2.5pt} %
    \begin{tabular}{l|llll|llll|llll}
        \toprule
        \multirow{2}{*}{Method} & \multicolumn{4}{c|}{Segmentation $\uparrow$} & \multicolumn{4}{c|}{Depth estimation $\downarrow$} & \multicolumn{4}{c}{Normal estimation $\downarrow$}  \\
        \cmidrule(lr){2-5} \cmidrule(lr){6-9} \cmidrule(lr){10-13}
        & \multicolumn{1}{c}{FP32} & \multicolumn{1}{c}{TVQ-INT4} & \multicolumn{1}{c}{TVQ-INT2} & \multicolumn{1}{c|}{RTVQ} & \multicolumn{1}{c}{FP32} & \multicolumn{1}{c}{TVQ-INT4} & \multicolumn{1}{c}{TVQ-INT2} & \multicolumn{1}{c|}{RTVQ}  & \multicolumn{1}{c}{FP32} & \multicolumn{1}{c}{TVQ-INT4} & \multicolumn{1}{c}{TVQ-INT2} & \multicolumn{1}{c}{RTVQ}  \\
        \midrule
        Individual & 52.0 & 52.0 \perf{0.0}& 37.7 \perf{-14.3}&\multicolumn{1}{c|}{-} & 41.5 & 41.4 \revperf{-0.1}& 62.5 \revperf{21.0}&\multicolumn{1}{c|}{-} & 24.2 & 24.2 \revperf{0.0}& 34.2 \revperf{10.0}& \multicolumn{1}{c}{-} \\
        \midrule

        Task arithmetic~\cite{TaskArithmetic_2023_ICLR} & 31.6 & 31.5 \perf{-0.1}  & 36.4 \perf{4.8} & 36.1 \perf{4.5} & 24.0 & 24.0 \revperf{0.0} & 26.2 \revperf{2.2} & 24.6 \revperf{0.6} & 30.6 & 30.6 \revperf{0.0} & 36.1 \revperf{5.5} & 32.6 \revperf{2.0} \\
        
        Ties merging~\cite{TiesMerging_2023_NeurIPS} & 39.9 & 40.0 \perf{0.1} & 36.1 \perf{-3.8} & 37.0 \perf{-2.9} & 27.3 & 27.2 \revperf{-0.1} & 26.5 \revperf{-0.8} & 24.6 \revperf{-2.7} & 36.2 & 36.2 \revperf{0.0} & 37.0 \revperf{0.8} & 32.6 \revperf{-3.6} \\
        
        MagMax~\cite{MagMax_2024_ECCV} & 24.7 & 25.4 \perf{0.7} & 29.9 \perf{5.2} & 29.4 \perf{4.7} & 23.9 & 24.2 \revperf{0.3} & 25.6 \revperf{1.7} & 24.7 \revperf{0.8} & 30.3 & 30.0 \revperf{-0.3} & 32.2 \revperf{1.9} & 31.1 \revperf{0.8} \\

        Breadcrumbs~\cite{Breadcrumbs_2024_ECCV} & 34.1 & 34.3 \perf{0.2} & 32.2 \perf{-1.9} & 34.0 \perf{-0.1} & 27.2 & 27.2 \revperf{0.0} & 28.4 \revperf{1.2} & 27.7 \revperf{0.5} & 36.9 & 37.0 \revperf{0.1} & 40.6 \revperf{3.7} & 38.3 \revperf{1.4} \\

        EMR-Merging~\cite{EMRMerging_2024} & 41.5 & 44.8 \perf{3.3} & 21.3 \perf{-20.2} & 34.1 \perf{-7.4} & 19.4 & 18.8 \revperf{-0.6} & 25.5 \revperf{6.1} & 22.1 \revperf{2.7} & 26.5 & 26.6 \revperf{0.1} & 45.2 \revperf{18.7} & 35.0 \revperf{8.5} \\
        \bottomrule
    \end{tabular}
    
    \caption{Comparison of our proposed quantization methods on three dense prediction tasks from the NYUv2 dataset, using ResNet-50 models. We evaluate segmentation using mIoU ($\uparrow$), depth estimation using Relative Error ($\downarrow$), and normal estimation using Mean angular error ($\downarrow$). RTVQ quantizes both the base vector and offset to 2 bits. Note that RTVQ was designed to maintain stable performance at extremely low bit-widths, effectively mitigating the performance degradation observed in 2-bit quantization. {Similar to the classification tasks, performance significantly drops in FQ (see Appendix~\ref{appendix:supple_dense}).}}
    \label{tab:main_dense_pred_tasks}
\end{table*}

\begin{figure}[t]
    \centering
    \includegraphics[width=0.4\textwidth]{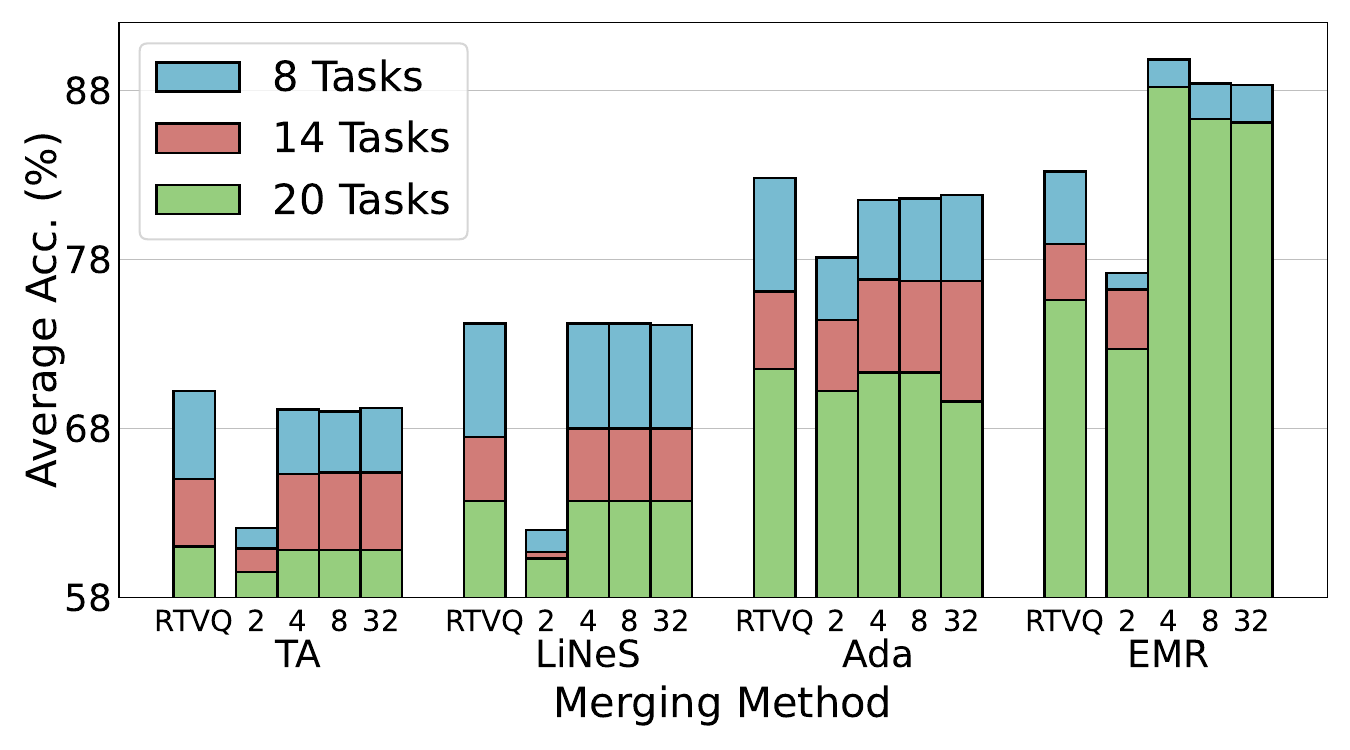}
    \vspace{-1em}
    \caption{Evaluation of the our quantization methods when scaling to 8, 14, and 20 tasks. 
    We compare diverse merging methods using 2-bit to 8-bit TVQ and an FP32 baseline.
    RTVQ mitigates performance degradation at extremely low precision (around 2 bits) even as the number of tasks increases.}
    \label{fig:more_vision}
    \vspace{-1mm}
\end{figure}

\noindent\textbf{Merging 14 and 20 classification tasks.} We further investigate the scalability of our quantization method. With 14 and 20 tasks, storing full-precision fine-tuned checkpoints becomes more impractical and highlights the need for effective quantization.
Fig.~\ref{fig:more_vision} illustrates the performance results for 8, 14, and 20 tasks under various quantization bitwidths. 
Consistent with our previous results, performance remains close to the FP32 baseline even at 4-bit precision, regardless of the number of tasks.
While 2-bit TVQ still introduces noticeable performance drops, the degradation reduces as more tasks are merged, possibly due to improved generalization from diverse task vectors. 
Additionally, RTVQ consistently maintains stable performance at significantly low bit, requiring approximately 2.375, 2.2, and 2.15 bits per task for 8, 14, and 20 tasks, respectively.
Note that, as the base vector is globally shared among tasks, RTVQ scales favorably by effectively reducing the per-task bit requirement as task numbers increase.

\begin{figure}[t]
    \centering
    \includegraphics[width=0.48\textwidth]{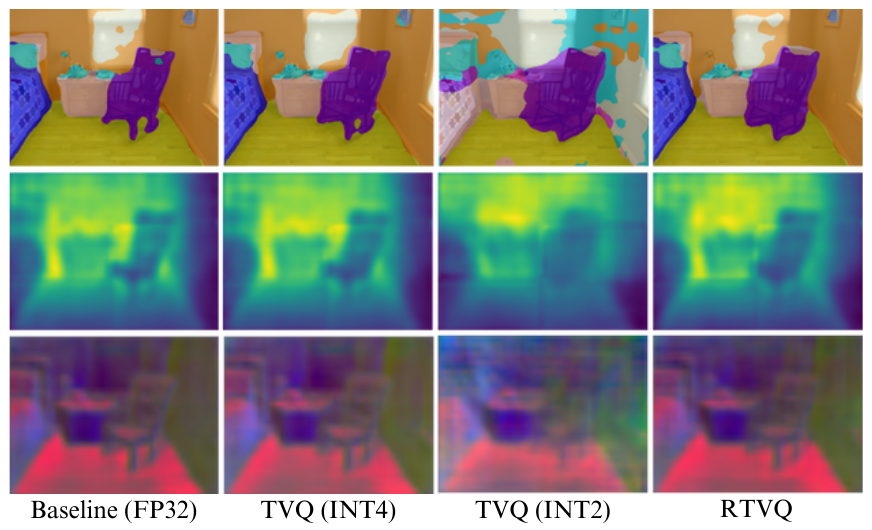}
    \caption{Qualitative results on segmentation (top), depth estimation (middle), and normal estimation (bottom), comparing the FP32 baseline and our quantization methods. Notably, RTVQ, which operates at a similar bit precision as 2-bit TVQ, preserves structural details better.
    }
    \label{fig:qualitative_results}
\end{figure}

\noindent\textbf{Merging dense prediction tasks.}
Table~\ref{tab:main_dense_pred_tasks} presents the performance of our methods on dense prediction tasks, including semantic segmentation, depth estimation, and normal estimation.  
Overall, TVQ maintains performance up to 4-bit, but normal estimation shows a noticeable drop at 2-bit precision. 
In particular, applying 2-bit TVQ to EMR-Merging causes the largest performance degradation across all tasks. 
Other merging methods mitigate this to some extent by improving cross-task performance\footnote{We define the task on which the model is fine-tuned as the target task and the others as cross tasks. While 2-bit TVQ reduces target task performance, it improves cross-task performance. See \textsection{5.3} for details.}.  
However, unlike classification, dense prediction tasks have lower task similarity, making EMR-Merging more sensitive to quantization errors due to its reliance on task-specific masks.  
RTVQ significantly outperforms 2-bit TVQ, highlighting the importance of reducing quantization error with base vectors to preserve task information.  
Fig.~\ref{fig:qualitative_results} shows qualitative comparisons, demonstrating that our quantization methods effectively capture scene structures. Additional results are provided in Appendix~\ref{appendix:supple_dense}.

\subsection{Analyses}
\label{sec:analyses}

\begin{table}[t]
    \centering
    \footnotesize
    \renewcommand{\arraystretch}{1.1}
    \setlength{\tabcolsep}{4pt} %
    
    \begin{tabular}{lcccccc}
        \toprule
        {Task} & {Baseline} & \multicolumn{4}{c}{{TVQ (ours)}} & {RTVQ (ours)} \\
        \cmidrule(lr){2-2} \cmidrule(lr){3-6} \cmidrule(lr){7-7}
        & FP32 & INT8 & INT4 & INT3 & INT2 & B3O2  \\
        \midrule
        Target  & 90.5 & 90.5 & 90.5 & \textbf{90.7} & 83.5 & 88.1  \\
        Cross  & 38.7 & 38.7 & 38.7 & 41.5 & 48.4 & \textbf{58.2}  \\
        \bottomrule
    \end{tabular}
 
    \caption{Comparison of target and cross-task performance across 8 classification tasks using ViT-B/32. Each task is the target once, with the remaining 7 as cross tasks. Results are averaged for overall accuracy. B3O2 denotes RTVQ with a 3-bit base vector and 2-bit offset vector.}
    \label{tab:task-vector-comparison}
\end{table}

\begin{figure}[t]
    \centering
    \begin{subfigure}[b]{0.43\columnwidth}
        \centering
        \includegraphics[width=\textwidth]{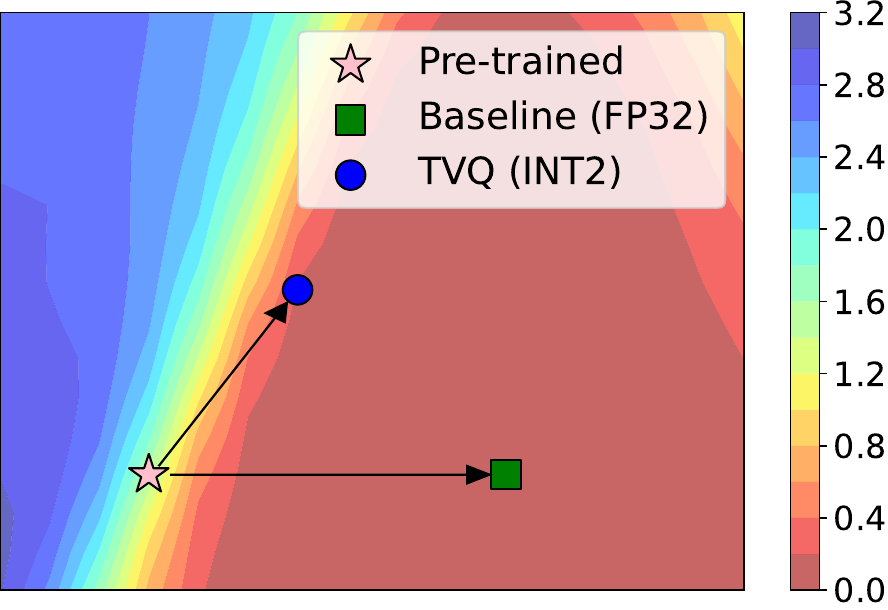}
        \caption{Target task}
        \label{fig:loss_landscape_cross_specific}
    \end{subfigure}
    \hfill
    \begin{subfigure}[b]{0.43\columnwidth}
        \centering
        \includegraphics[width=\textwidth]{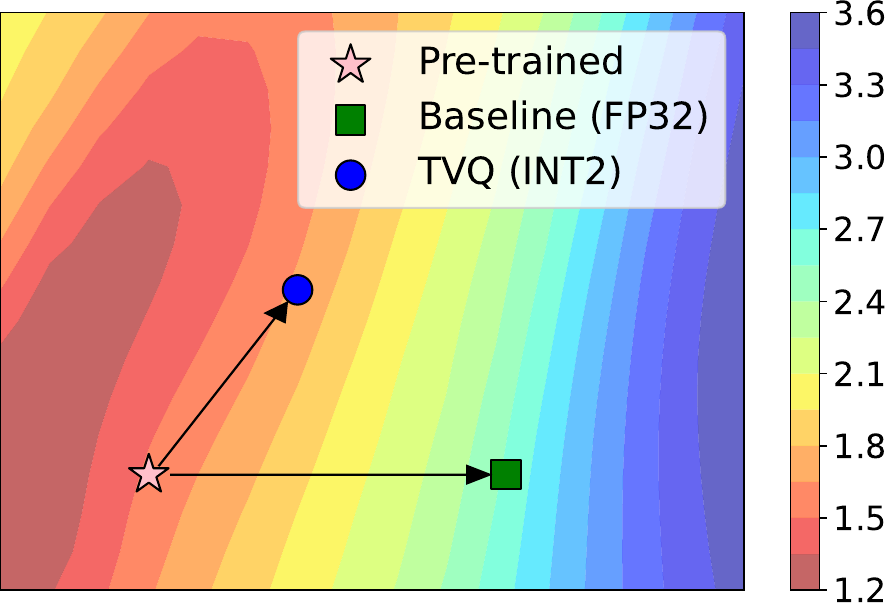}
        \caption{Cross task}
        \label{fig:loss_landscape_cross}
    \end{subfigure}
    \caption{Loss landscape visualization of the baseline and 2-bit TVQ. (a) uses the EuroSAT model, and (b) uses the GTSRB model, both evaluated on the EuroSAT task. TVQ increases target task loss but lowers cross task loss compared to the baseline.}
    \label{fig:loss_landscape}
    \vspace{-3mm}
\end{figure}

\noindent\textbf{Enhancing cross-task generalization.}
One might wonder how compressing a task vector to such a low bit-width can preserve or even improve performance, as observed with RTVQ. To investigate this, we evaluate target and cross-task performance across 8 classification tasks using ViT-B/32.  
Table~\ref{tab:task-vector-comparison} shows that quantization up to 3-bit (TVQ INT3) maintains target task accuracy at a level comparable to FP32 while significantly improving cross-task performance. When further reducing precision to 2-bit, task-specific details are lost, leading to a noticeable drop in target task accuracy. However, this extreme compression unexpectedly enhances cross-task generalization, with accuracy increasing from 38.7\% to 48.4\%. RTVQ provides an even better trade-off between target and cross-task performance. It boosts cross-task accuracy to 58.2\% while limiting the drop in target task accuracy. This suggests that by reducing quantization error, RTVQ better preserves task-specific properties while benefiting from improved cross-task generalization.  
Fig.~\ref{fig:loss_landscape} further illustrates how quantized task vectors deviate from their original directions in the loss landscape~\cite{mode_connectivity_fge_2018_NeurIPS}, sometimes shifting toward directions beneficial for other tasks. RTVQ follows a similar trend to TVQ. See Appendix~\ref{appendix:loss_landscape} for further details.

\begin{figure}[t]
    \centering
    \begin{subfigure}[b]{0.47\columnwidth}
        \centering
        \includegraphics[width=\textwidth]{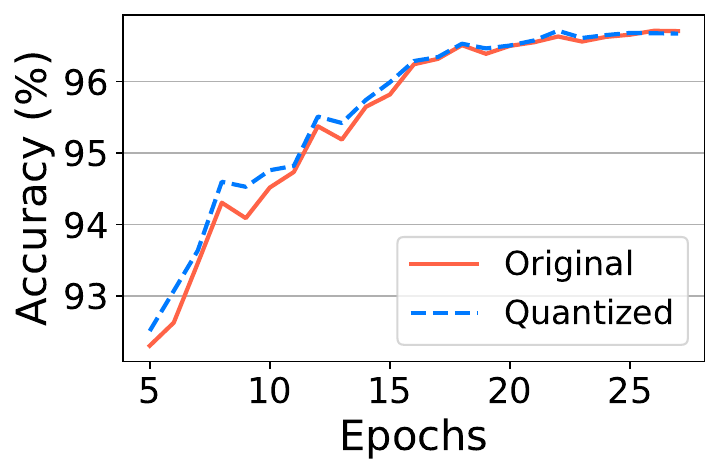}
        \caption{Training accuracy}
        \label{fig:epochs_train_acc}
    \end{subfigure}
    \hfill
    \begin{subfigure}[b]{0.47\columnwidth}
        \centering
        \includegraphics[width=\textwidth]{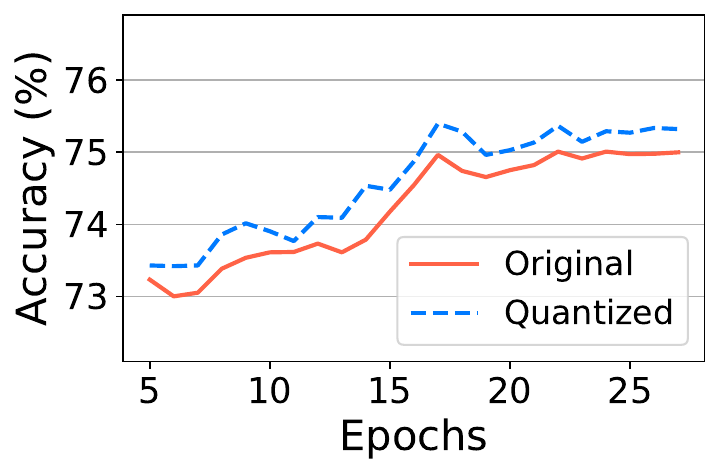}
        \caption{Test accuracy}
        \label{fig:epochs_test_acc}
    \end{subfigure}
    \vspace{-1mm}
    \caption{Comparison of training and test accuracy between the original and 3-bit quantized task vectors on SUN397 using ViT-B/32 over training epochs.}
    \label{fig:epochs_acc_error}
\end{figure}

\noindent\textbf{Overfitting reduction with quantization.}
Although task vector quantization reduces precision by using fewer bits, which may lead to some loss of task-specific information, previous experiments show that performance can sometimes improve. We attribute this improvement to the regularization effect of quantization, which reduces overfitting~\cite{abdolrashidi2021pareto,askarihemmat2022qreg}.  
To test this hypothesis, we evaluate the most challenging dataset among the 8 classification tasks, SUN397~\cite{SUN397_2010_CVPR}.  
We compare models trained on SUN397 using the original task vector and a 3-bit quantized task vector (TVQ).  As shown in Fig.~\ref{fig:epochs_train_acc}, both task vectors yield nearly identical training accuracy across epochs.  However, Fig.~\ref{fig:epochs_test_acc} shows that the quantized task vector consistently achieves higher test accuracy throughout training.  
This confirms that task vector quantization suppresses overfitting and enhances generalization.  
{We further suggest that quantization suppresses small-magnitude weights in task vectors to zero more significantly than in standard fine-tuned weights, which induces pruning and improves model performance. See Appendix~\ref{appendix:quantization_analyses} for details.}

\begin{figure}[t]
\begin{center}
\centering
\def\arraystretch{0.5}
\begin{tabular}{@{\hskip 0.0\linewidth}c@{\hskip 0.0\linewidth}c@{}c}
\hspace{-4mm}
\includegraphics[width=0.99\linewidth]{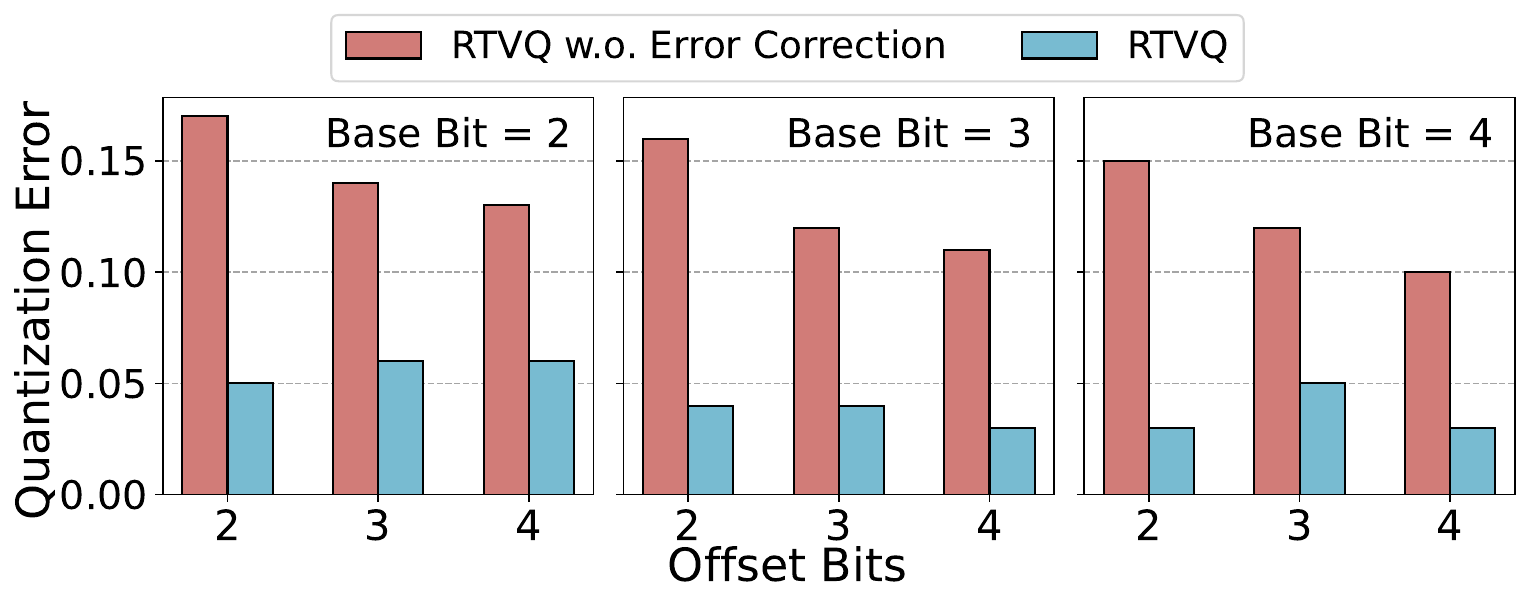} 
\end{tabular}
\end{center}
\vspace{-5mm}
\caption{Comparison of quantization error for Residual Task Vector Quantization (RTVQ) with and without error correction across different base and offset bit configurations.}
 \vspace{-2mm}
\label{fig:exp:quant_error_bitcorrection}
\end{figure}

\noindent\textbf{Error correction in RTVQ.}
Fig.~\ref{fig:exp:quant_error_bitcorrection} compares the quantization error of RTVQ with and without error correction across different base bit configurations and offset bit settings (2, 3, and 4). The results indicate that the inclusion of error correction in RTVQ significantly reduces quantization error across all configurations. Notably, as the base bit increases, the quantization error generally decreases for both methods, but the reduction is more pronounced in RTVQ with error correction. This highlights the effectiveness of error correction in minimizing quantization artifacts, particularly when the weight range is constrained by lower-bit representations.

\begin{table}[t]
    \centering
    \footnotesize
    \renewcommand{\arraystretch}{1.1}
    \setlength{\tabcolsep}{6pt} %
    
    \begin{tabular}{cccccc}
        \toprule
        {\# Tasks} & {Baseline} & \multicolumn{3}{c}{{TVQ (ours)}} & {RTVQ (ours)} \\
        \cmidrule(lr){2-2} \cmidrule(lr){3-5} \cmidrule(lr){6-6}
        & FP32 & INT8 & INT4  & INT2 & B3O2  \\
        \midrule
        8    & 9.1 GB  & 2.3 GB & 1.1 GB  & 0.6 GB & 0.7 GB  \\
        14   & 16.0 GB & 4.0 GB & 2.0 GB  & 1.0 GB & 1.2 GB  \\
        20   & 22.8 GB & 5.7 GB & 2.9 GB  & 1.4 GB & 1.7 GB  \\
        \bottomrule
    \end{tabular}
    
    \caption{Comparison of practical storage consumption for ViT-L/14 under different quantization schemes. The table presents the storage requirements for different numbers of tasks. B3O2 denotes RTVQ with a 3-bit base vector and a 2-bit offset vector.}
    \label{tab:paratical-memory-comparison}
    \vspace{-2mm}
\end{table}

\noindent\textbf{Storage cost comparison.}
Table~\ref{tab:paratical-memory-comparison} demonstrates the substantial storage reduction achieved by our techniques compared to the FP32 baseline for the ViT-L/14 model. As the number of tasks increases, storage consumption scales accordingly across all bit precisions. The FP32 baseline demands the highest memory, reaching 22.8 GB for 20 tasks. The proposed TVQ method significantly reduces memory usage, achieving up to a 16× reduction (6.25\%) with INT2. Notably, RTVQ with B3O2 not only maintains a compact footprint (7.5\% compared to FP32) across different task sizes but also outperforms INT2 in accuracy, as shown in previous experiments (see Table~\ref{tab:main_vitl14_8_tasks}).

\noindent\textbf{{Comparison with prior compression methods.}}
{To demonstrate the effectiveness of our method, we compare against prior approaches such as TALL-Mask \cite{Tall_Mask_2024_ICML} and TSV-C \cite{gargiulo2025task}. Following the TALL-Mask protocol, we evaluate both storage and accuracy.  As shown in Table \ref{tab:compression}, our 3-bit TVQ achieves the highest compression among all methods. It also supports flexible compression via adjustable bitwidths. When applied to the singular components of TSV-C, our quantization yields additional savings with minimal accuracy loss.}

\begin{table}[h]
  \centering
  \scriptsize
  \setlength{\tabcolsep}{2pt}       %
  \begin{minipage}[t]{0.48\columnwidth}
    \vspace{2mm}
    \centering
    \begin{tabular}{lcc}
      \toprule
      Method & Acc.\,(\%)$\uparrow$ & Bits (Gb)$\downarrow$ \\ \midrule
      TALL Mask      & 95.7 & 23.1 \\
      TSV-C          & 95.7 & 20.7 \\
      \midrule
      Ours           & 95.9 & 18.3 \\
      TSV-C + Ours   & 95.7 & 11.9 \\ \bottomrule
    \end{tabular}
    \vspace{-0.1mm}
        \captionof{table}{Comparison on ViT-B/32 across 8 tasks.}
    \label{tab:compression}
    \vspace{-5mm}
  \end{minipage}\hfill%
  \begin{minipage}[t]{0.49\columnwidth}
    \vspace{0mm}
    \centering
    \includegraphics[width=\textwidth]{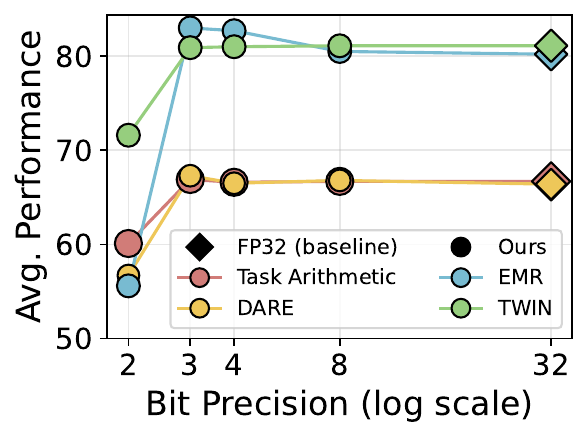}
    \vspace{-7.2mm}
    \captionof{figure}{NLP tasks analysis.}
    \label{fig:nlp_task}
  \end{minipage}
\end{table}

\noindent\textbf{{Task generalizability of the proposed method.}}
{To assess the generality of our method, we conducted additional experiments on NLP tasks (Fig.~\ref{fig:nlp_task}). Remarkably, our approach maintains stable and competitive performance, closely matching the full-precision baseline even at 3-bit quantization. This demonstrates the robustness and adaptability of our method across diverse tasks and compression levels.
}

\section{Conclusion}
\label{sec:conclusion}

{We address the memory overhead in model merging by introducing Task Vector Quantization, which exploits the narrow weight range of task vectors for effective low-precision storage. To further reduce errors, we propose Residual Task Vector Quantization, decomposing vectors into base and residual parts for higher accuracy where needed. Experiments on various tasks show that our method preserves or improves performance while reducing storage by up to 92\% without any architectural changes.
}

\section*{Acknowledgment}
\label{sec:acknowledgment}

This research was partly supported by the MSIT(Ministry of Science and ICT), Korea, under the Graduate School of Metaverse Convergence support program(IITP-2025-RS-2023-00254129) supervised by the IITP(Institute for Information \& Communications Technology Planning \& Evaluation). This research was partly supported by the MSIT(Ministry of Science, ICT), Korea, under the Global Research Support Program in the Digital Field program(RS-2024-00425354) supervised by the IITP(Institute for Information \& Communications Technology Planning \& Evaluation). This research was partly supported by the MSIT(Ministry of Science and ICT), Korea, under the ICAN(ICT Challenge and Advanced Network of HRD) support program(RS-2024-00436934) supervised by the IITP(Institute for Information \& Communications Technology Planning \& Evaluation).

{
    \small
    \bibliographystyle{ieeenat_fullname}
    \bibliography{main}
}

\clearpage
\twocolumn[
  \begin{@twocolumnfalse}
    \begin{center}
        {\Large \bf Supplementary Material: \\ Task Vector Quantization for Memory-Efficient Model Merging \par}
        \vspace*{12pt}
        {\large
        Youngeun Kim\textsuperscript{1*} \quad
        Seunghwan Lee\textsuperscript{2*} \quad
        Aecheon Jung\textsuperscript{2*} \quad
        Bogon Ryu\textsuperscript{2} \quad
        Sungeun Hong\textsuperscript{2†}\\[5pt]
        \textsuperscript{1}Yale University \quad \textsuperscript{2}Sungkyunkwan University\\[5pt]
        {\tt\small youngeun.kim@yale.edu \qquad \{simon2, kasurashan, bogon.ryu, csehong\}@skku.edu}
        \par}
        \vspace*{24pt}
    \end{center}
  \end{@twocolumnfalse}
]

\appendix
\renewcommand\thefigure{\Alph{figure}}    
\setcounter{figure}{0}  
\renewcommand\thetable{\Alph{table}}
\setcounter{table}{0}

\noindent  This Appendix provides an overview of our experimental details, further empirical analyses, and additional results. The detailed descriptions of each section are summarized as follows:

\begin{itemize}
    \item {Appendix~\ref{appendix:setup}}: Details of the datasets, model checkpoints, and merging methods used in our experiments.
    \item {Appendix~\ref{appendix:more_analyses}}: Further analyses on the effects of quantization in terms of weight pruning and task vector similarity; includes a sensitivity analysis on bit allocations.
    \item {Appendix~\ref{appendix:more_results}}: Detailed results on the 14 and 20 classification tasks, task-level performance, additional quantitative and qualitative results for dense prediction tasks, and loss landscape visualizations.

\end{itemize}

\section{Experimental Details}\label{appendix:setup}

\subsection{Employed datasets}

We evaluated our approach on a broad and diverse set of datasets that span three major categories: image classification, dense prediction, and natural language processing.

For image classification, we conducted experiments on a total of 19 datasets that cover a wide range of domains and visual characteristics. Specifically, we used SUN397~\cite{SUN397_2010_CVPR} for scene recognition, Cars~\cite{Cars_2013_ICCVW} for fine-grained vehicle classification, and RESISC45~\cite{RESISC45_2017} and EuroSAT~\cite{EuroSAT_2019} for remote sensing imagery. We also included digit and character recognition datasets such as SVHN~\cite{SVHN_2011}, MNIST~\cite{MNIST_1998}, EMNIST~\cite{EMNIST_2017}, FashionMNIST~\cite{FashionMNIST_2017}, and KMNIST~\cite{KMNIST_2018}. Additional benchmarks included GTSRB~\cite{GTSRB_2011} for traffic sign recognition, DTD~\cite{DTD_2014} for texture classification, CIFAR-10/100~\cite{CIFAR_10_100_2009} and STL10~\cite{STL10_2011} for general object recognition, FER2013~\cite{FER2013_2013} for facial expression recognition, Flowers102~\cite{Flowers102_2008} and Oxford-IIIT Pet~\cite{OxfordIIITPet_2012} for fine-grained species classification, PCAM~\cite{PCAM_2018} for histopathology image analysis, Food101~\cite{Food101_2014} for food recognition, and Rendered SST-2~\cite{RenderedSST2_2019}, which contains rendered images generated from sentiment classification data.

For dense prediction tasks, we used NYUv2~\cite{silberman2012indoor}, which provides RGB-D indoor images annotated for multiple tasks, including 13-class semantic segmentation, depth estimation, and surface normal estimation. This dataset was chosen to evaluate the ability of our method to handle multi-modal dense prediction problems.

Finally, for NLP tasks, we adopted the GLUE benchmark~\cite{GLUE_2018}, which consists of multiple language understanding tasks. We reported the Matthews correlation coefficient for CoLA~\cite{CoLA_2019}, Pearson and Spearman correlations for STS-B~\cite{STSB_2017}, and accuracy for the remaining tasks, including SST-2~\cite{SST2_2013}, MRPC~\cite{MRPC_2005}, QQP~\cite{QQP_2017}, MNLI~\cite{MNLI_2017}, QNLI~\cite{QNLI_2016_EMNLP}, and RTE~\cite{RTE_2007}. These benchmarks comprehensively evaluate our method across different input modalities and task types.

{\subsection{Model checkpoints}
For our experiments, we address three distinct domains. In image classification, we utilize Vision Transformers of varying scales: ViT-B/32 and ViT-L/14, initialized with pretrained CLIP weights~\cite{CLIP_2021_ICML}\footnote{\url{https://github.com/mlfoundations/open_clip}}.
To ensure a fair comparison across the 8 classification tasks, we adopt the publicly released model checkpoints from Task Arithmetic~\cite{TaskArithmetic_2023_ICLR}\footnote{\url{https://github.com/mlfoundations/task_vectors}}. For our extended experiments with 14 and 20 classification tasks, we fine-tune CLIP ViT-B/32 following Tall-Mask~\cite{Tall_Mask_2024_ICML} to obtain task-specific models. 
For dense prediction tasks, we employ ResNet-50~\cite{ResNet_2016_CVPR} as the backbone, initializing with ImageNet pre-trained weights. We then fine-tune this model following FusionBench~\cite{tang2024fusionbench} for segmentation, depth estimation, and normal estimation.
For NLP tasks, we use RoBERTa-base~\cite{RoBERTa_2019} with the publicly available fine-tuned weights from EMR-Merging~\cite{EMRMerging_2024}\footnote{\url{https://github.com/harveyhuang18/EMR_Merging}}.}

\subsection{Merging method baseline}

We applied our quantization method against diverse merging strategies, from simple to advanced. Notably, we limited our investigation to approaches that leverage task vectors for the purpose of implementing our proposed quantization scheme. Furthermore, we reimplemented all merging methods and conducted extensive experiments to ensure a fair and comprehensive evaluation. Below, we briefly summarize the key insights of each approach:

\noindent\textbf{Individual} fine-tunes the pretrained model separately for each task, which optimizes performance for the specific task. However, this approach cannot handle multiple tasks.

\noindent\textbf{Task Arithmetic}~\cite{TaskArithmetic_2023_ICLR} defines a task vector as the difference between a pretrained model and a fine-tuned model, then combines these vectors to form a multi-task model. 

\noindent\textbf{Ties Merging}~\cite{TiesMerging_2023_NeurIPS} mitigates interference during merging by addressing redundant parameters and sign conflicts among task vectors.

\noindent\textbf{MagMax}~\cite{MagMax_2024_ECCV} merges task vectors by selecting, for each parameter, the one with the largest magnitude change.

\noindent\textbf{Breadcrumbs}~\cite{Breadcrumbs_2024_ECCV} applies layer-wise filtering to remove extreme weight changes, including both large outliers and negligible values, constructing a unified multi-task model.

\noindent\textbf{Consensus TA}~\cite{Tall_Mask_2024_ICML} retains general weights that are important across multiple tasks while removing selfish weights to reduce task interference.

\noindent\textbf{LiNeS}~\cite{LiNeS_2025_ICLR} applies linear scaling to adjust layer-wise coefficients, which captures the relative importance of each task vector during the merging process.

\noindent\textbf{AdaMerging}~\cite{Adamerging_2024_ICLR} employs an unsupervised approach at test time to optimize merge coefficients, rather than setting them empirically.

\noindent\textbf{EMR-Merging}~\cite{EMRMerging_2024} constructs a unified model by electing a shared set of weights. It then applies task-specific binary masks and rescaling factors to adjust magnitudes.

\noindent\textbf{TSV-Compress}~\cite{gargiulo2025task} exploits the low-rank structure of layer-wise task matrices using singular value decomposition, enabling effective compression of task vectors.

\section{More Empirical Analyses}\label{appendix:more_analyses}

\begin{figure}[t]
    \centering
    \includegraphics[width=0.48\textwidth]{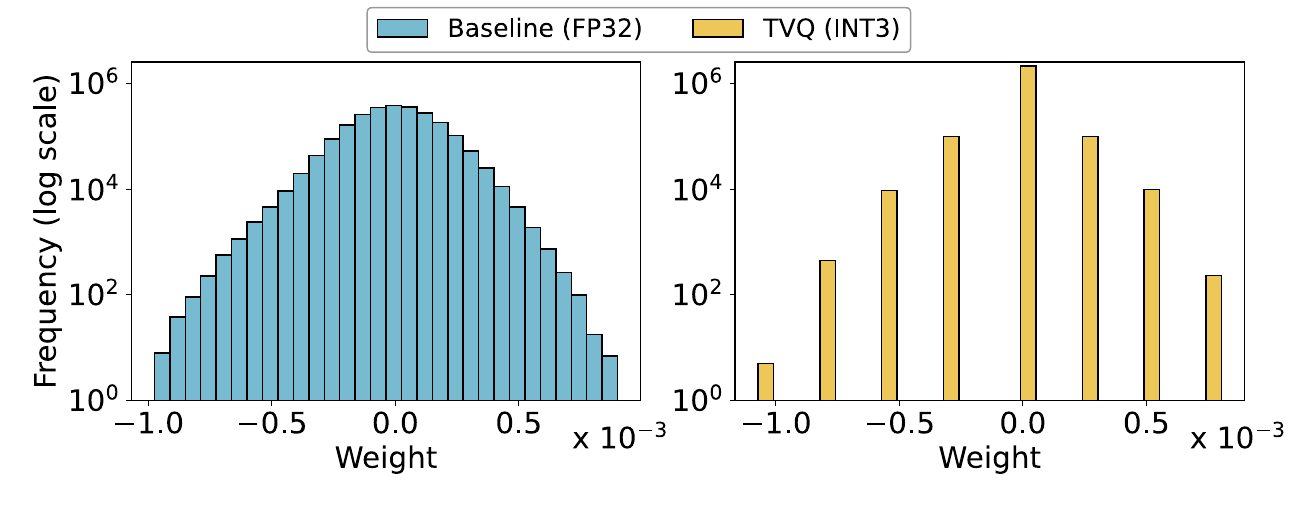}

    \caption{Histogram of task vector weight distributions before and after quantization. After quantization, smaller weight values are mapped to zero, leading to a substantial increase in sparsity.}

    \label{fig:analyses_histogram}

\end{figure}

\subsection{Impact of quantization}
\label{appendix:quantization_analyses}

In our main experiments, we observed that quantization can occasionally lead to performance improvements. Building on the analysis presented in main paper, which showed that quantization might help prevent overfitting and improve generalization, we further conducted a more extensive investigation using various approaches. In particular, we focused on 3 bit quantization because it exhibited consistent performance gains across diverse tasks.

\noindent\textbf{Weight pruning in task vector.} Quantization typically serves as an implicit regularizer, spreading out weight values across discrete levels and thereby reducing overfitting. However, we specifically quantize {task vectors}, which represent differences between pretrained and fine-tuned model weights, exhibiting a high concentration of values near zero (see Fig.~\ref{fig:method:observation} in the main paper). To better understand how quantization affects these task vectors, we visualized their weight distributions before and after quantization. As shown in Fig.~\ref{fig:analyses_histogram}, quantization maps small-magnitude weights to exactly zero, effectively pruning less impactful parameters. This process increases the proportion of zero-valued weights to 56.7\%, highlighting the pruning effect of quantization in the context of task vectors. Interestingly, this processes mirror strategies commonly employed in task vector-based model merging~\cite{TiesMerging_2023_NeurIPS, EMRMerging_2024, Tall_Mask_2024_ICML, Breadcrumbs_2024_ECCV}. Consequently, our quantization approach naturally introduces sparsity, simultaneously providing efficiency gains and potential improvements in model performance.

\begin{figure*}[t!]
\centering
    \begin{minipage}{0.49\linewidth}
        \includegraphics[width=\linewidth]{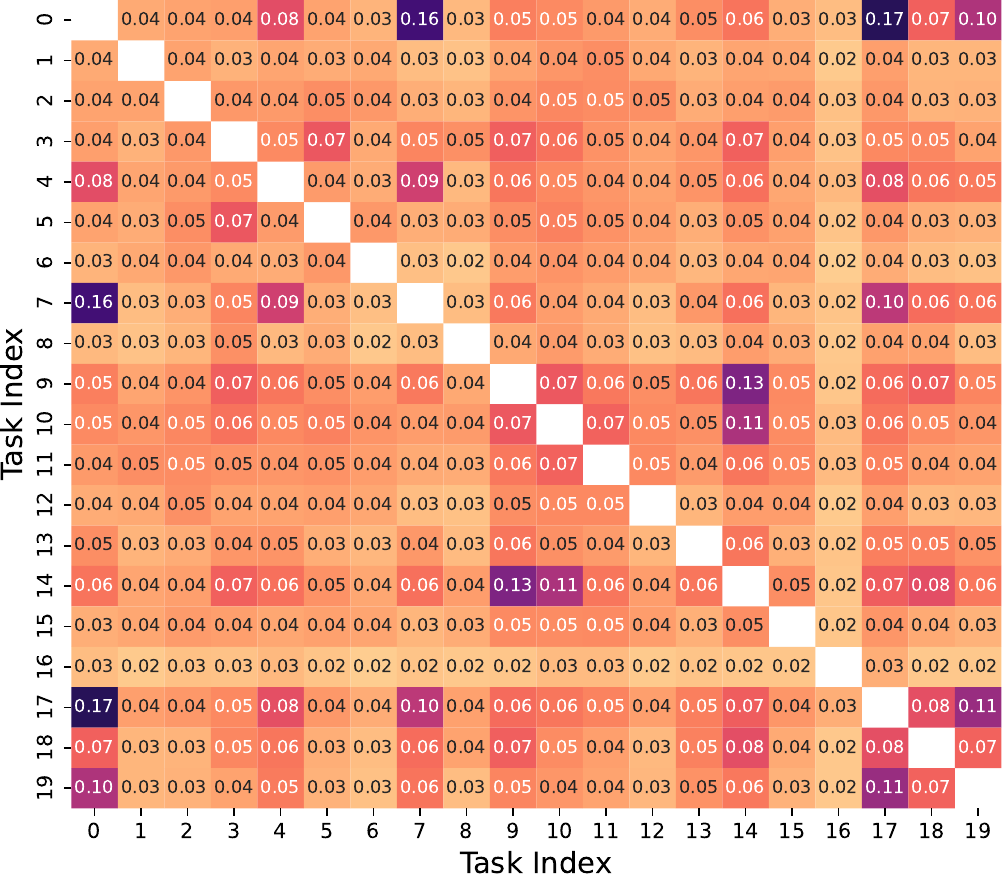}
        \subcaption[]{Original task vector (FP32)}
    \end{minipage}
    \hspace{.6em}
    \begin{minipage}{0.47\linewidth}
        \includegraphics[width=\linewidth]{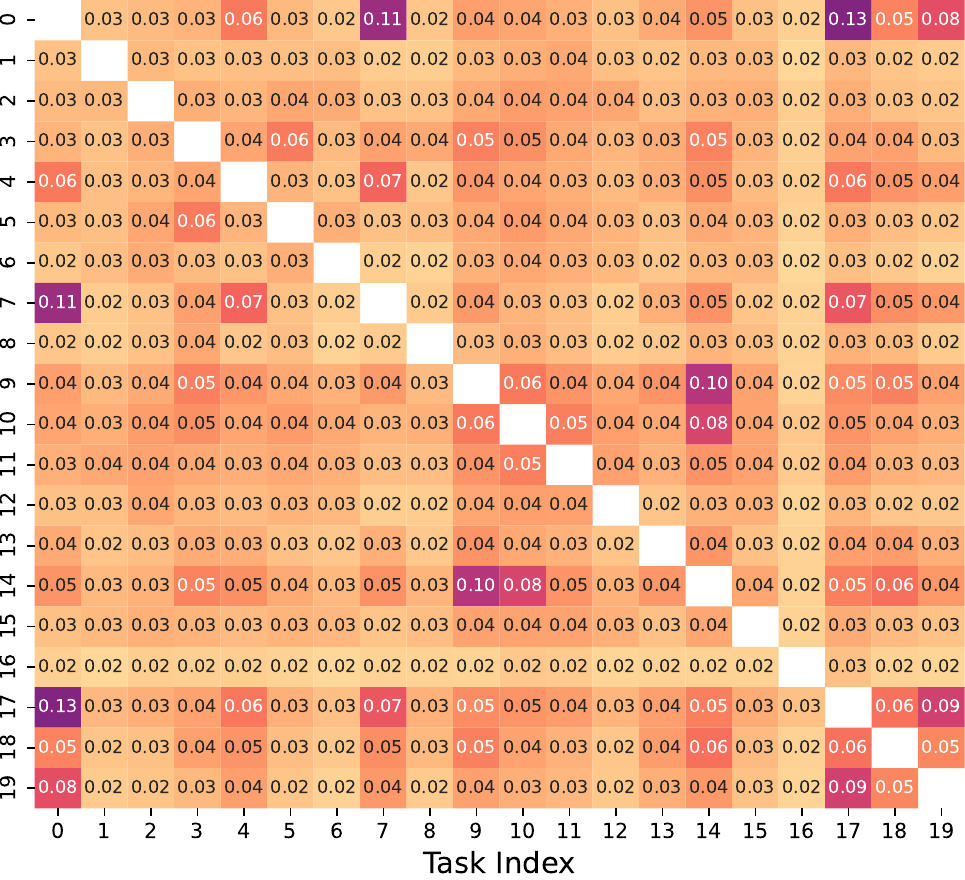}
        \subcaption[]{Quantized task vector (INT3)}
    \end{minipage}
    \caption{Confusion matrices of the cosine similarity among 20 classification task vectors for (a) full-precision (FP32) and (b) 3-bit quantized settings. The diagonal entries are excluded for a clearer comparison. {Indices 0 to 19, respectively, correspond to: MNIST, Cars, DTD, EuroSAT, GTSRB, RESISC45, SUN397, SVHN, PCAM, CIFAR100, STL10, OxfordIIITPet, Flowers102, FER2013, CIFAR10, Food101, RenderedSST2, EMNIST, FashionMNIST, and KMNIST.}}
    \label{fig:supple_similarity}
    
\end{figure*}

\noindent\textbf{Task vector similarity.} We measure cosine similarities of the task vectors before and after quantization by constructing confusion matrices for 20 classification tasks. As shown in Fig.~\ref{fig:supple_similarity}, quantization reduces off-diagonal similarities, indicating that distinct tasks become more orthogonal. Prior work~\cite{TaskArithmetic_2023_ICLR} has observed that task vectors tend to be nearly orthogonal, which facilitates more effective merging. Given that quantization further increases orthogonality, we speculate that this process helps reduce task interference and strengthens the robustness of the merged model.

\begin{table}[t!]
  \footnotesize 

  \centering
  \label{tab:bitwidth-accuracy}
  \begin{tabular}{lcccc}
    \toprule
    Offset \textbackslash{} Base & INT2 & INT3 & INT4 & INT8 \\
    \midrule
    INT2 & 69.5 & 70.2 & 69.7 & 69.6 \\
    INT3 & \textbf{70.9} & 69.9 & 69.5 & 69.5 \\
    INT4 & 69.0 & 69.0 & 69.0 & 69.0 \\
    INT8 & 68.9 & 69.0 & 69.0 & 69.0 \\
    \bottomrule
  \end{tabular}
  \caption{Average accuracy (\%) for various base and offset bit-width configurations in our Residual Task Vector Quantization (RTVQ). All experiments were performed on merging 8 classification tasks using Task Arithmetic.}
\label{tab:supple_sensitivity}
\end{table}

\subsection{Sensitivity analysis}
\label{appendix:sensitivity}

We conducted a sensitivity analysis of the quantization precision for the base and offset vectors in our Residual Task Vector Quantization (RTVQ). Table~\ref{tab:supple_sensitivity} summarizes the performance across 16 distinct bitwidth configurations ranging from {2, 3, 4, 8} bits. While one might expect allocating more bits to either the shared base vector or the task-specific offset vector would yield better performance, our results did not show a clear pattern. Instead, balanced configurations, such as combining a 2-bit base vector with a 3-bit offset vector, provided optimal accuracy. Notably, nearly all configurations surpassed the performance of 2-bit quantization (62.1\%) and approached the FP32 baseline (69.2\%). These results indicate that RTVQ remains robust across diverse bitwidth configurations, achieving near-FP32 performance even with relatively low-bit allocations.

\section{Additional Results}\label{appendix:more_results}

\subsection{Full results on 14 and 20 classification tasks}
\label{appendix:scaling_vision}

In Table~\ref{tab:supple_vitb32_14_tasks} and Table~\ref{tab:supple_vitb32_20_tasks}, we present comprehensive results corresponding to Figure~\ref{fig:more_vision} in the main paper, which shows that our quantization method becomes more stable as model size increases. Similarly, stability improves as the number of tasks grows from 14 to 20, particularly for 3-bit TVQ and RTVQ. Additionally, while full-precision models face increasing storage overhead as tasks scale, our quantization method keeps storage requirements manageable. This ensures efficiency gains without compromising accuracy, making it well-suited for large-scale or multi-task scenarios where storage scalability is crucial.

\subsection{Additional results for dense prediction tasks}
\label{appendix:supple_dense}
\begin{figure*}[!t]
    \centering
    \includegraphics[width=0.85\textwidth,keepaspectratio]{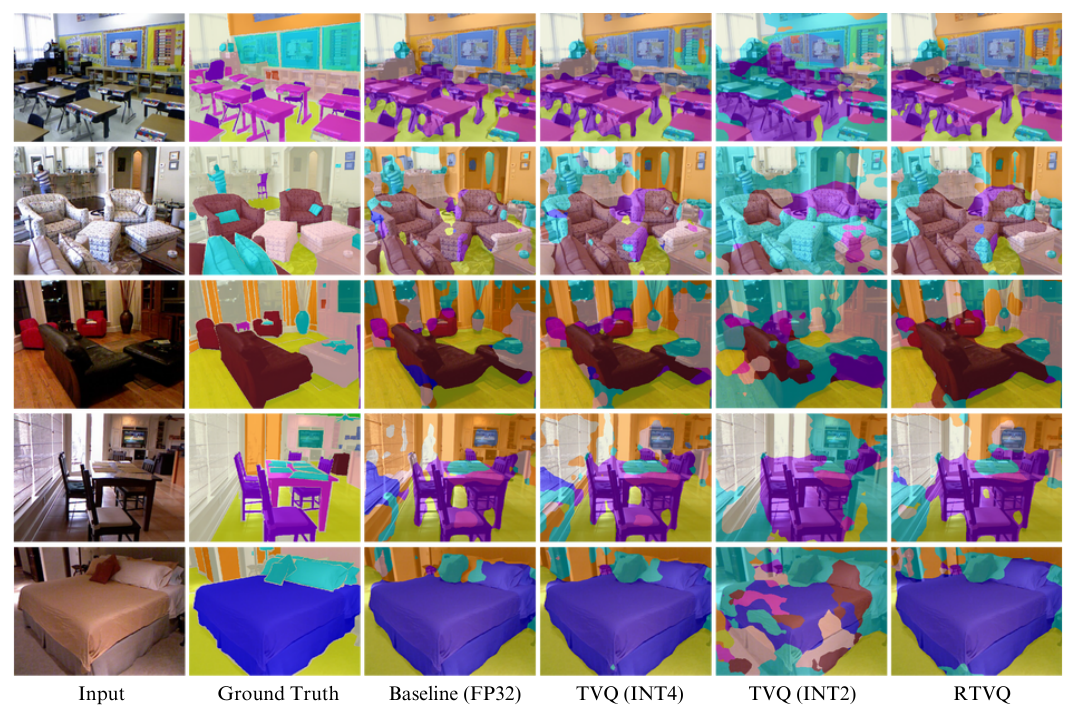}
        \caption{Qualitative results of segmentation task across different quantization methods. RTVQ quantizes both the base vector and offset to 2 bits.}
    \label{fig:supple_qualitative_seg}
\end{figure*}

\begin{figure*}[!t]
    \centering
    \includegraphics[width=0.85\textwidth,keepaspectratio]{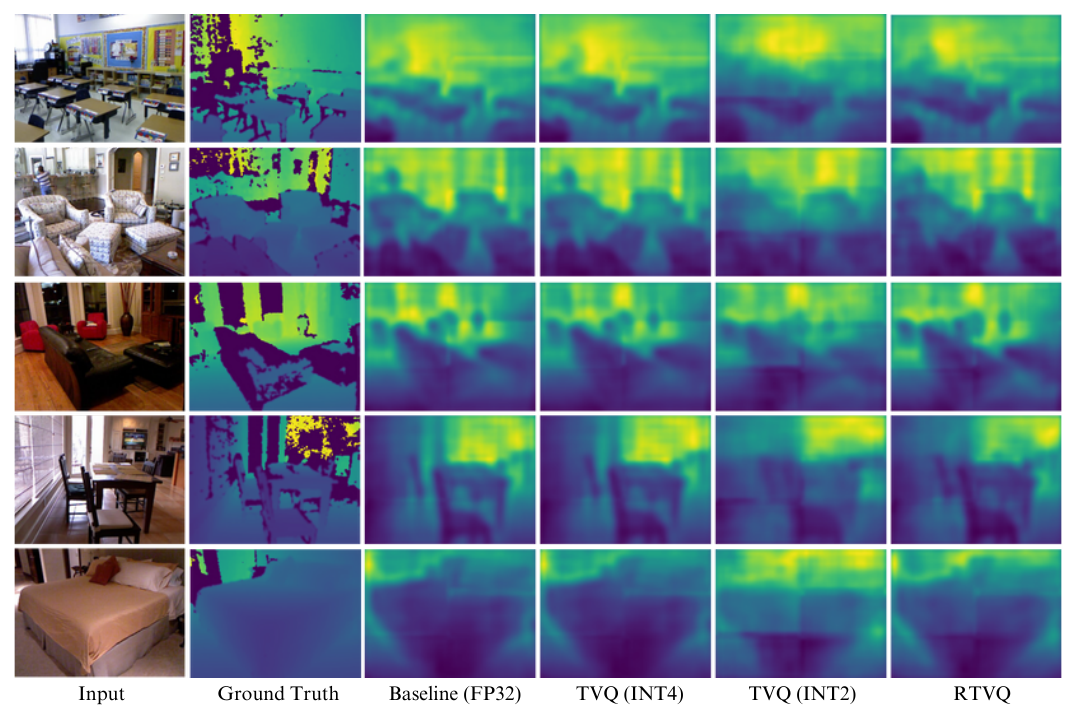}
        \caption{Qualitative results of depth estimation task across different quantization methods. RTVQ quantizes both the base vector and offset to 2 bits.}
    \label{fig:supple_qualitative_depth}
\end{figure*}

\begin{figure*}[!t]
    \centering
    \includegraphics[width=0.85\textwidth,keepaspectratio]{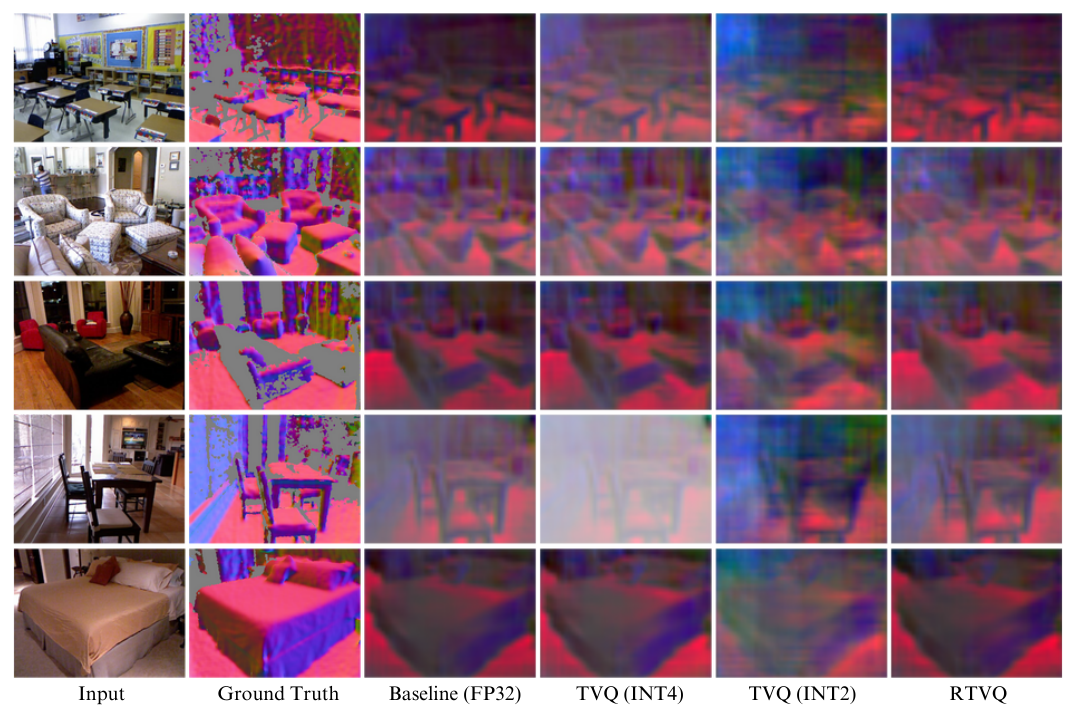}
        \caption{Qualitative results of normal estimation task across different quantization methods. RTVQ quantizes both the base vector and offset to 2 bits.}
    \label{fig:supple_qualitative_normal}
\end{figure*}

We provide detailed results corresponding to Table~\ref{tab:main_dense_pred_tasks} in the main paper for merging dense prediction tasks. Table~\ref{tab:supple_dense} expands on these results by comparing a broader range of quantization methods and evaluation metrics. Additionally, Fig.~\ref{fig:supple_qualitative_seg}, ~\ref{fig:supple_qualitative_depth}, and ~\ref{fig:supple_qualitative_normal} present qualitative visualizations for semantic segmentation, depth estimation, and normal estimation using the state-of-the-art EMR-Merging method. These results further validate the trends consistently observed throughout the paper.

\subsection{Comprehensive task-level results}
\label{appendix:task-level}

We report the average accuracy across all tasks in Table~\ref{tab:main_vitb32_8_tasks} and Table~\ref{tab:main_vitl14_8_tasks} of the main paper. This provides an overall comparison of different quantization methods and shows how well each method generalizes across multiple tasks.
We also present detailed task-specific results for ViT-B/32 and ViT-L/14 in Table~\ref{tab:supple_vitb32_tasklevel} and ~\ref{tab:supple_vitl14_tasklevel}. These results enable closer inspection of per-task behavior and reveal where certain methods perform well or poorly. Such analysis is important for understanding how sensitive quantization strategies are to different task characteristics.

\subsection{Loss landscape visualizations}
\label{appendix:loss_landscape}
\begin{figure*}[t]
    \centering
    \renewcommand{\thesubfigure}{} %
    \setlength{\tabcolsep}{1pt} %
    \renewcommand{\arraystretch}{0.8} %
    \begin{tabular}{cccc} %
    
        \subfloat{\includegraphics[width=0.23\linewidth]{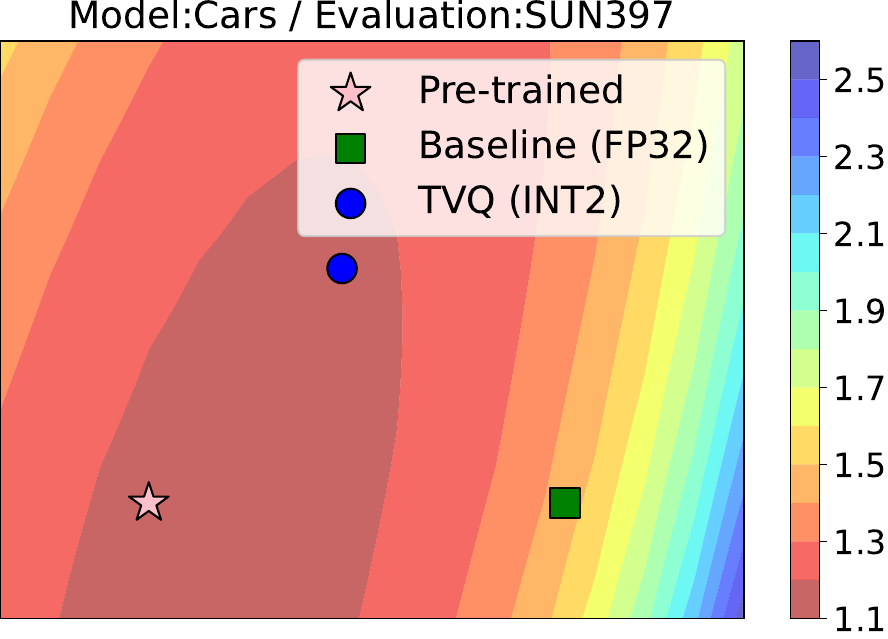}} &
        \subfloat{\includegraphics[width=0.23\linewidth]{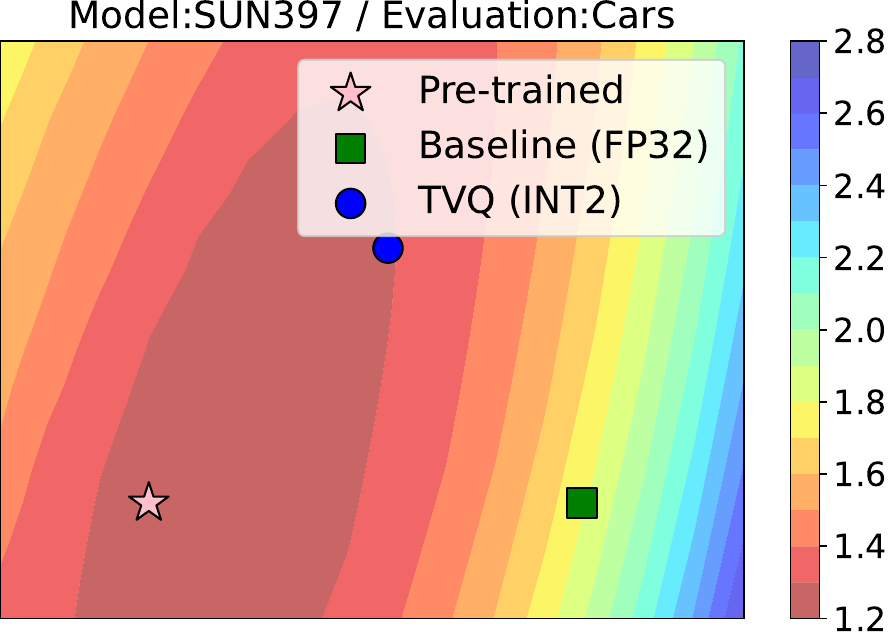}} &
        \subfloat{\includegraphics[width=0.23\linewidth]{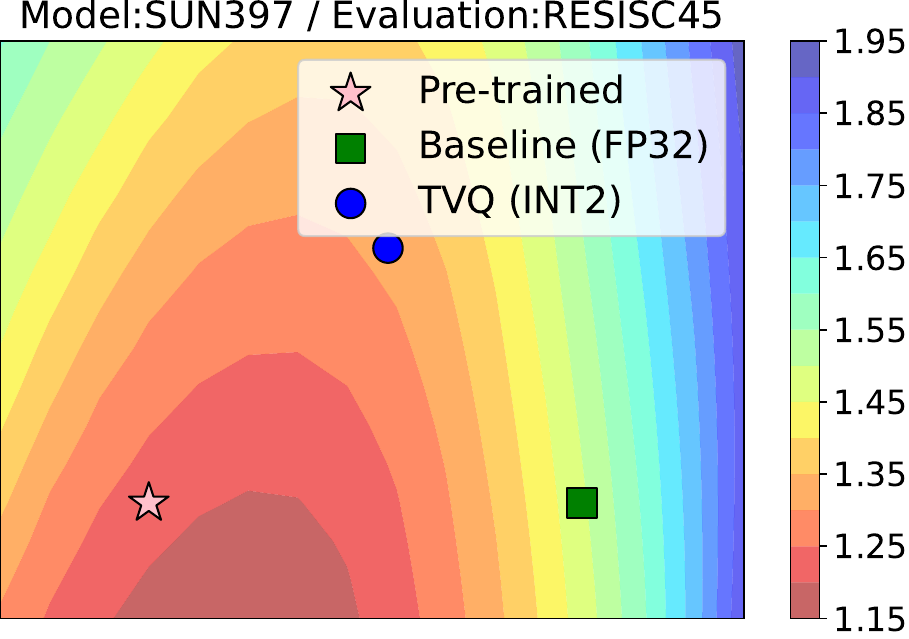}} &
        \subfloat{\includegraphics[width=0.23\linewidth]{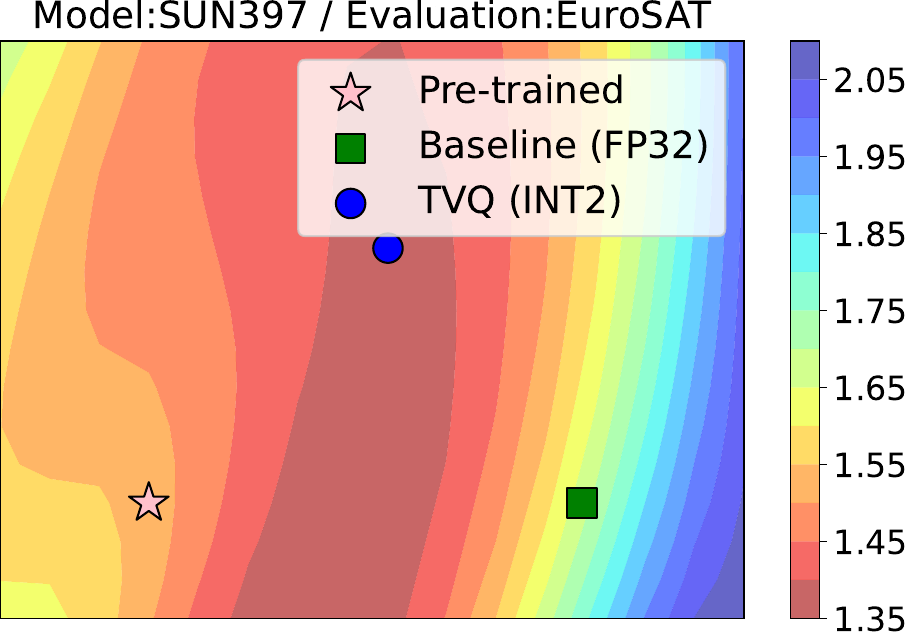}} \\
        
        \subfloat{\includegraphics[width=0.23\linewidth]{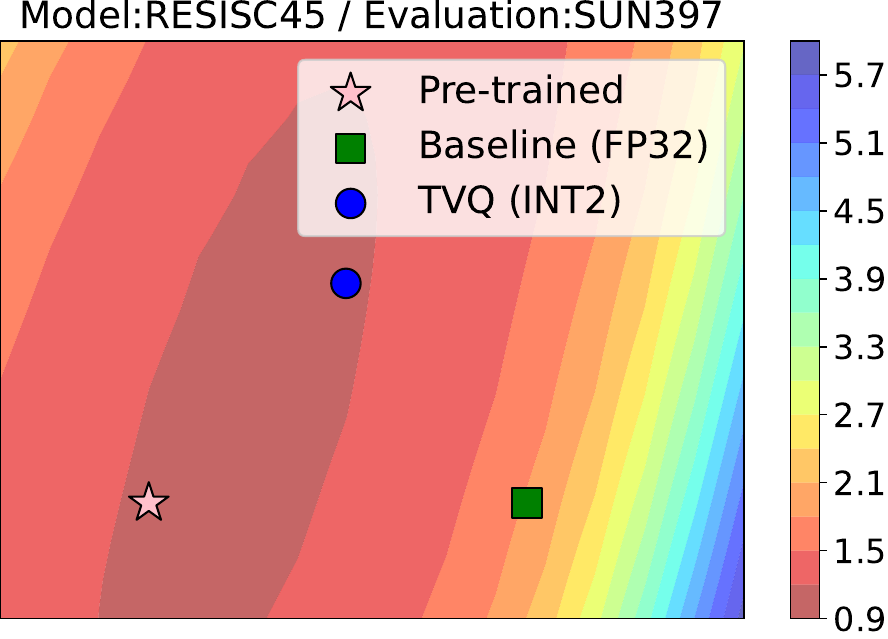}} &
        \subfloat{\includegraphics[width=0.23\linewidth]{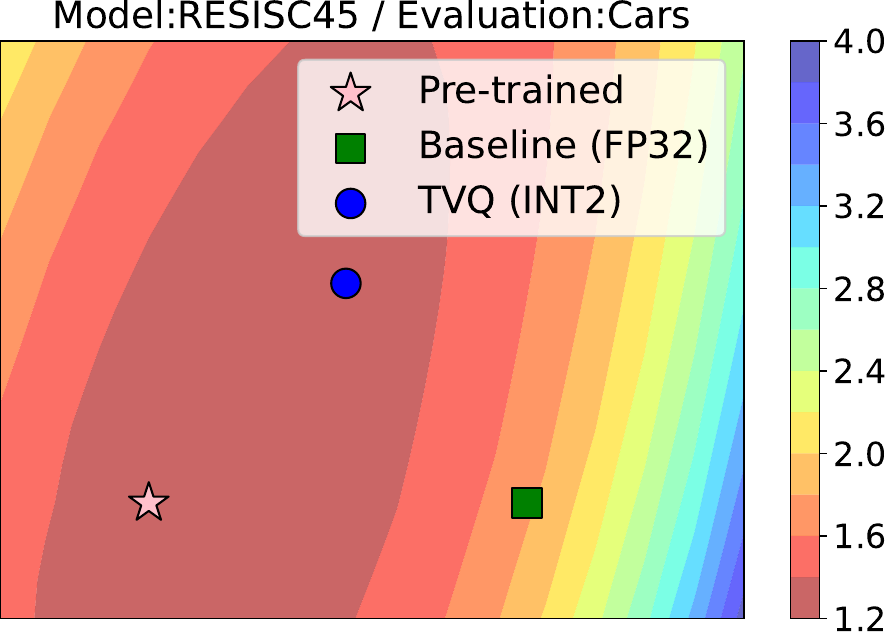}} &
        \subfloat{\includegraphics[width=0.23\linewidth]{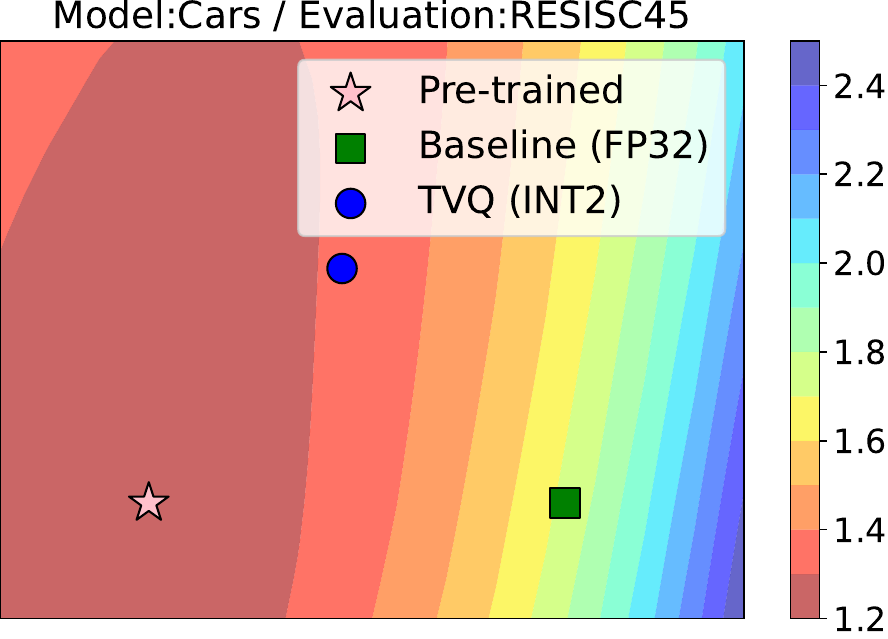}} &
        \subfloat{\includegraphics[width=0.23\linewidth]{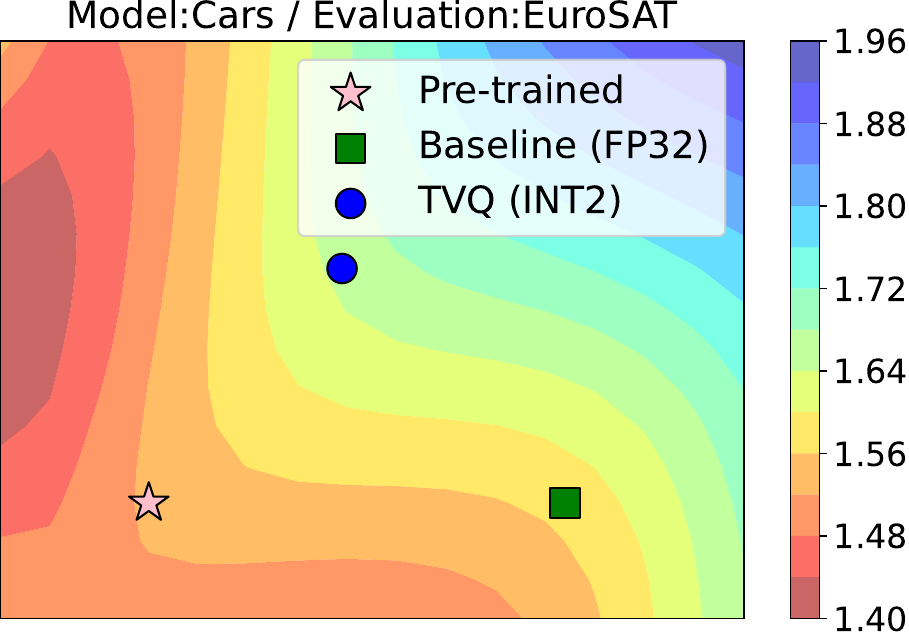}} \\
        
        \subfloat{\includegraphics[width=0.23\linewidth]{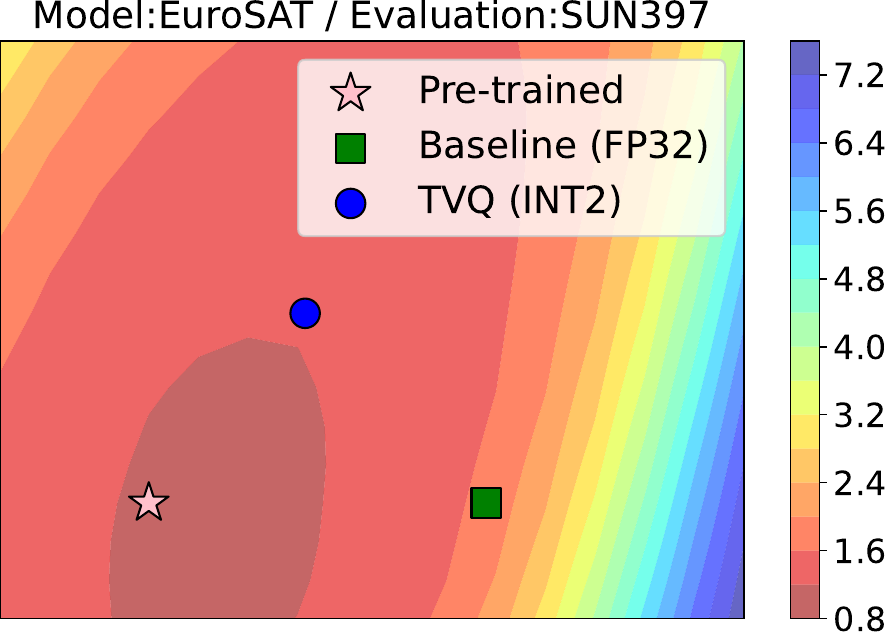}} &
        \subfloat{\includegraphics[width=0.23\linewidth]{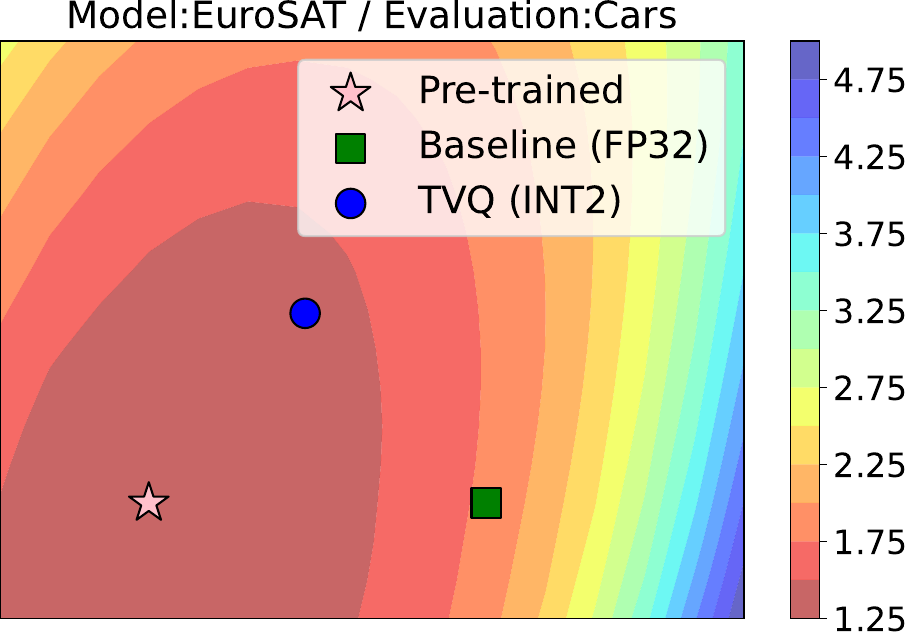}} &
        \subfloat{\includegraphics[width=0.23\linewidth]{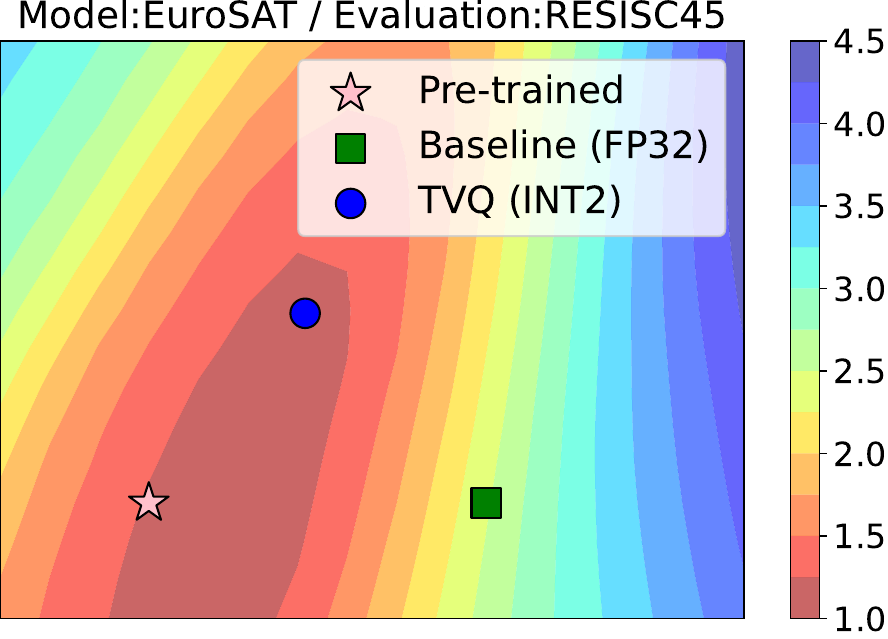}} &
        \subfloat{\includegraphics[width=0.23\linewidth]{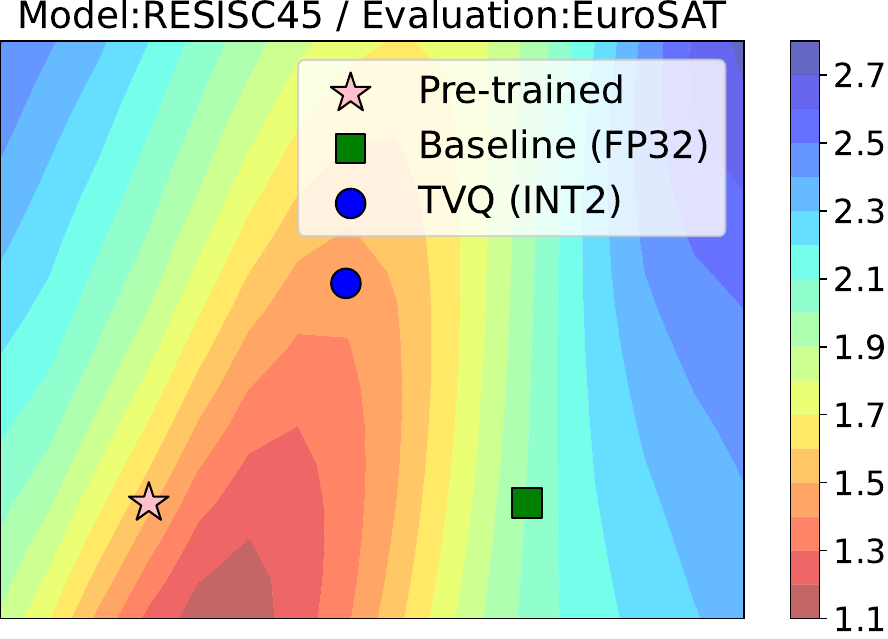}} \\
        
        \subfloat{\includegraphics[width=0.23\linewidth]{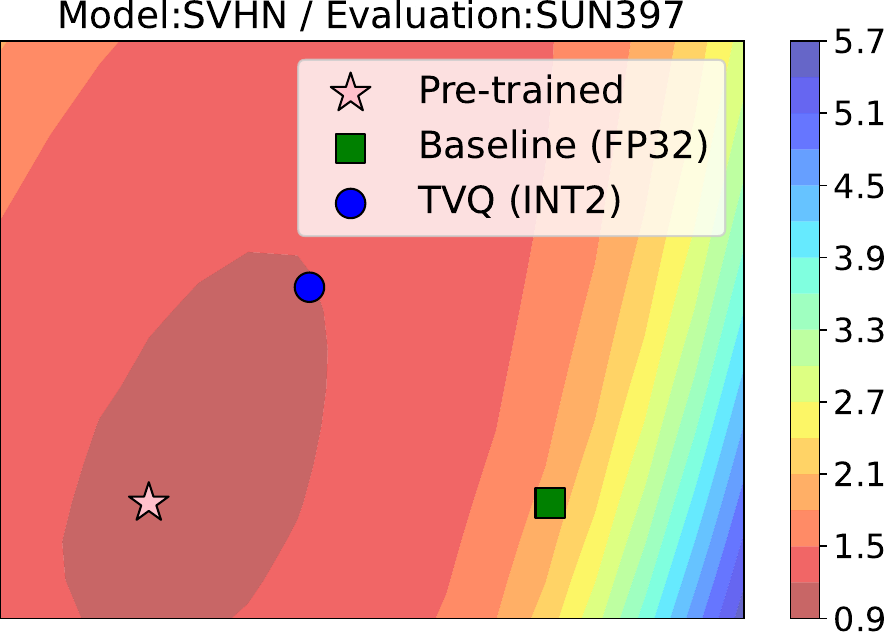}} &
        \subfloat{\includegraphics[width=0.23\linewidth]{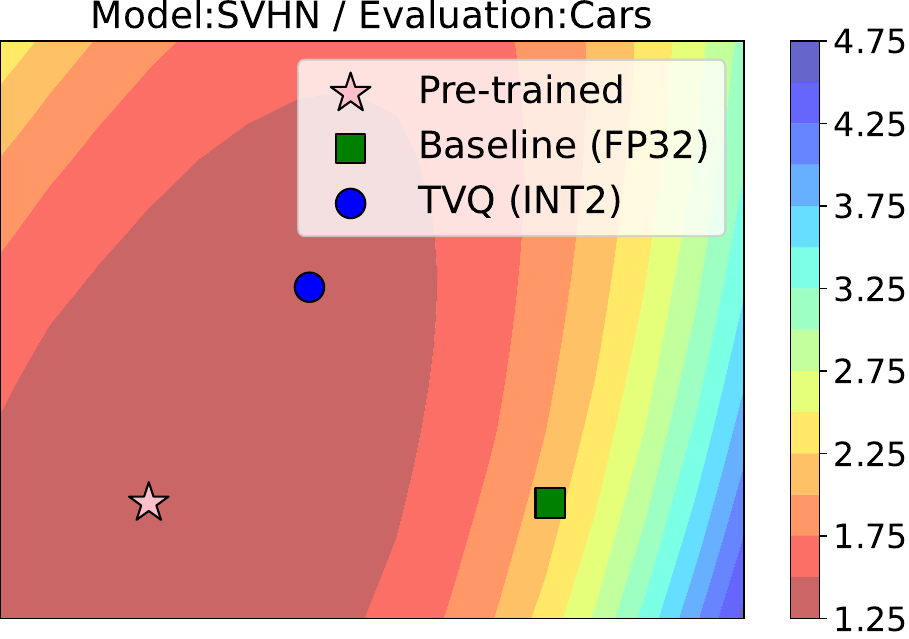}} &
        \subfloat{\includegraphics[width=0.23\linewidth]{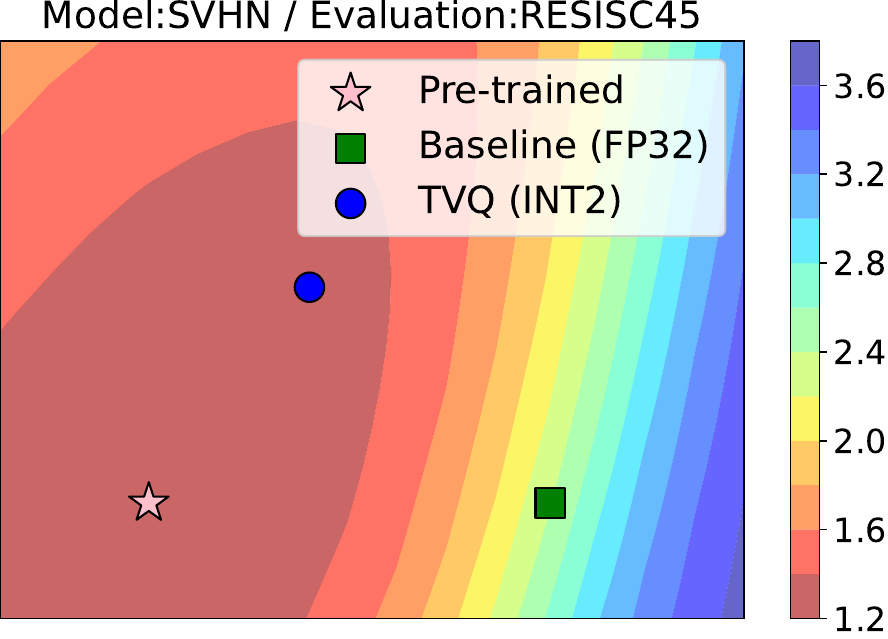}} &
        \subfloat{\includegraphics[width=0.23\linewidth]{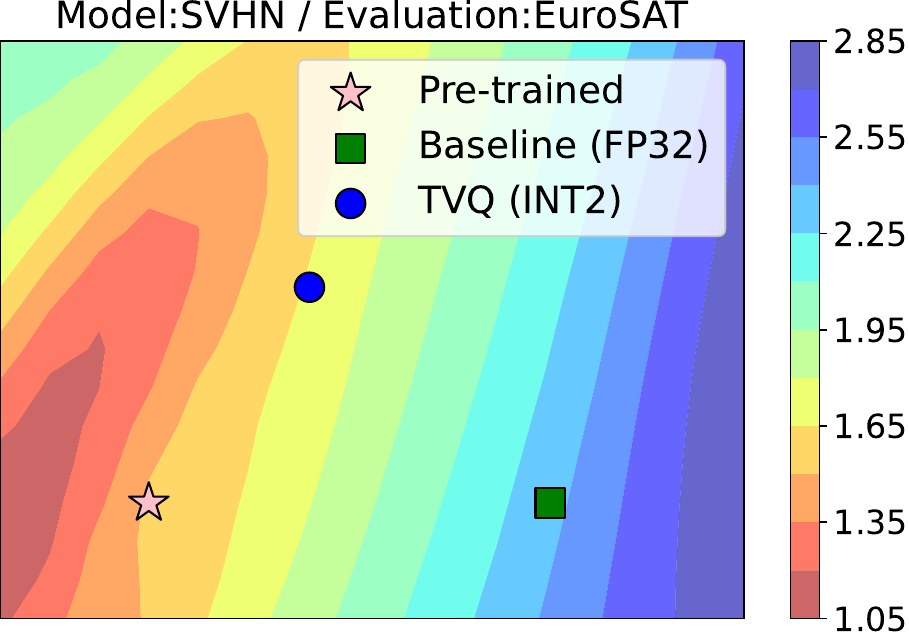}} \\
        
        \subfloat{\includegraphics[width=0.23\linewidth]{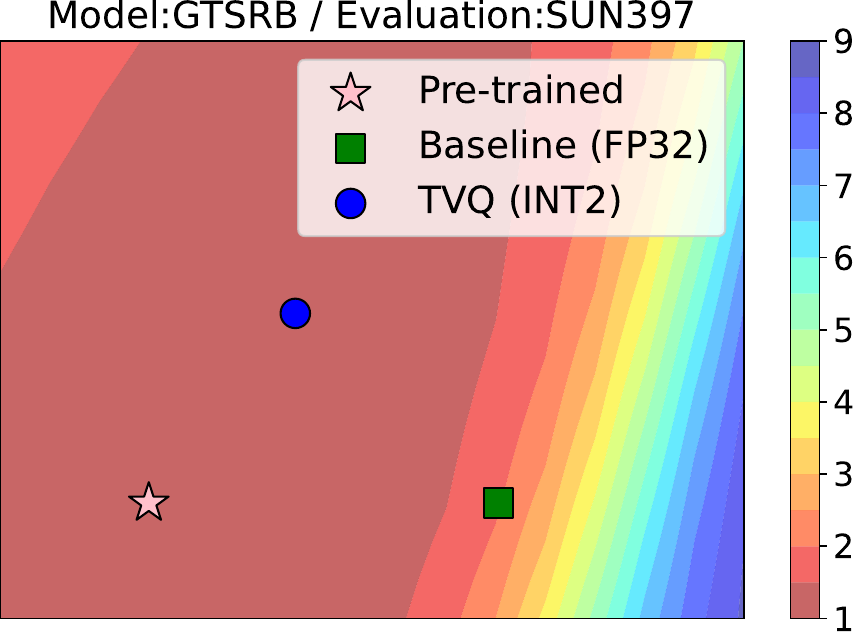}} &
        \subfloat{\includegraphics[width=0.23\linewidth]{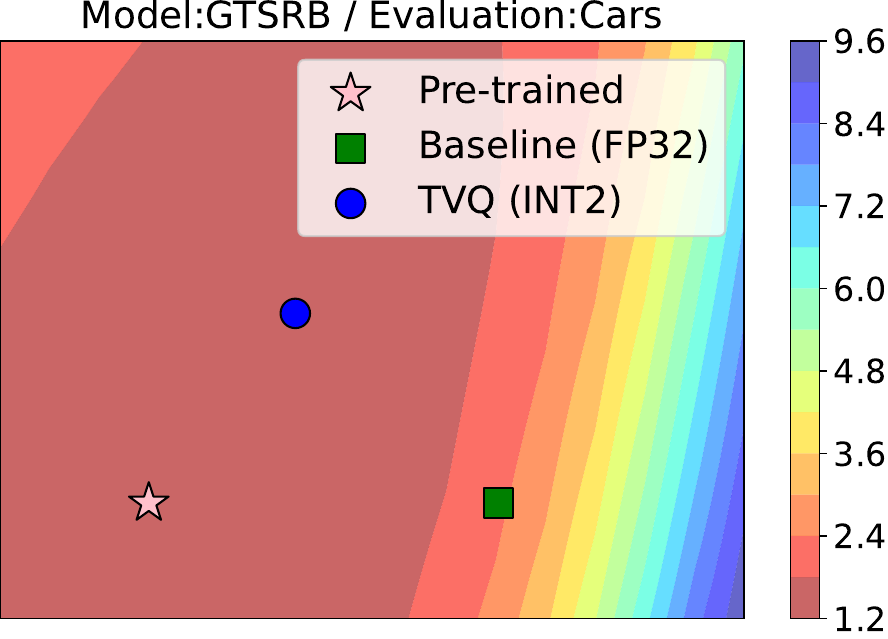}} &
        \subfloat{\includegraphics[width=0.23\linewidth]{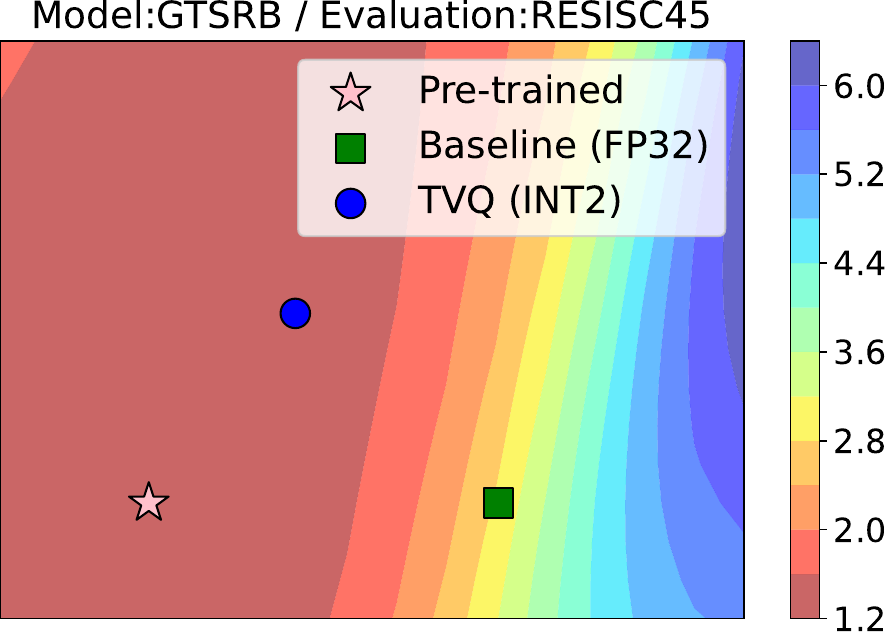}} &
        \subfloat{\includegraphics[width=0.23\linewidth]{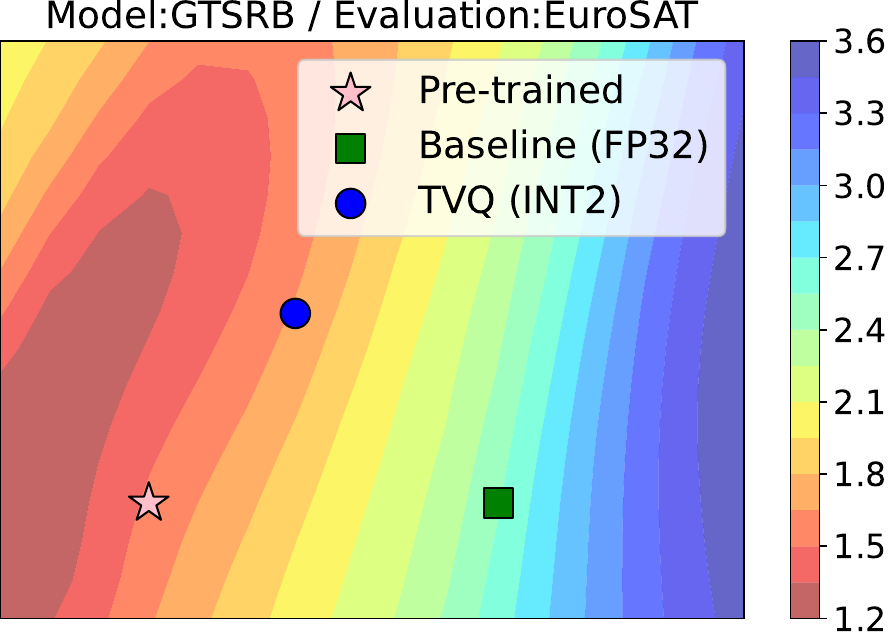}} \\
        
        \subfloat{\includegraphics[width=0.23\linewidth]{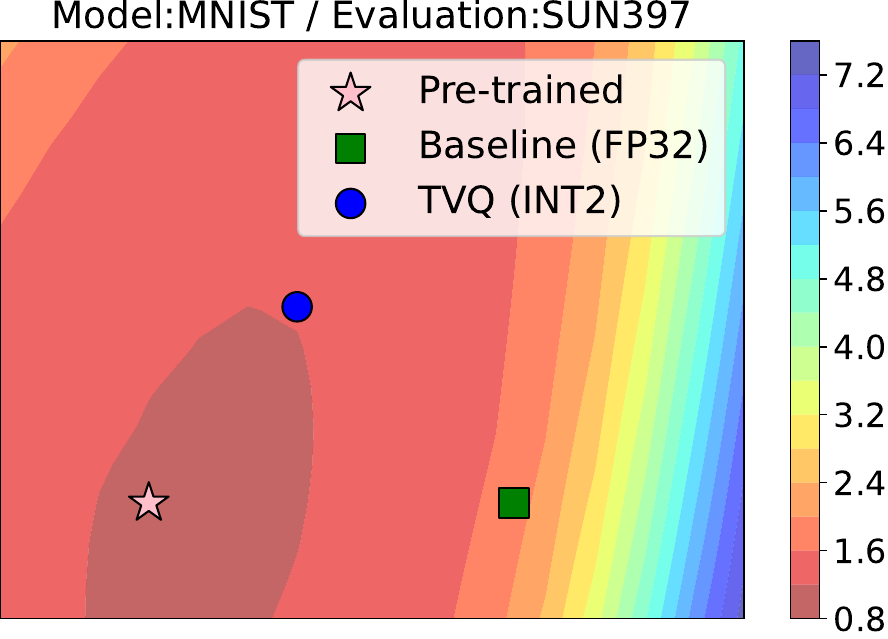}} &
        \subfloat{\includegraphics[width=0.23\linewidth]{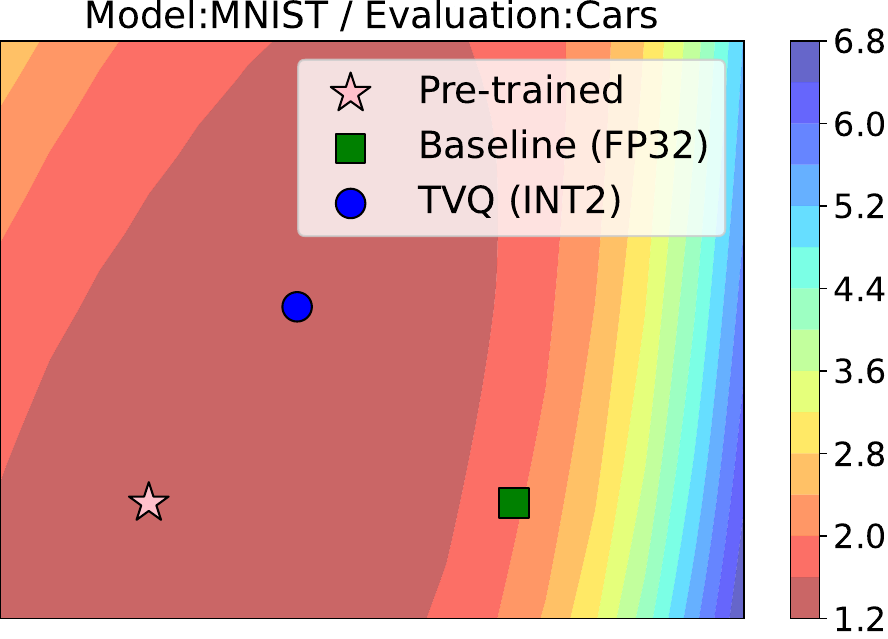}} &
        \subfloat{\includegraphics[width=0.23\linewidth]{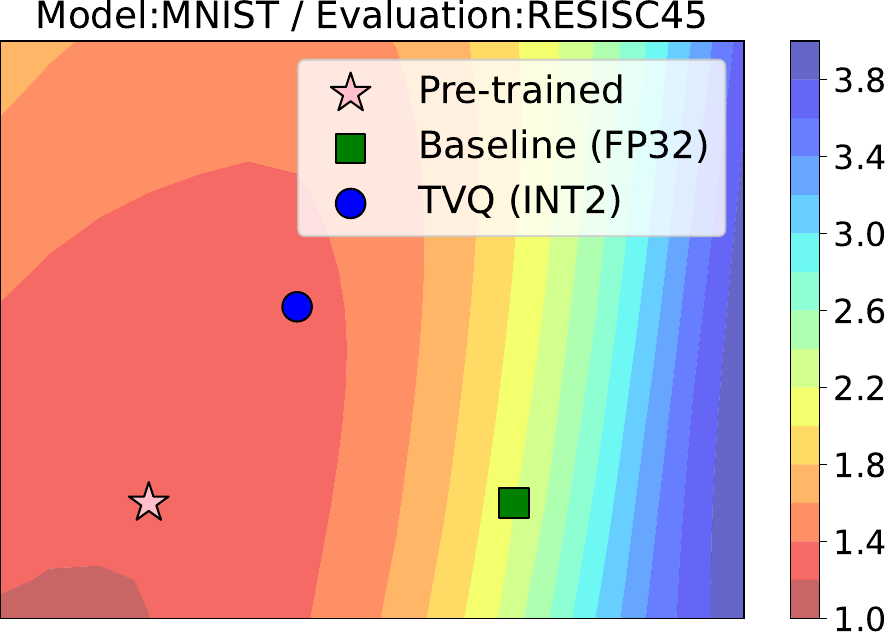}} &
        \subfloat{\includegraphics[width=0.23\linewidth]{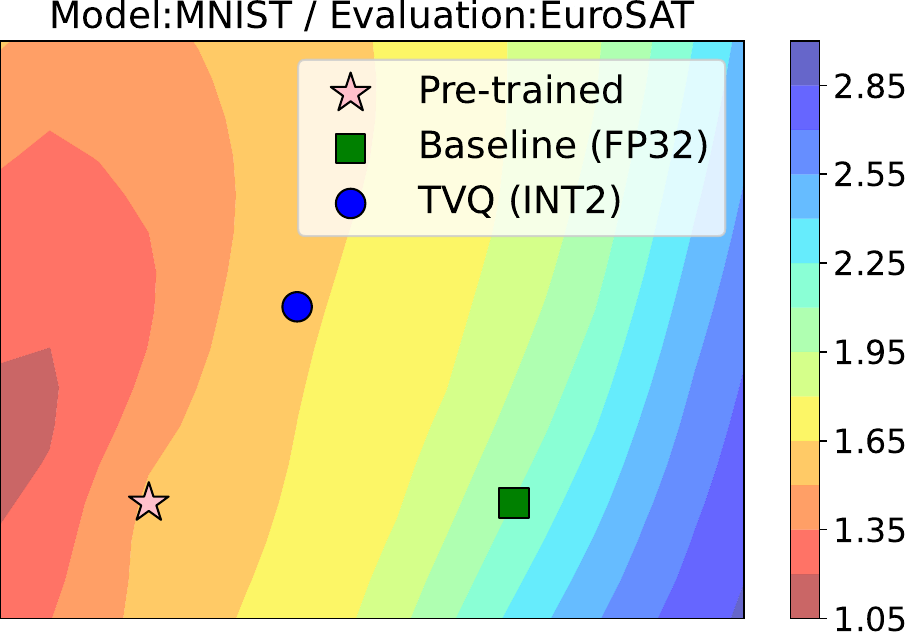}} \\
        
        \subfloat{\includegraphics[width=0.23\linewidth]{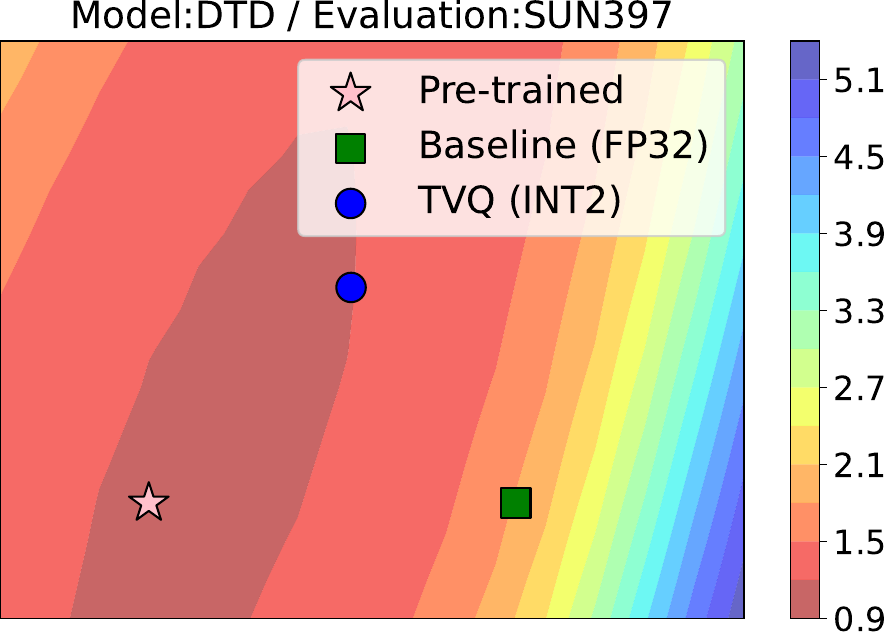}} &
        \subfloat{\includegraphics[width=0.23\linewidth]{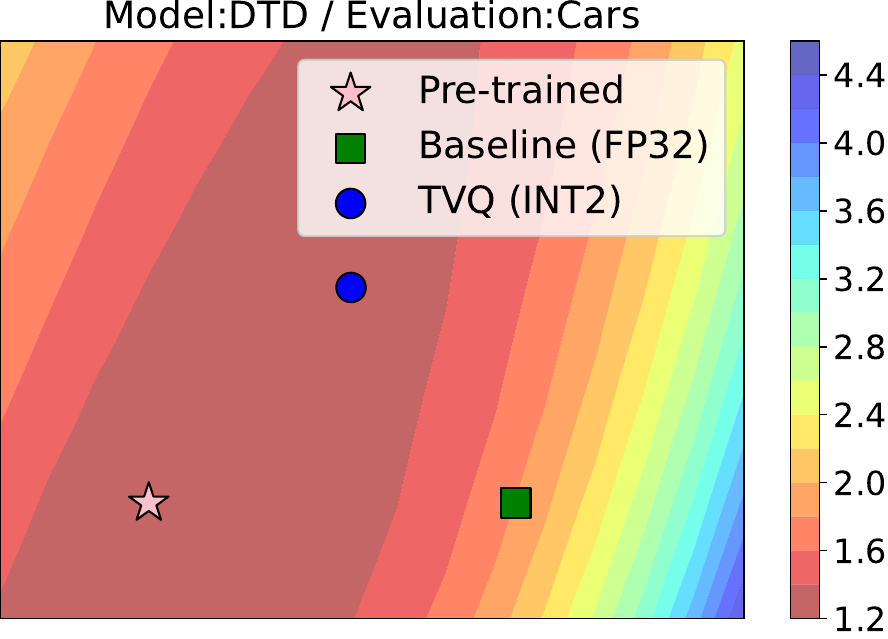}} &
        \subfloat{\includegraphics[width=0.23\linewidth]{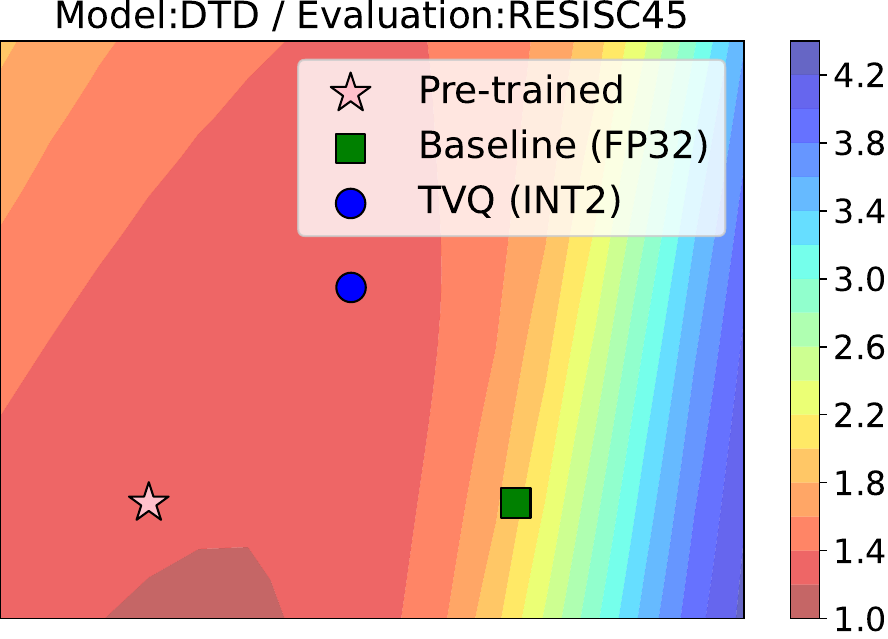}} &
        \subfloat{\includegraphics[width=0.23\linewidth]{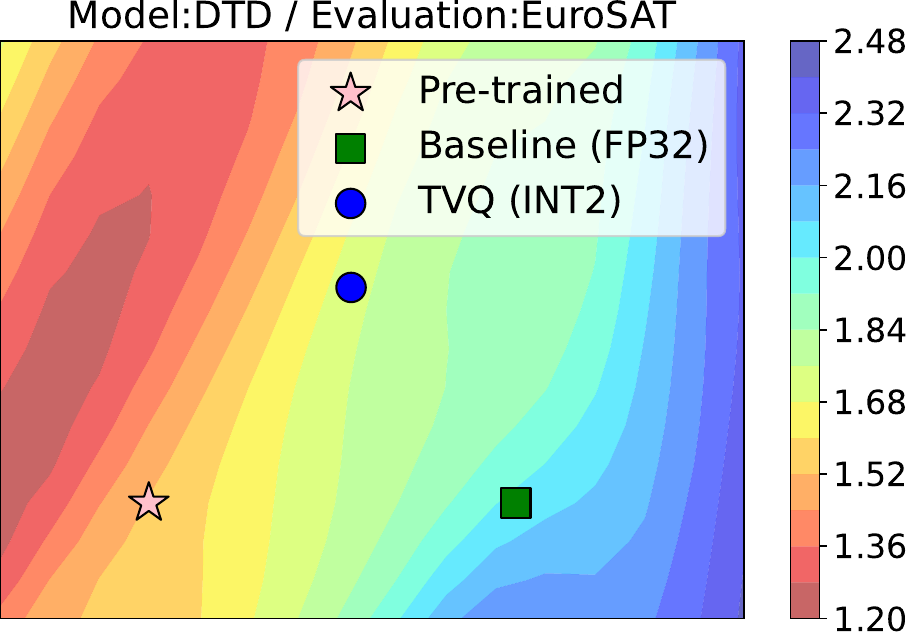}} \\
    
    \end{tabular}
    \caption{Loss landscape visualization of cross task pairs for 2-bit TVQ. The results show evaluations on SUN397, Cars, RESISC45, and EuroSAT.}
    \label{fig:tvq_landscape_control1}
\end{figure*}

\begin{figure*}[t]
    \centering
    \renewcommand{\thesubfigure}{} %
    \setlength{\tabcolsep}{1pt} %
    \renewcommand{\arraystretch}{0.8} %
    \begin{tabular}{cccc} %
    
        \subfloat{\includegraphics[width=0.23\linewidth]{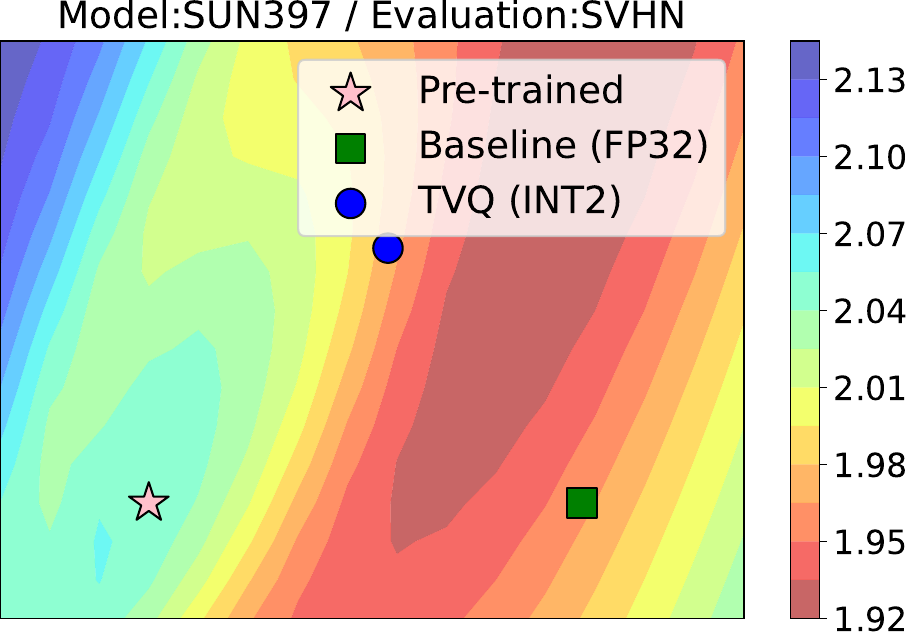}} &
        \subfloat{\includegraphics[width=0.23\linewidth]{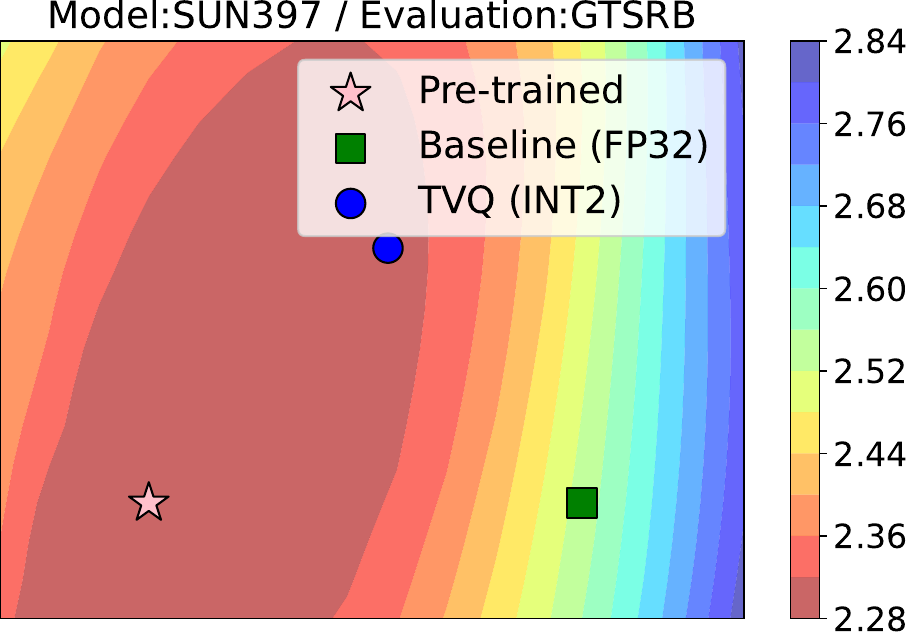}} &
        \subfloat{\includegraphics[width=0.23\linewidth]{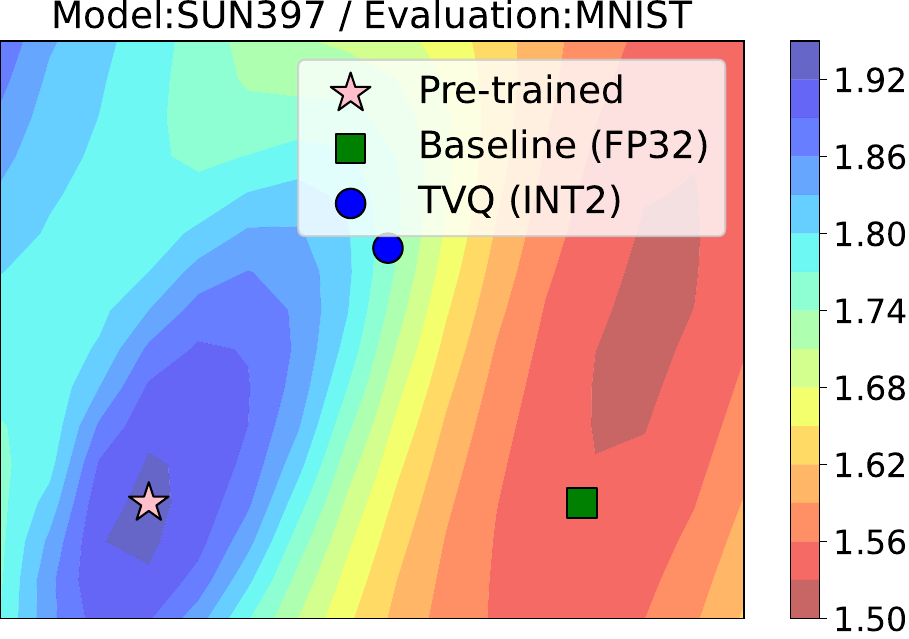}} &
        \subfloat{\includegraphics[width=0.23\linewidth]{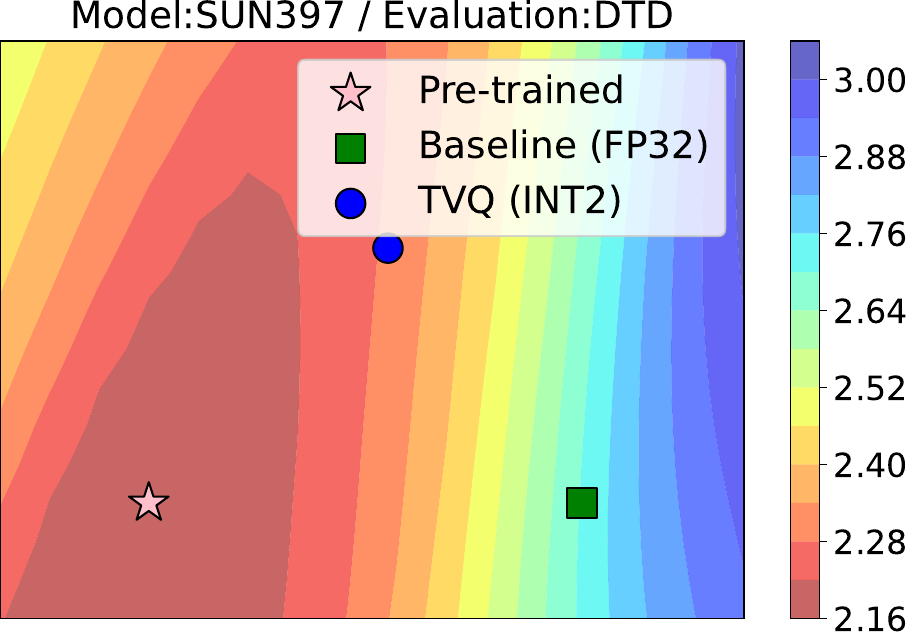}} \\
        
        \subfloat{\includegraphics[width=0.23\linewidth]{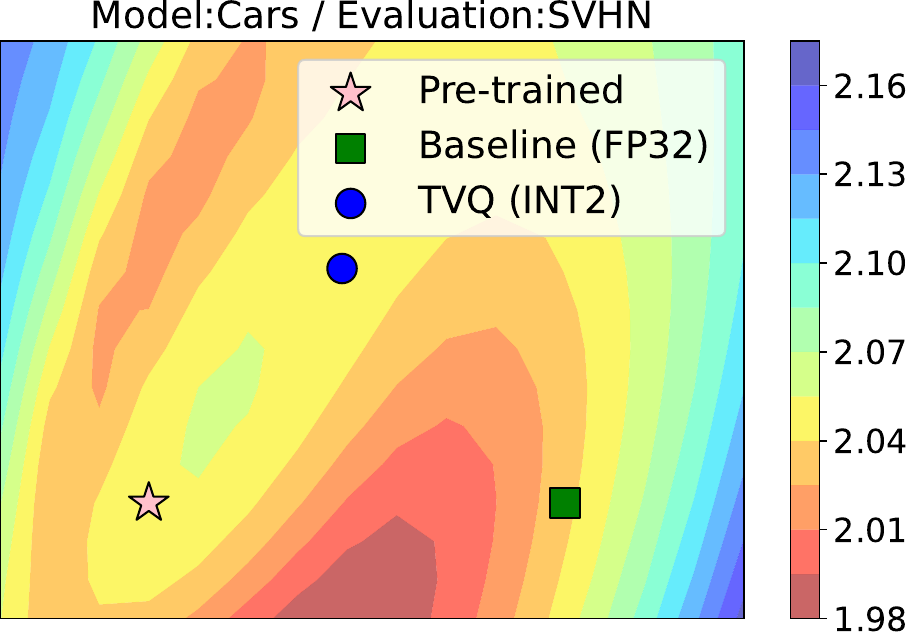}} &
        \subfloat{\includegraphics[width=0.23\linewidth]{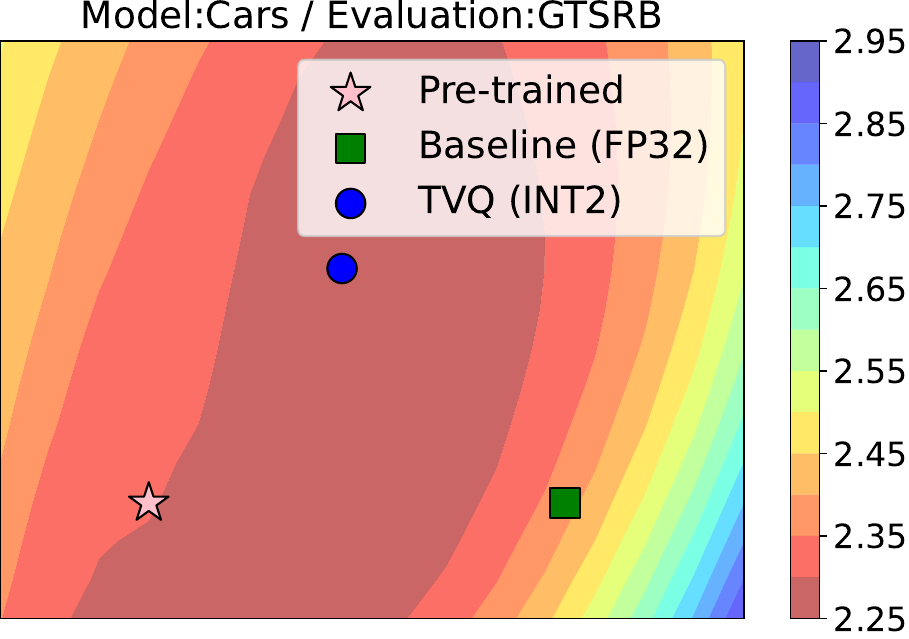}} &
        \subfloat{\includegraphics[width=0.23\linewidth]{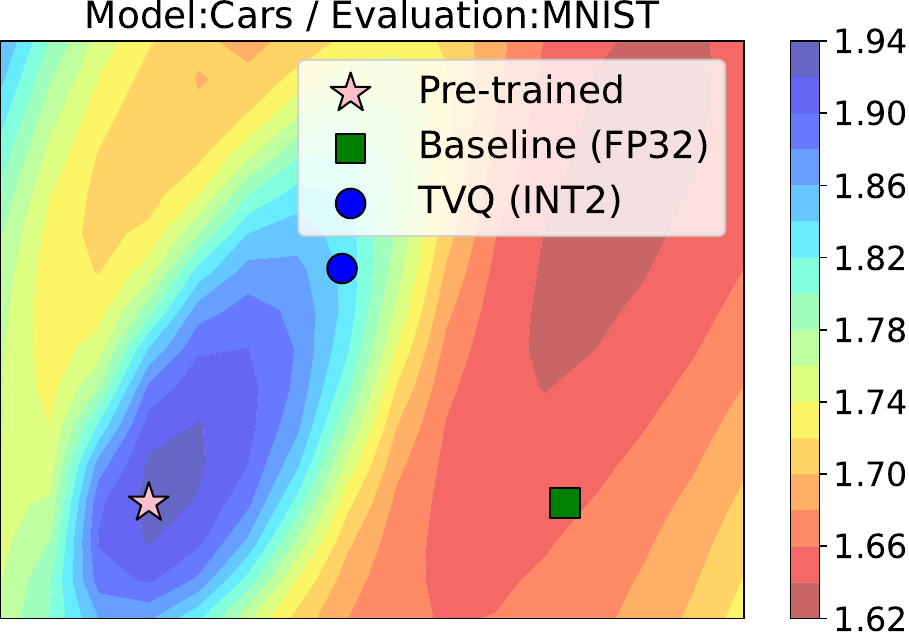}} &
        \subfloat{\includegraphics[width=0.23\linewidth]{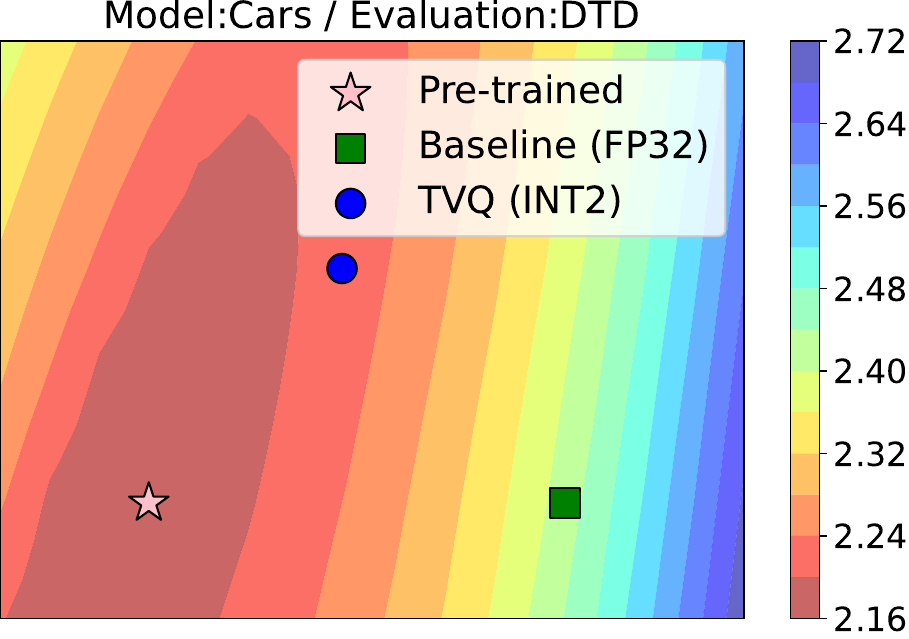}} \\
        
        \subfloat{\includegraphics[width=0.23\linewidth]{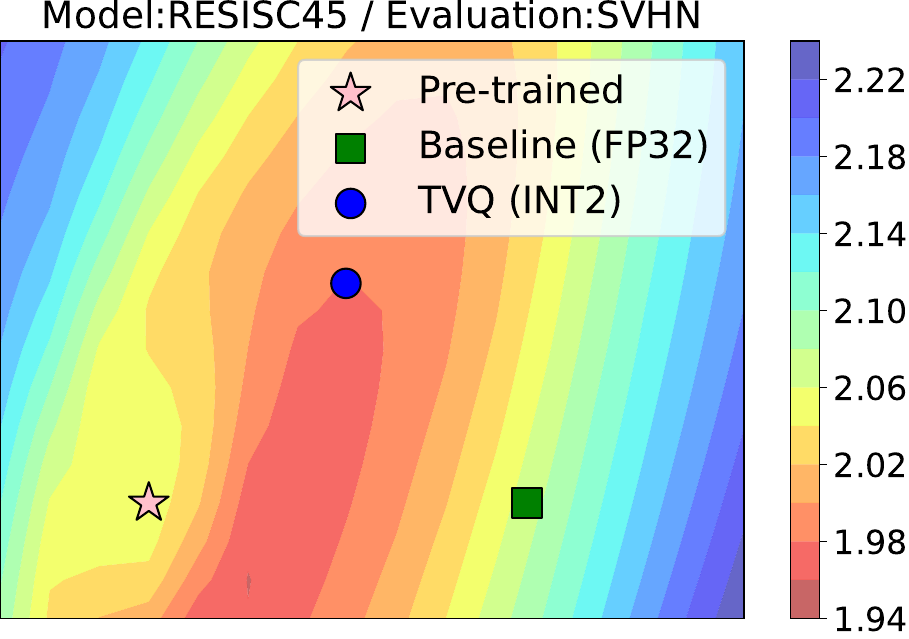}} &
        \subfloat{\includegraphics[width=0.23\linewidth]{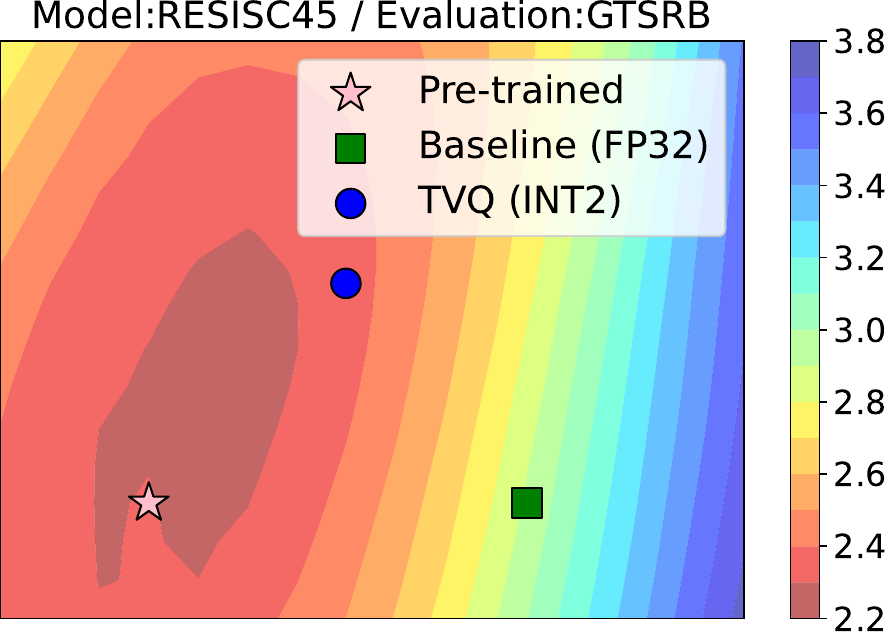}} &
        \subfloat{\includegraphics[width=0.23\linewidth]{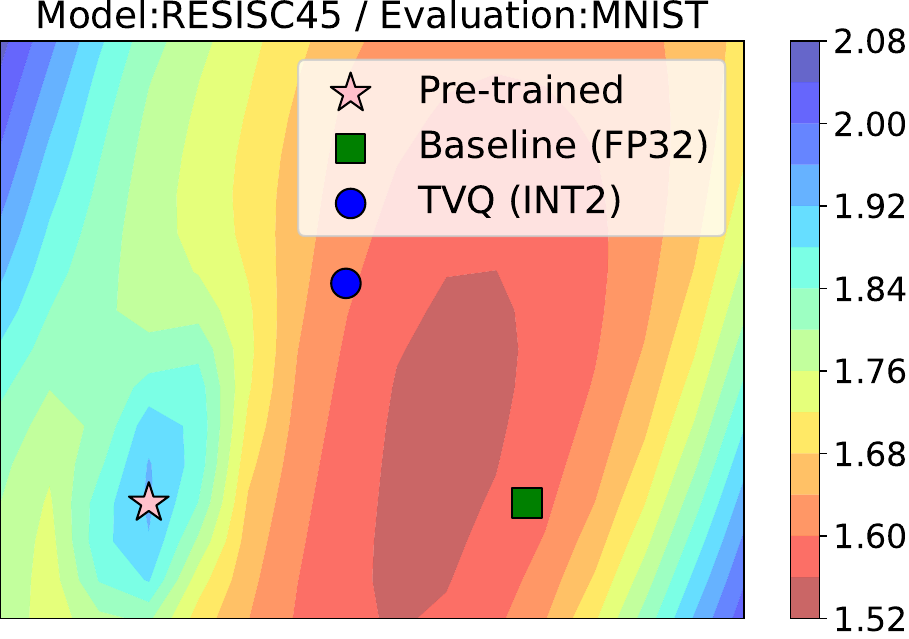}} &
        \subfloat{\includegraphics[width=0.23\linewidth]{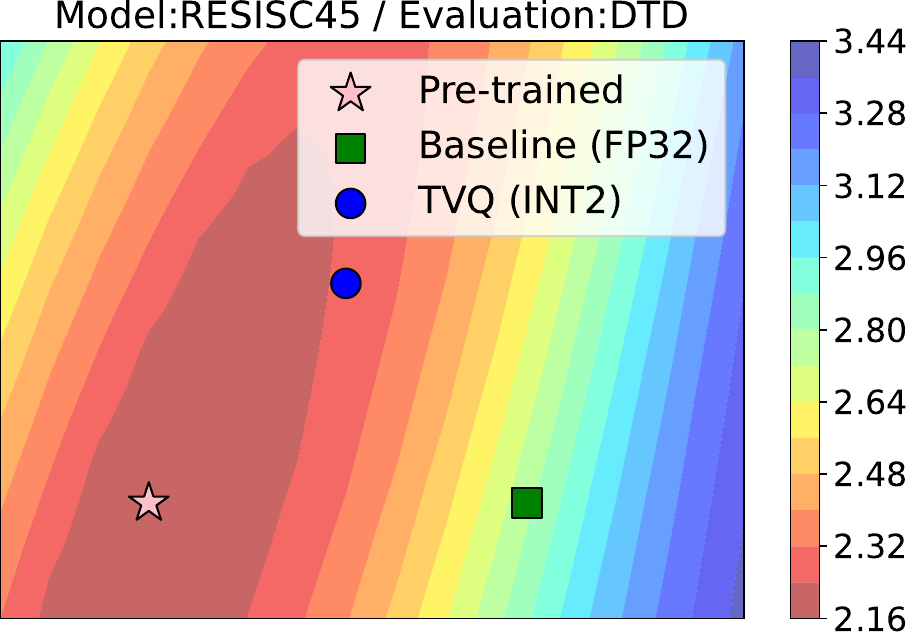}} \\
        
        \subfloat{\includegraphics[width=0.23\linewidth]{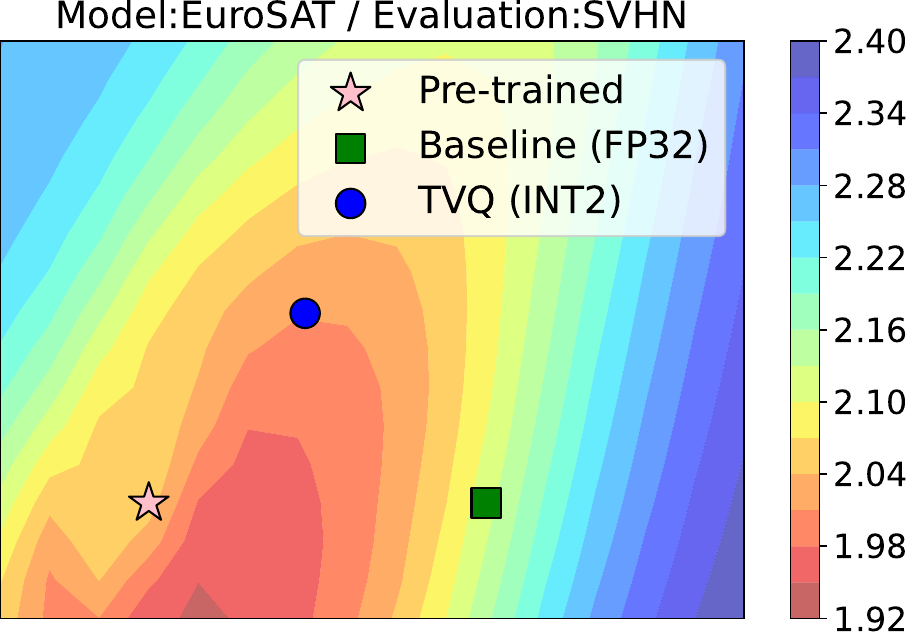}} &
        \subfloat{\includegraphics[width=0.23\linewidth]{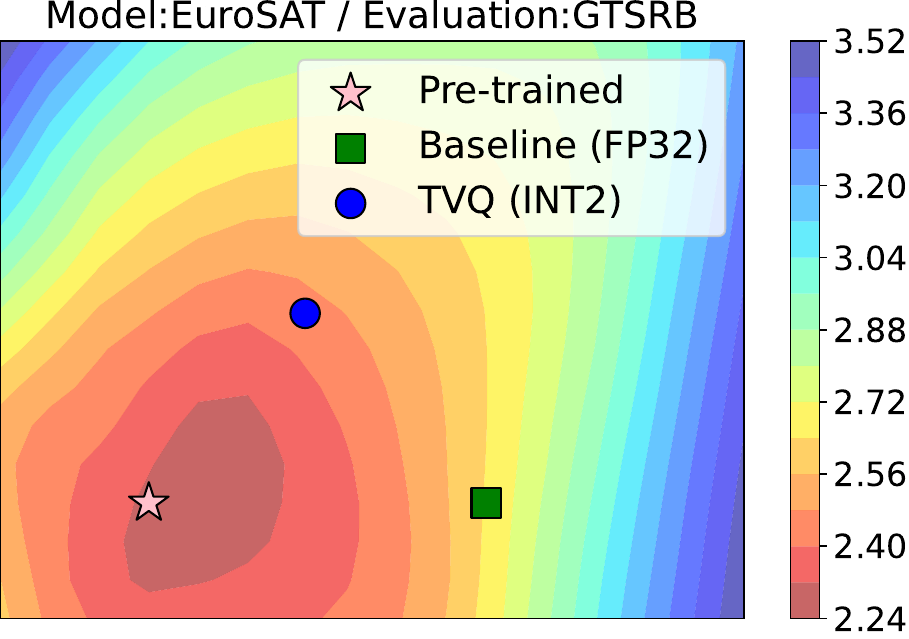}} &
        \subfloat{\includegraphics[width=0.23\linewidth]{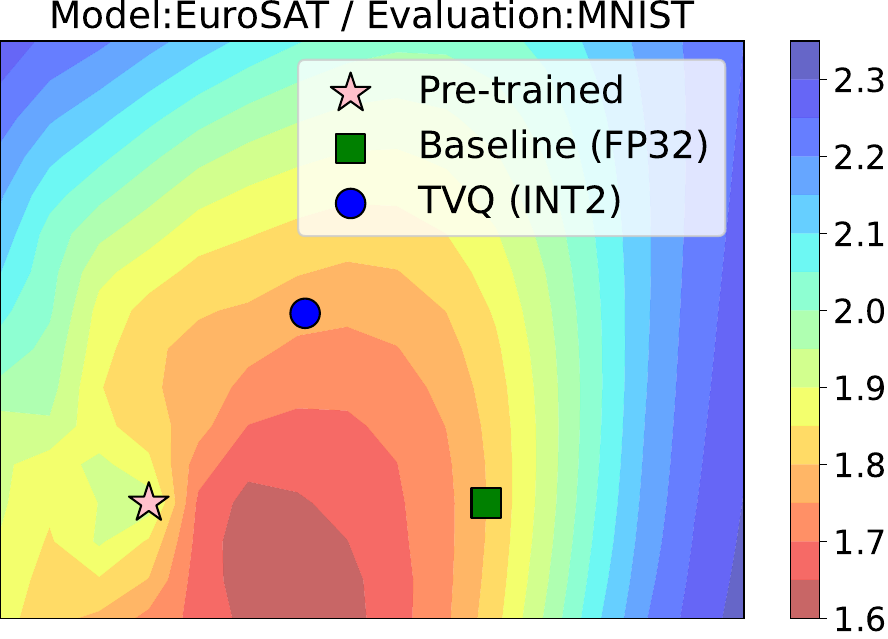}} &
        \subfloat{\includegraphics[width=0.23\linewidth]{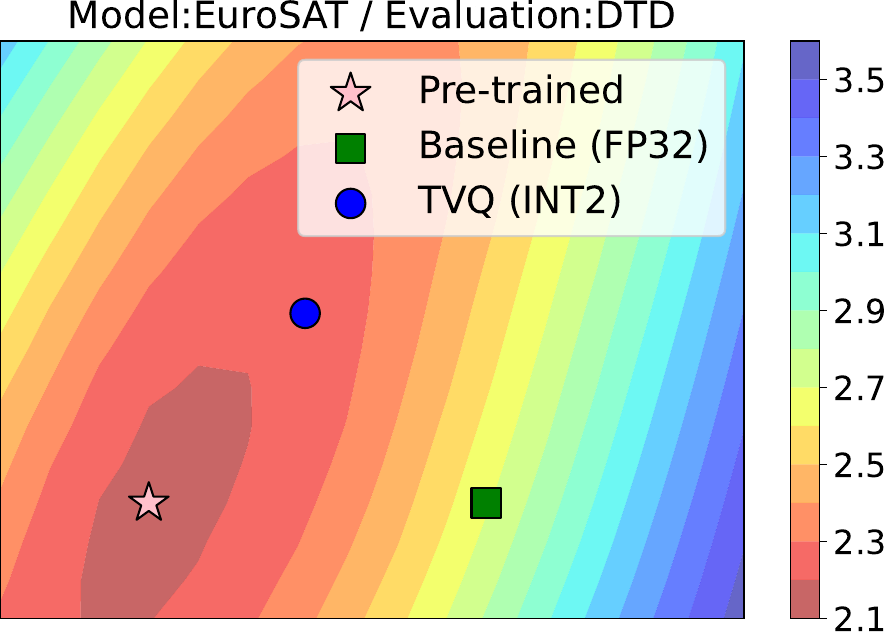}} \\
        
        \subfloat{\includegraphics[width=0.23\linewidth]{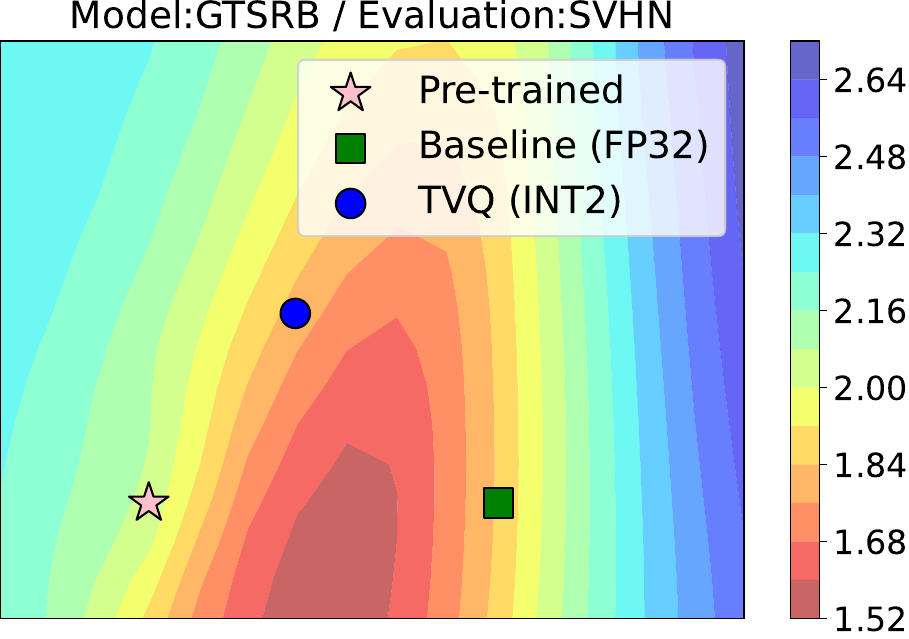}} &
        \subfloat{\includegraphics[width=0.23\linewidth]{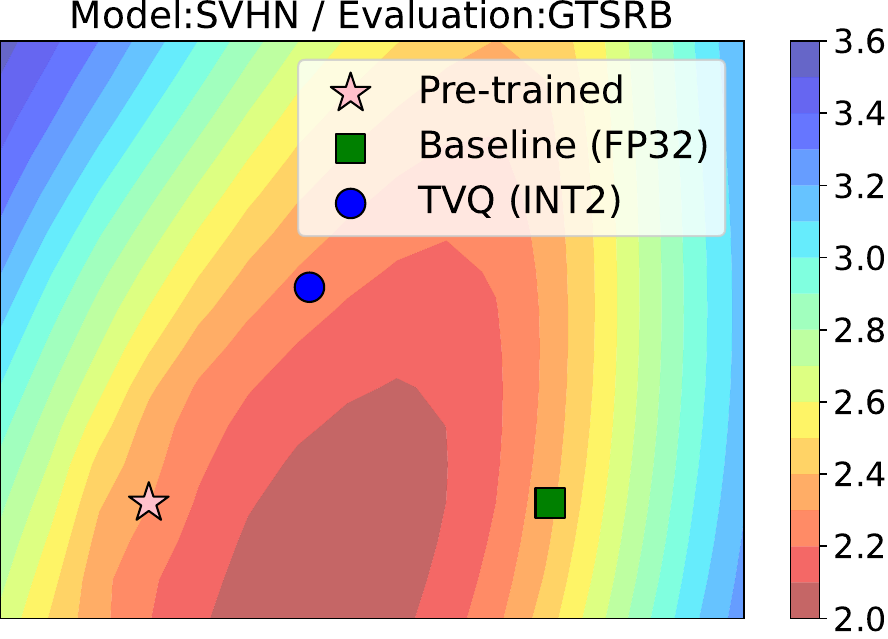}} &
        \subfloat{\includegraphics[width=0.23\linewidth]{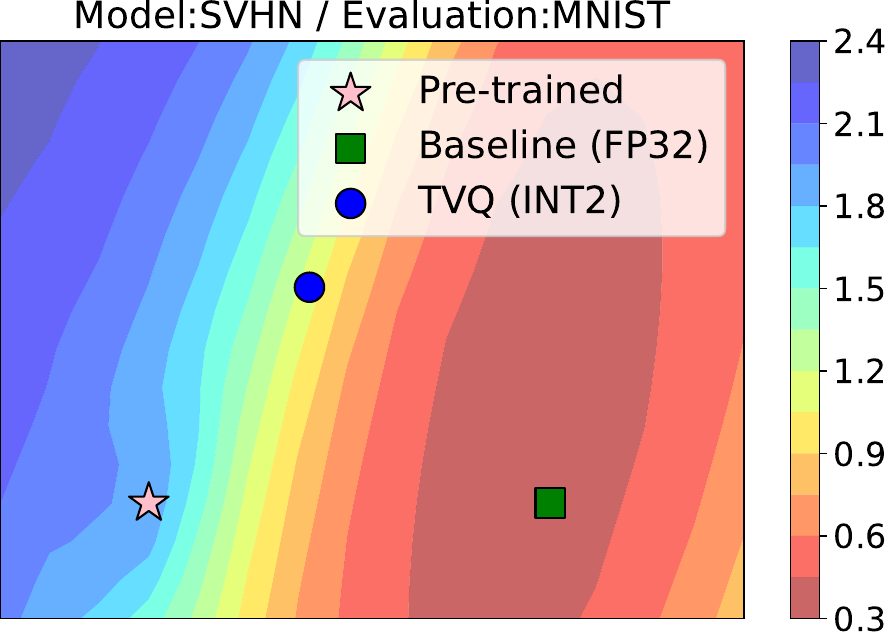}} &
        \subfloat{\includegraphics[width=0.23\linewidth]{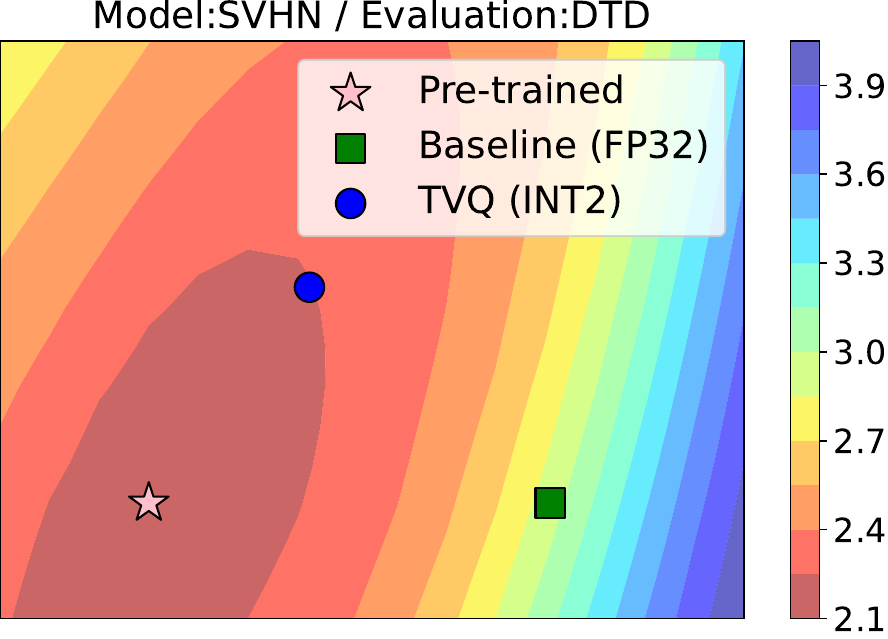}} \\
        
        \subfloat{\includegraphics[width=0.23\linewidth]{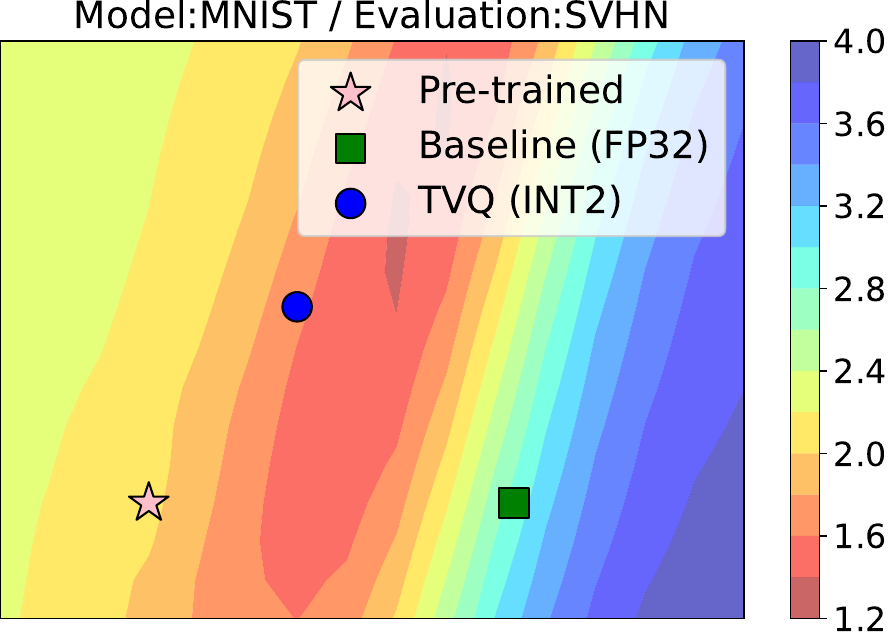}} &
        \subfloat{\includegraphics[width=0.23\linewidth]{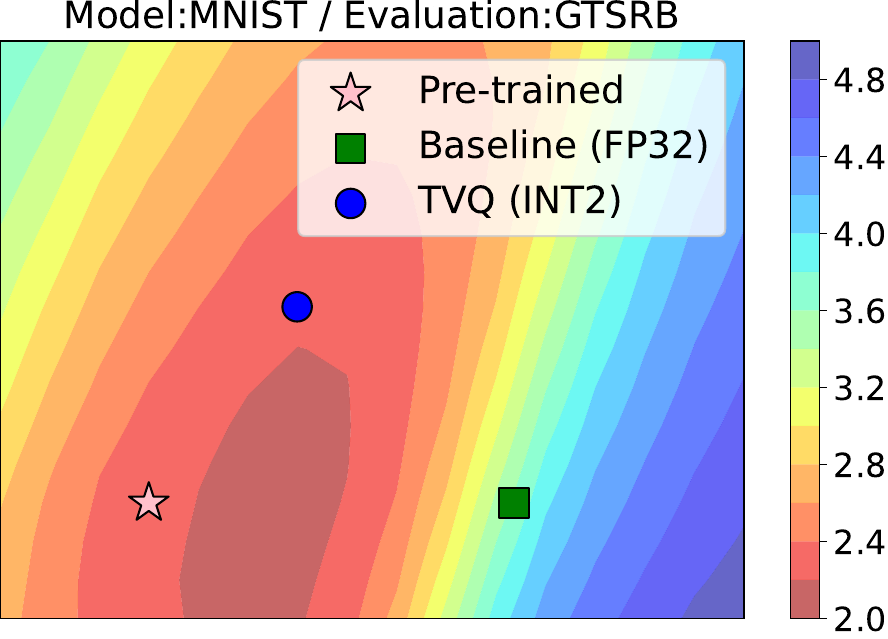}} &
        \subfloat{\includegraphics[width=0.23\linewidth]{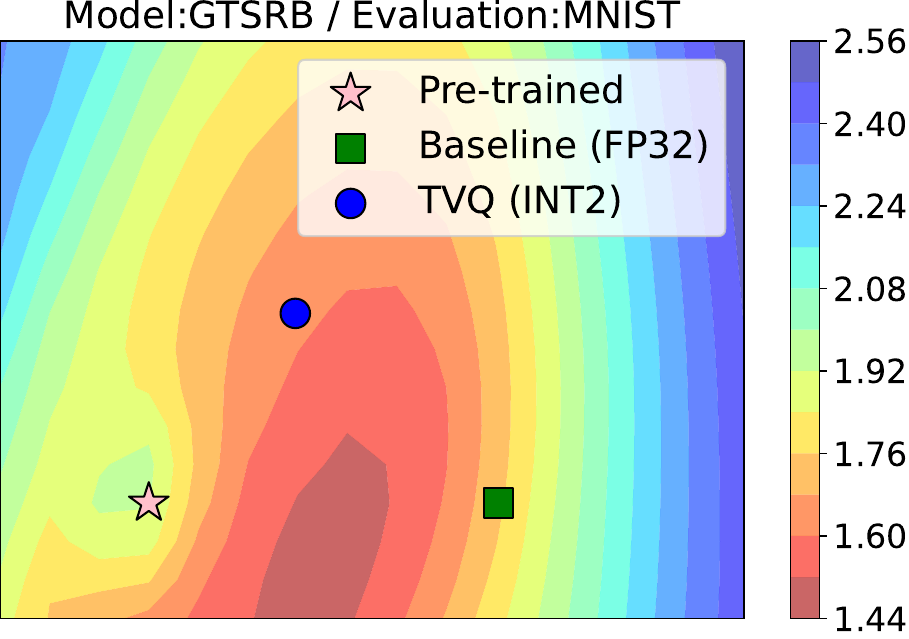}} &
        \subfloat{\includegraphics[width=0.23\linewidth]{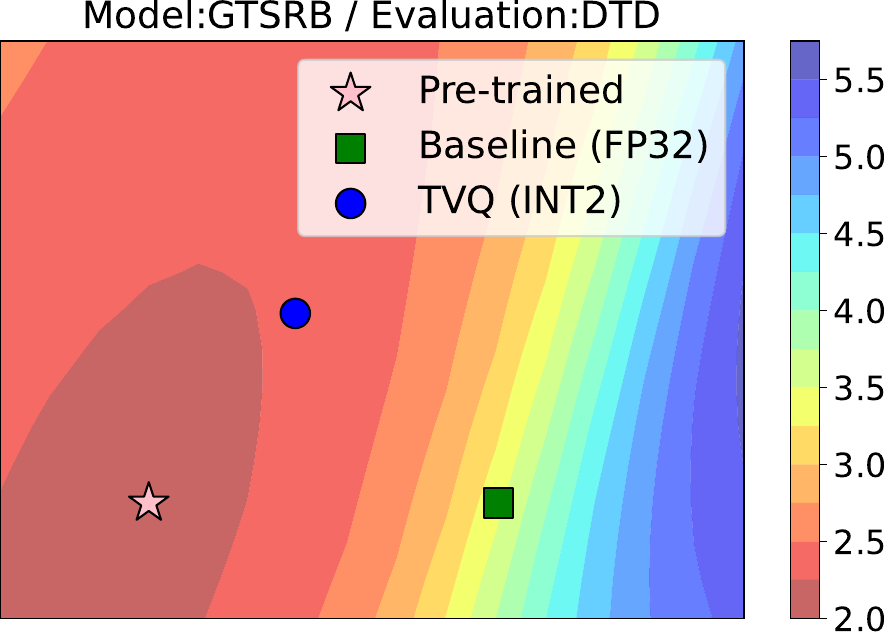}} \\
        
        \subfloat{\includegraphics[width=0.23\linewidth]{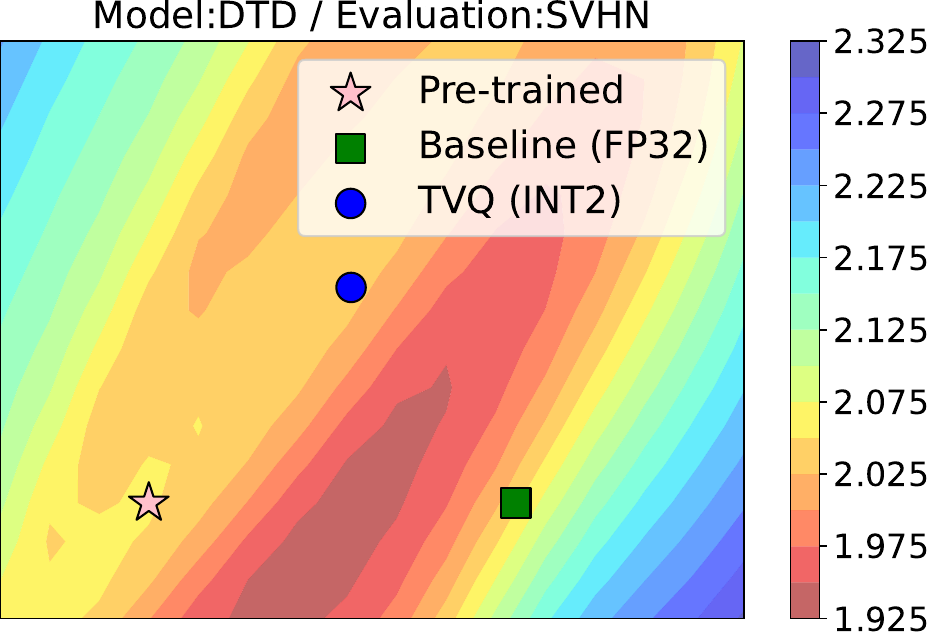}} &
        \subfloat{\includegraphics[width=0.23\linewidth]{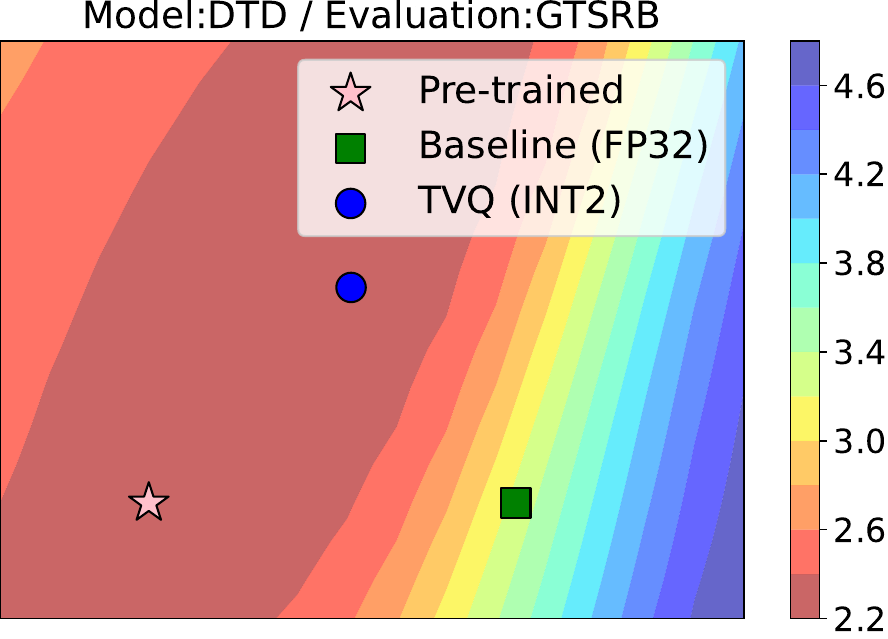}} &
        \subfloat{\includegraphics[width=0.23\linewidth]{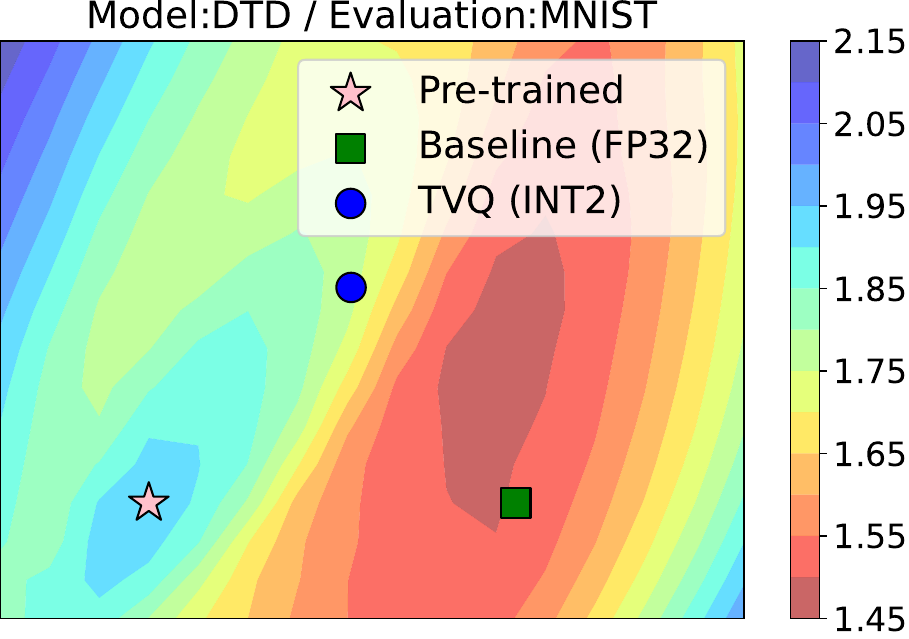}} &
        \subfloat{\includegraphics[width=0.23\linewidth]{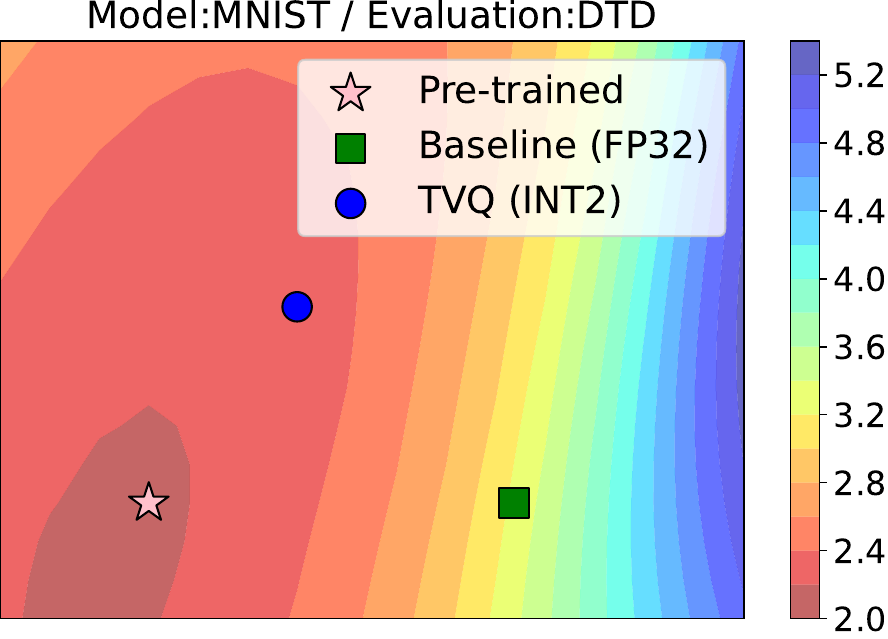}} \\
        
    \end{tabular}
    \caption{Loss landscape visualization of cross task pairs for 2-bit TVQ. The results show evaluations on SVHN, GTSRB, MNIST, and DTD.}
    \label{fig:tvq_landscape_control2}
\end{figure*}

\begin{figure*}[t]
    \centering
    \renewcommand{\thesubfigure}{} %
    \setlength{\tabcolsep}{1pt} %
    \renewcommand{\arraystretch}{0.8} %
    \begin{tabular}{cccc} %
    
        \subfloat{\includegraphics[width=0.23\linewidth]{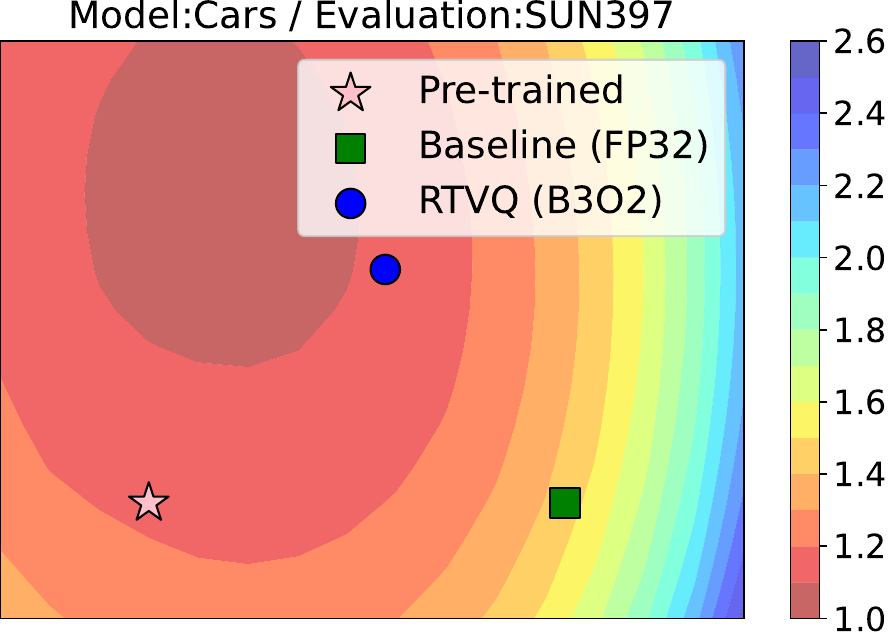}} &
        \subfloat{\includegraphics[width=0.23\linewidth]{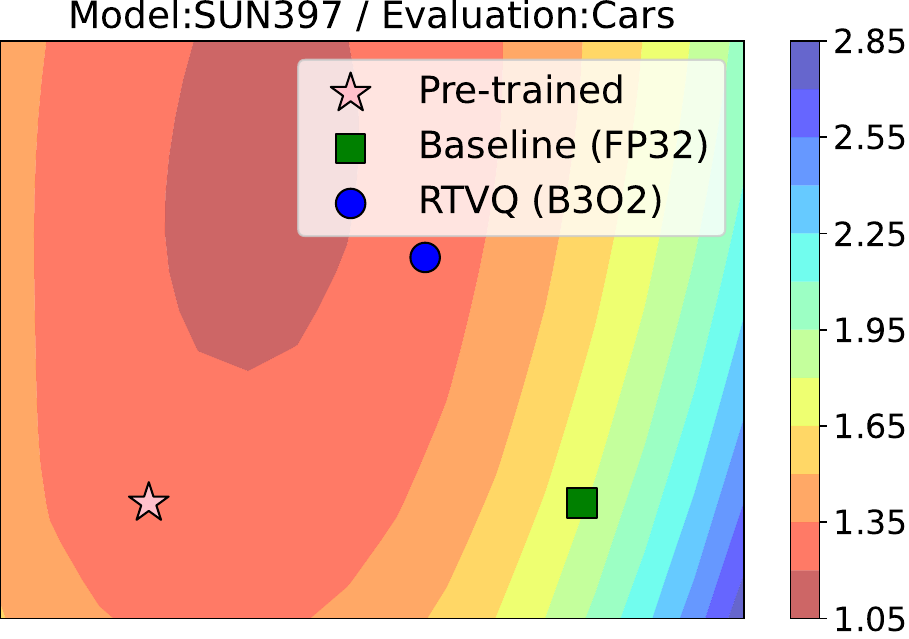}} &
        \subfloat{\includegraphics[width=0.23\linewidth]{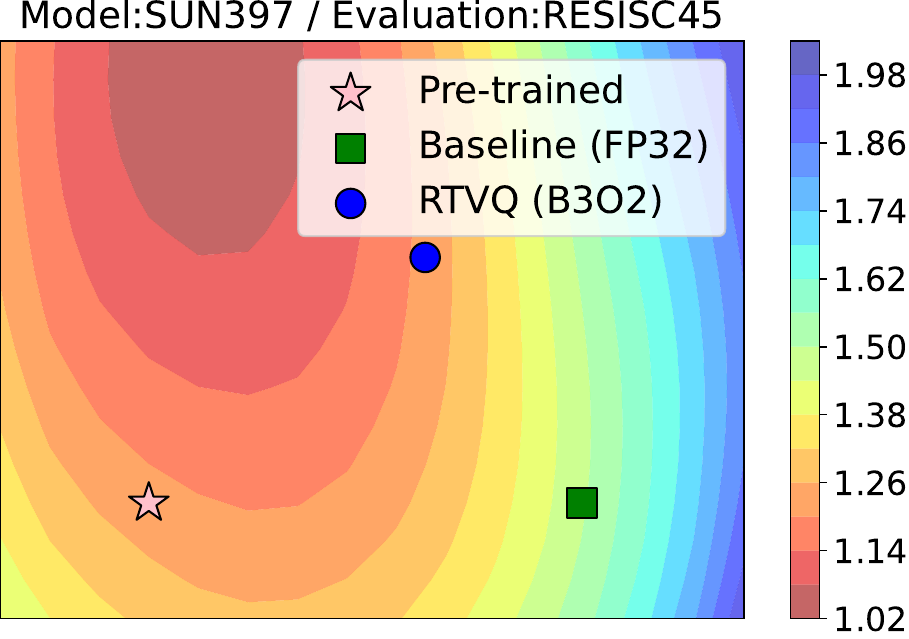}} &
        \subfloat{\includegraphics[width=0.23\linewidth]{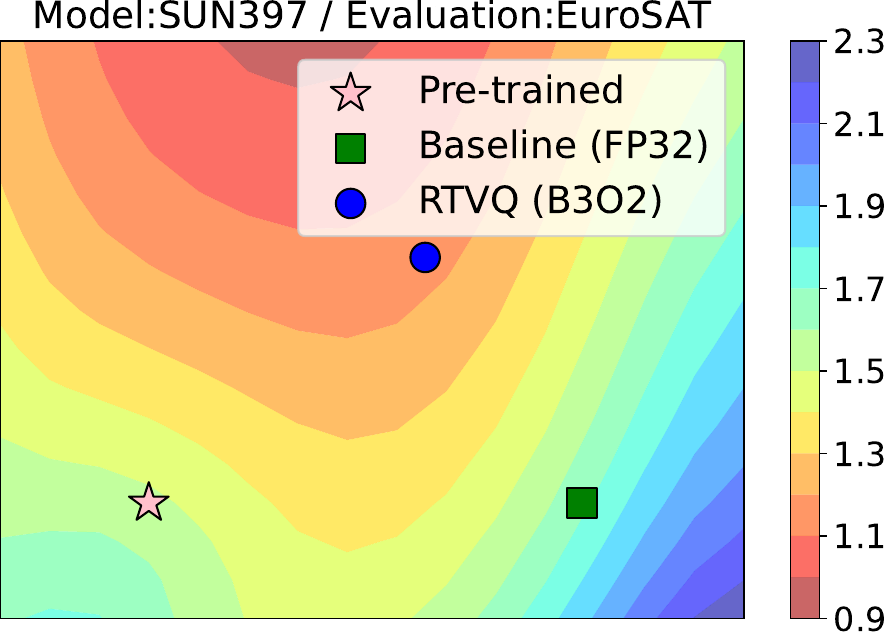}} \\
        
        \subfloat{\includegraphics[width=0.23\linewidth]{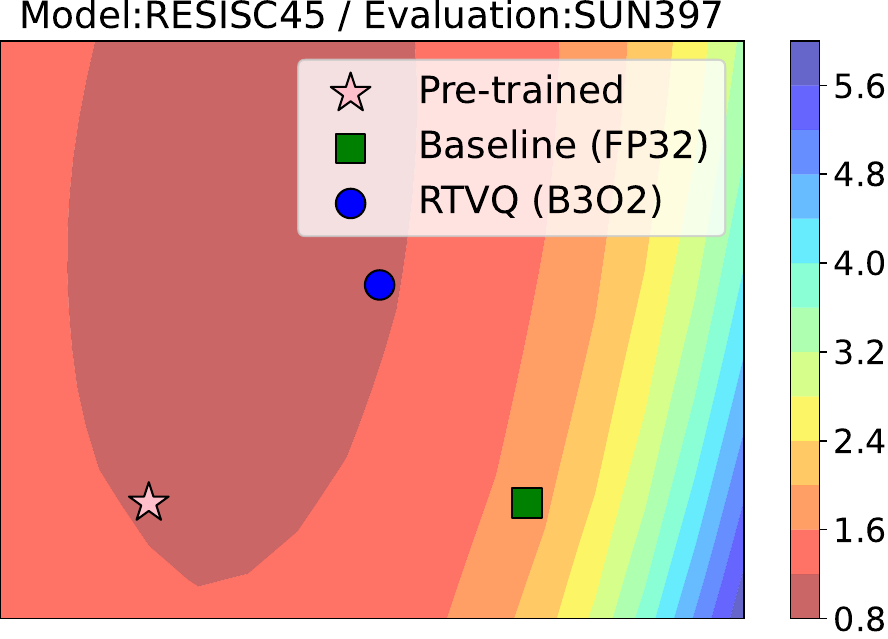}} &
        \subfloat{\includegraphics[width=0.23\linewidth]{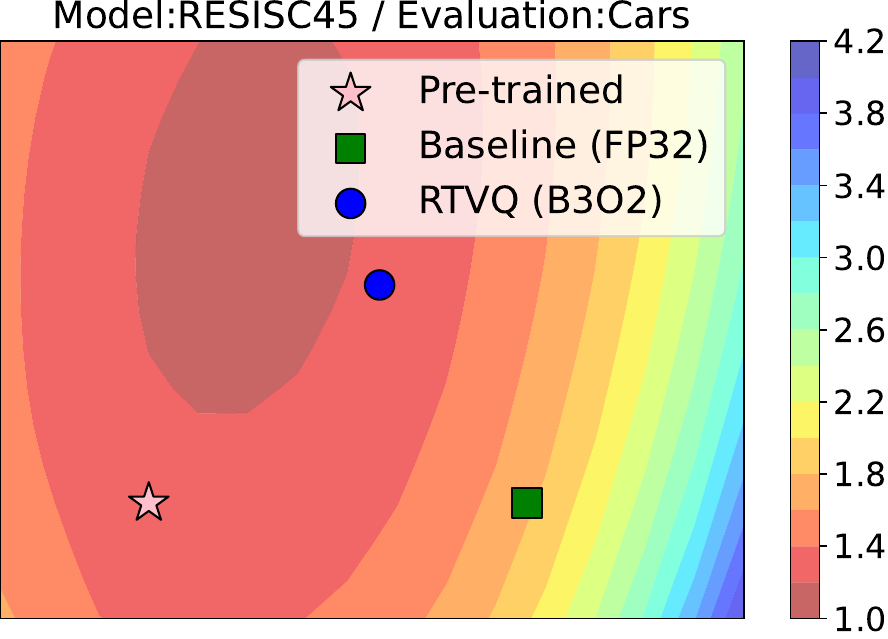}} &
        \subfloat{\includegraphics[width=0.23\linewidth]{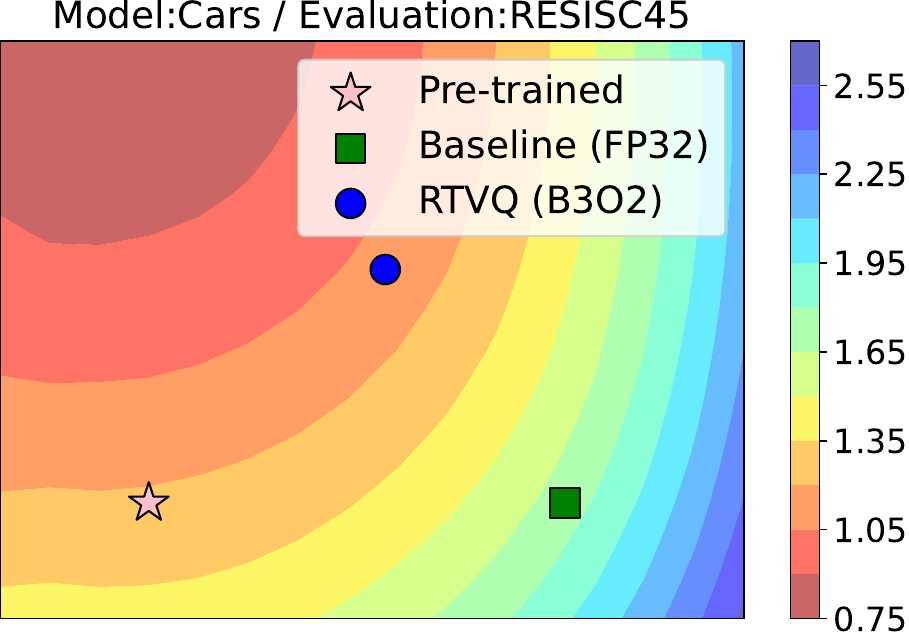}} &
        \subfloat{\includegraphics[width=0.23\linewidth]{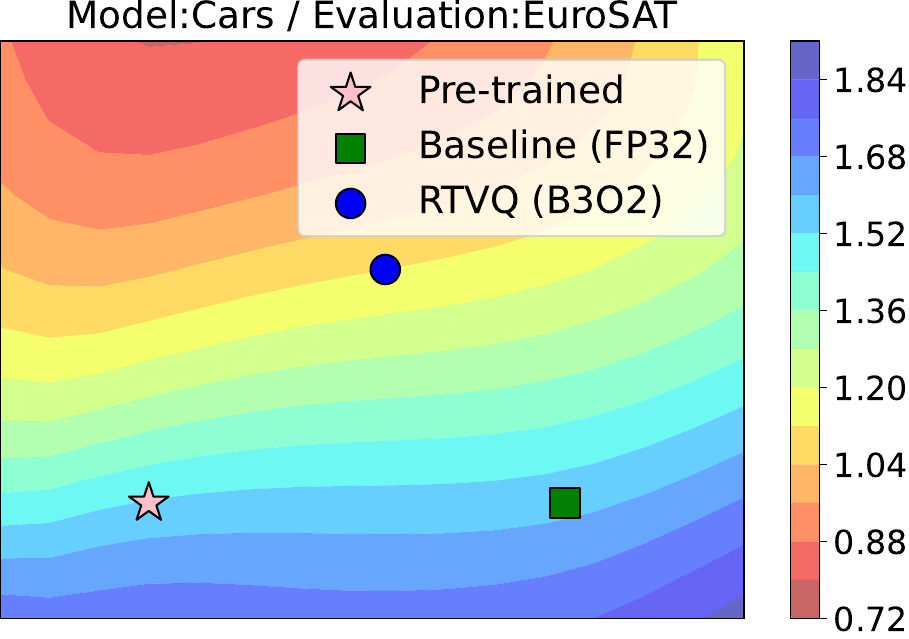}} \\
        
        \subfloat{\includegraphics[width=0.23\linewidth]{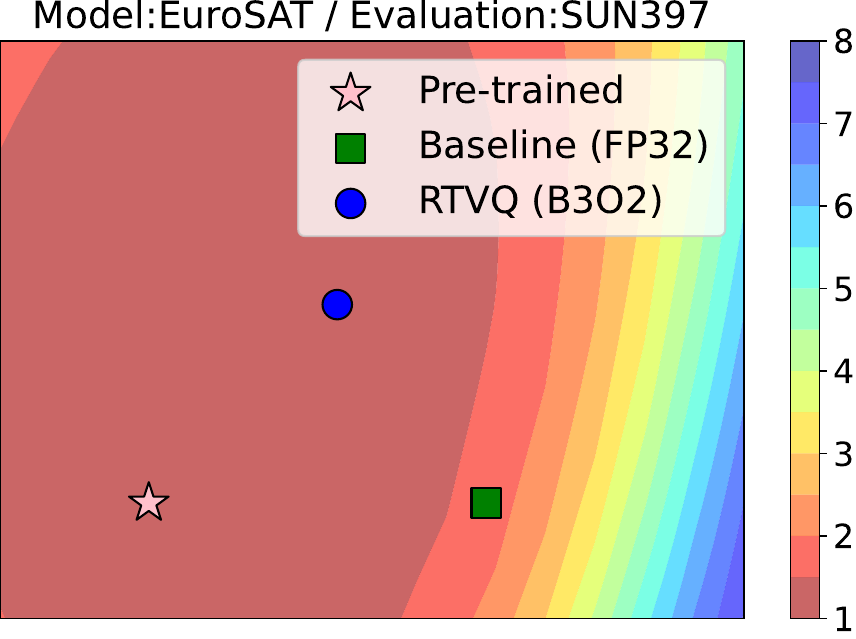}} &
        \subfloat{\includegraphics[width=0.23\linewidth]{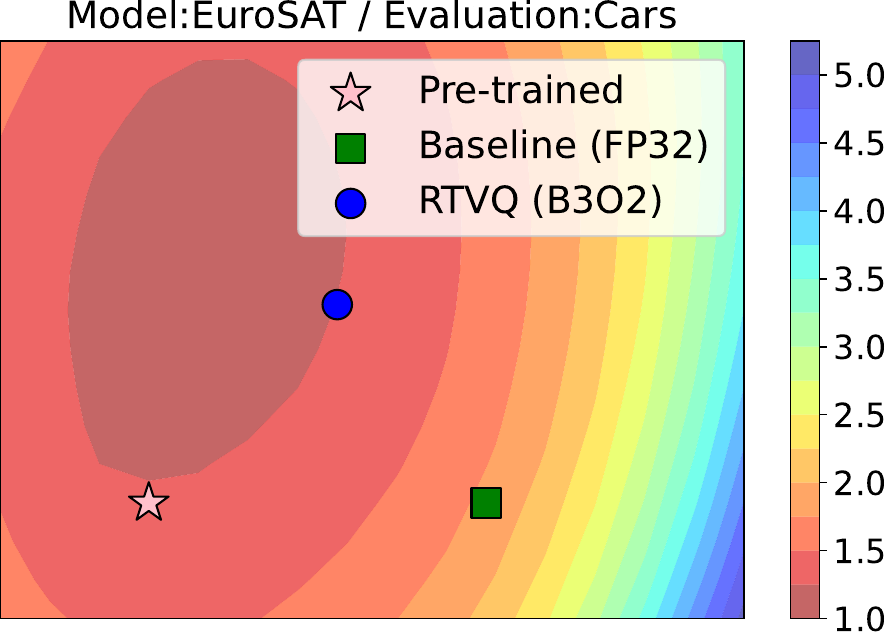}} &
        \subfloat{\includegraphics[width=0.23\linewidth]{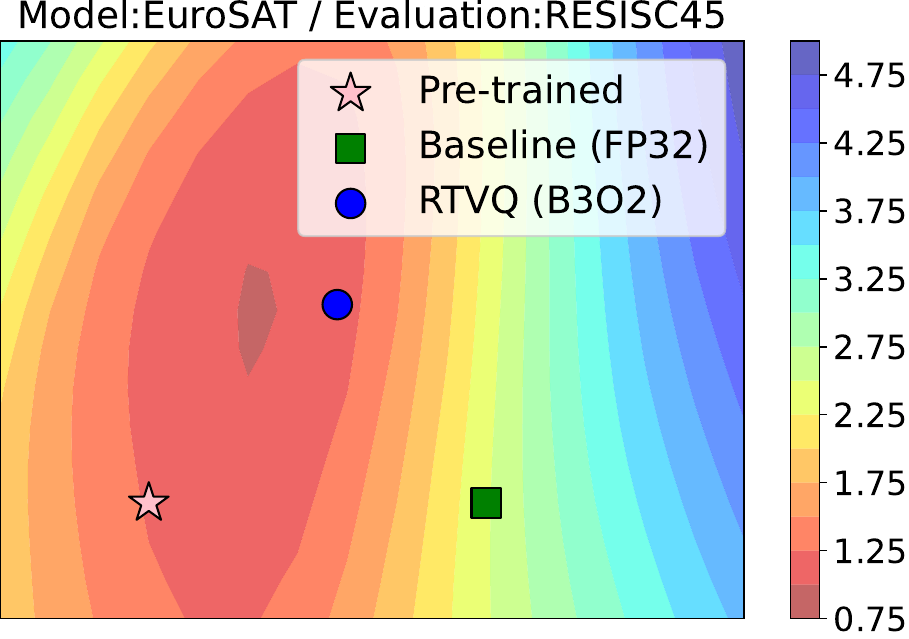}} &
        \subfloat{\includegraphics[width=0.23\linewidth]{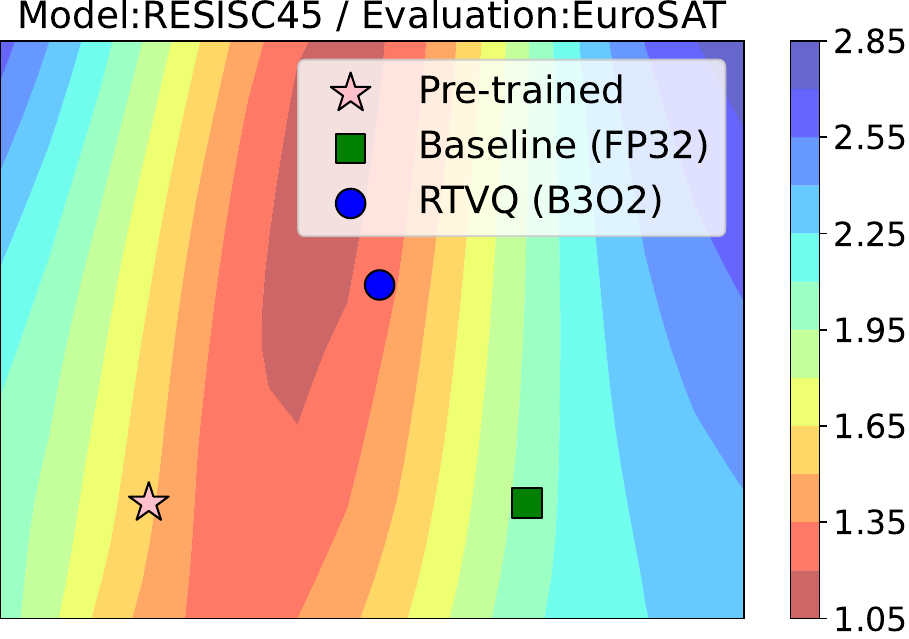}} \\
        
        \subfloat{\includegraphics[width=0.23\linewidth]{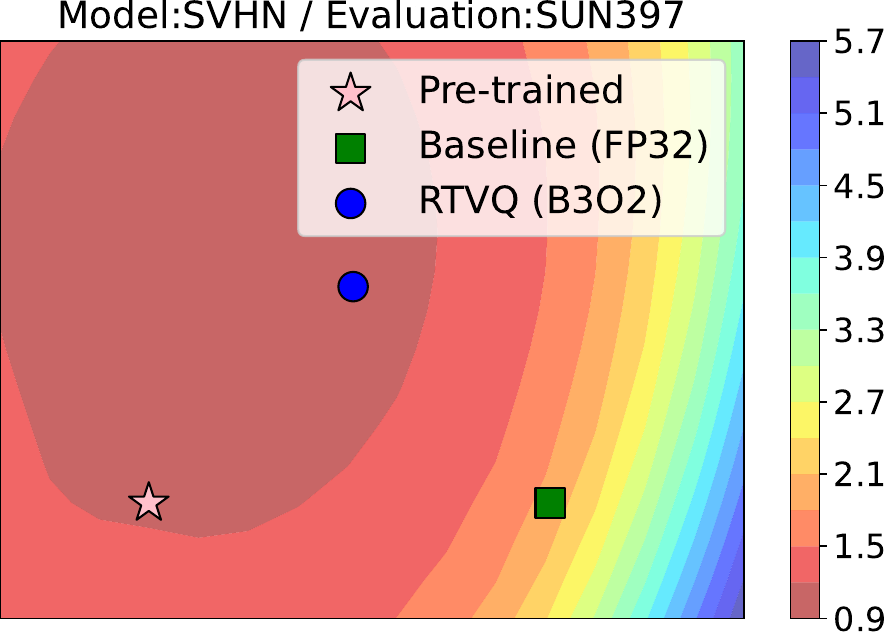}} &
        \subfloat{\includegraphics[width=0.23\linewidth]{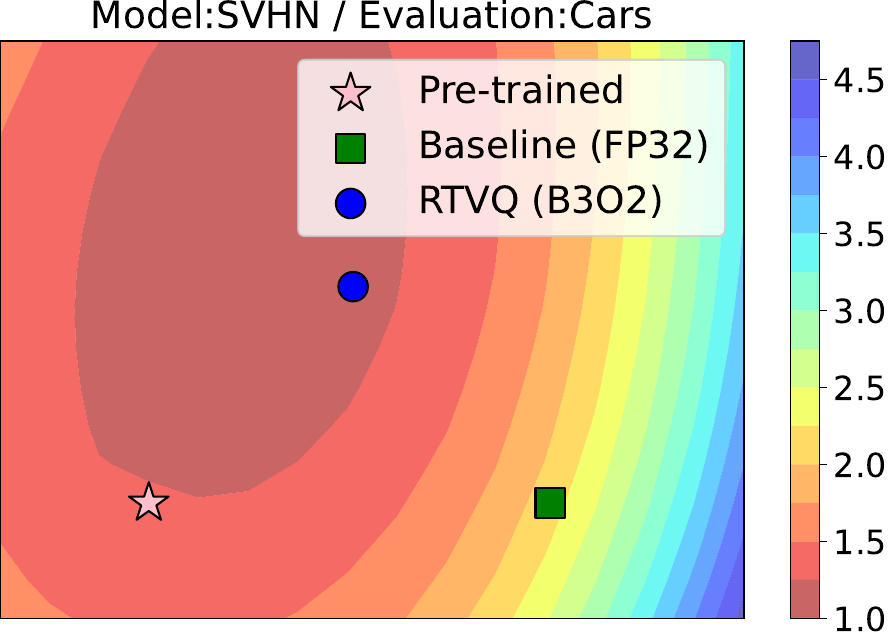}} &
        \subfloat{\includegraphics[width=0.23\linewidth]{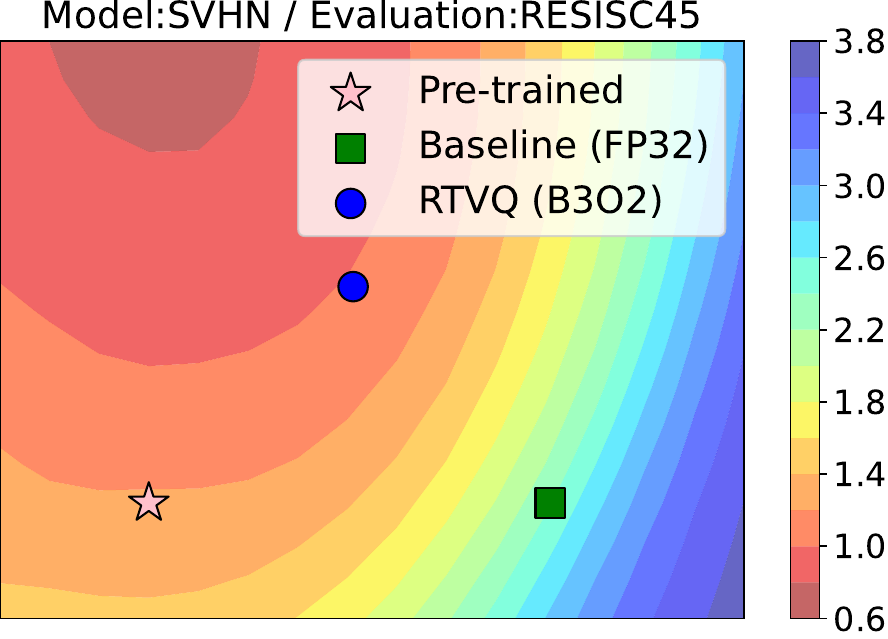}} &
        \subfloat{\includegraphics[width=0.23\linewidth]{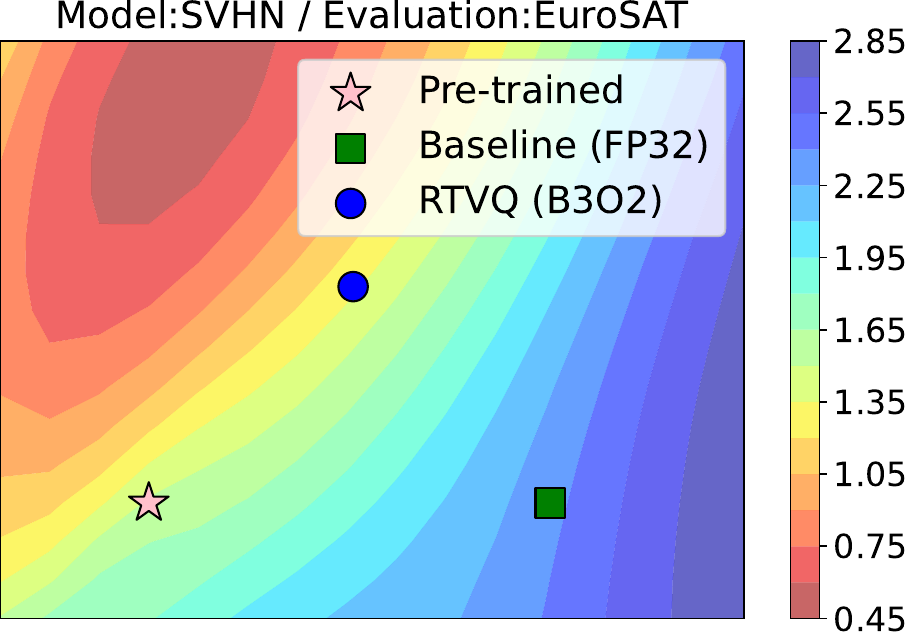}} \\
        
        \subfloat{\includegraphics[width=0.23\linewidth]{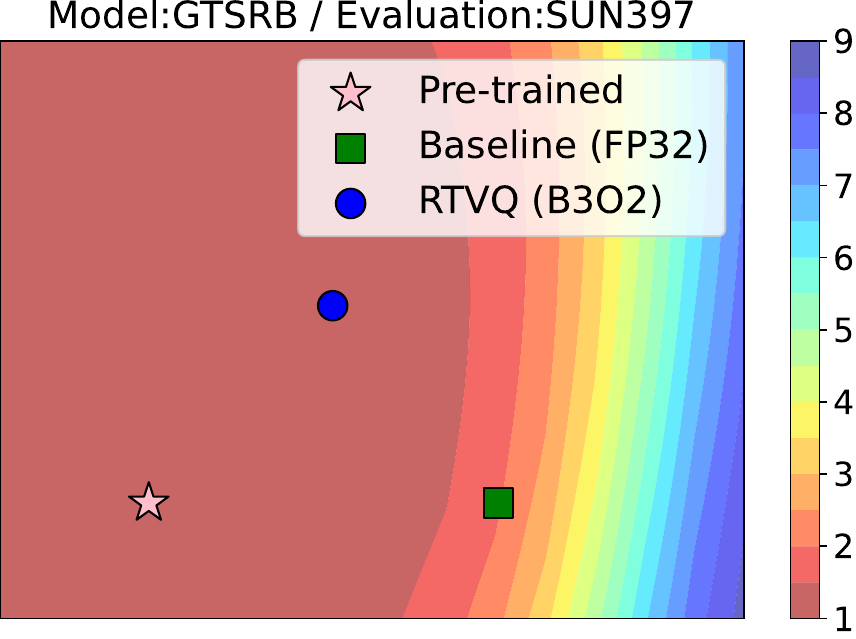}} &
        \subfloat{\includegraphics[width=0.23\linewidth]{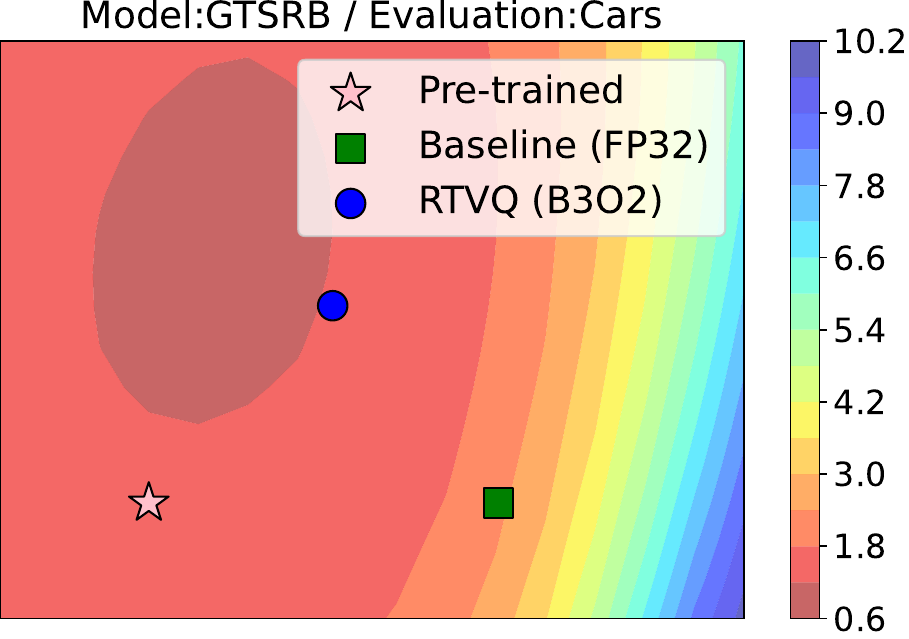}} &
        \subfloat{\includegraphics[width=0.23\linewidth]{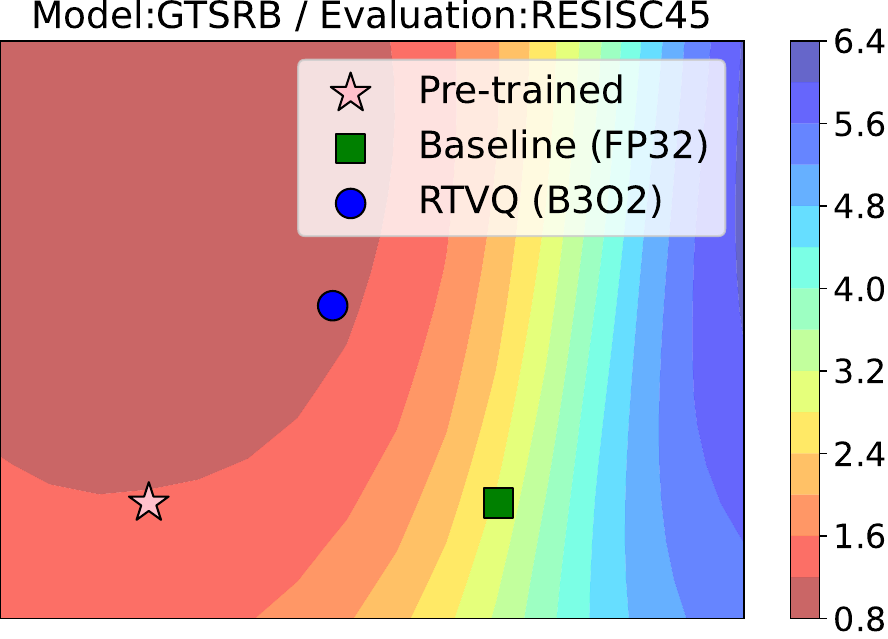}} &
        \subfloat{\includegraphics[width=0.23\linewidth]{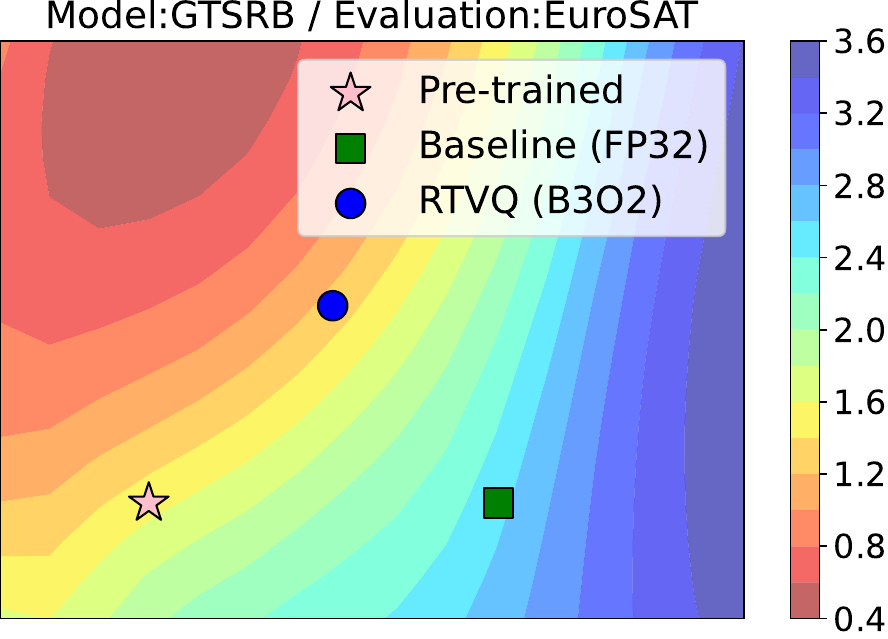}} \\
        
        \subfloat{\includegraphics[width=0.23\linewidth]{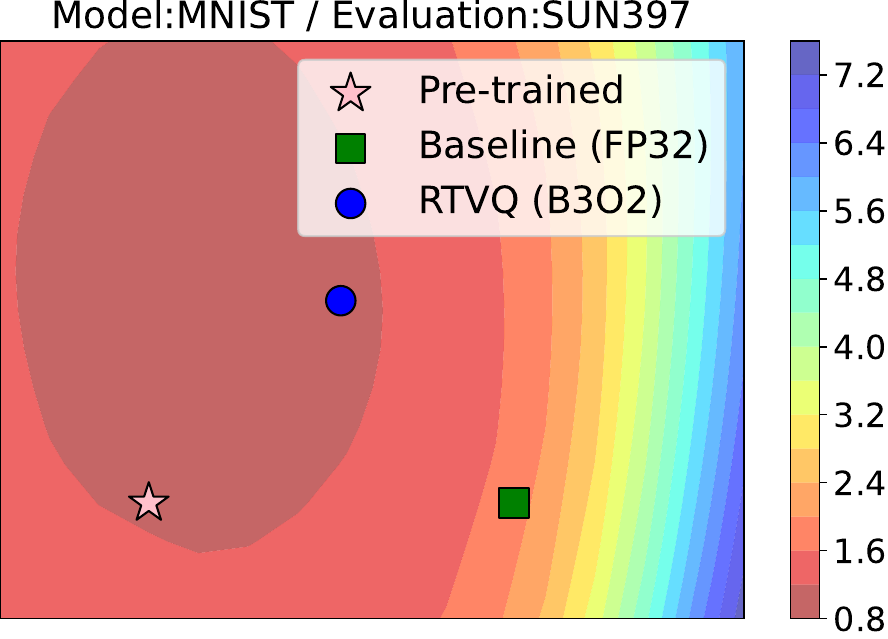}} &
        \subfloat{\includegraphics[width=0.23\linewidth]{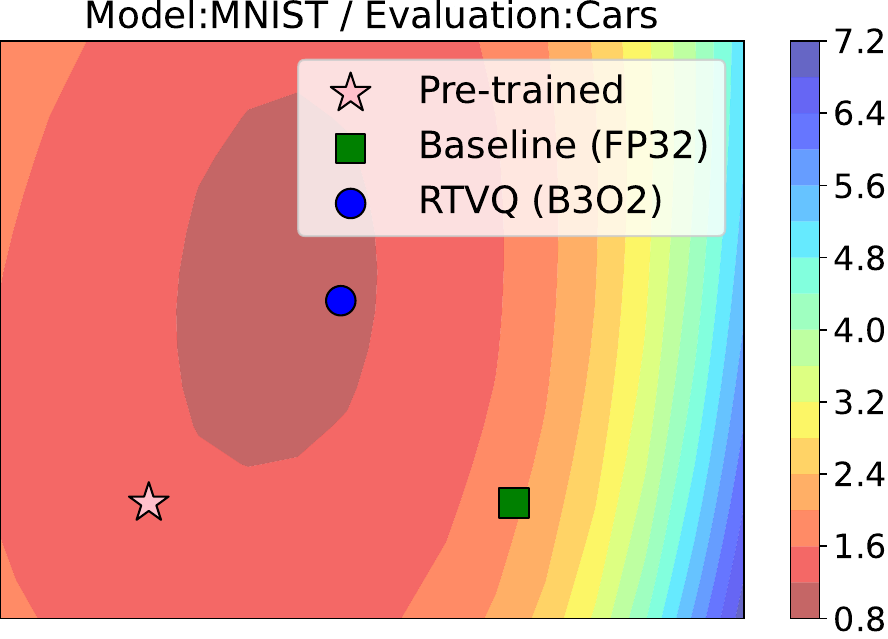}} &
        \subfloat{\includegraphics[width=0.23\linewidth]{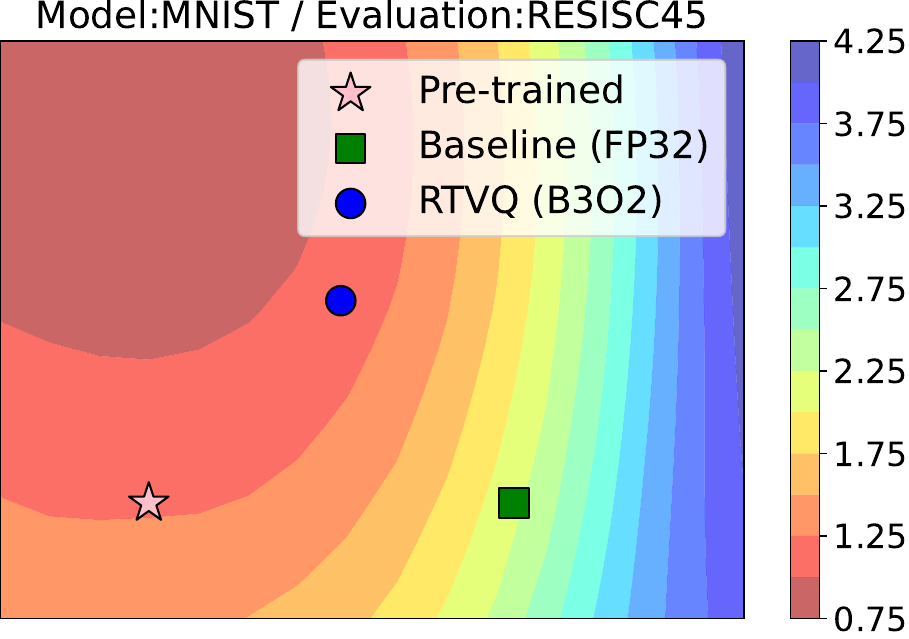}} &
        \subfloat{\includegraphics[width=0.23\linewidth]{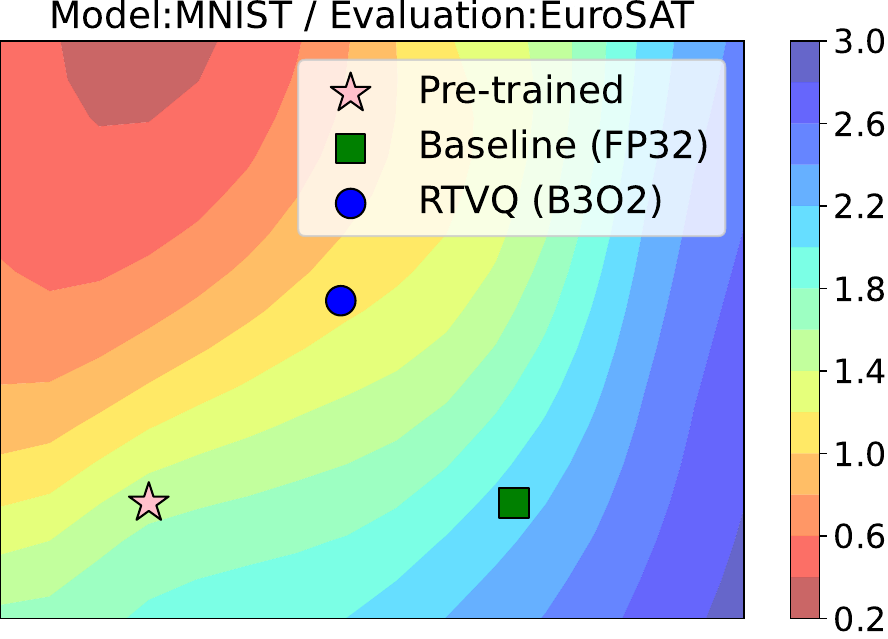}} \\
        
        \subfloat{\includegraphics[width=0.23\linewidth]{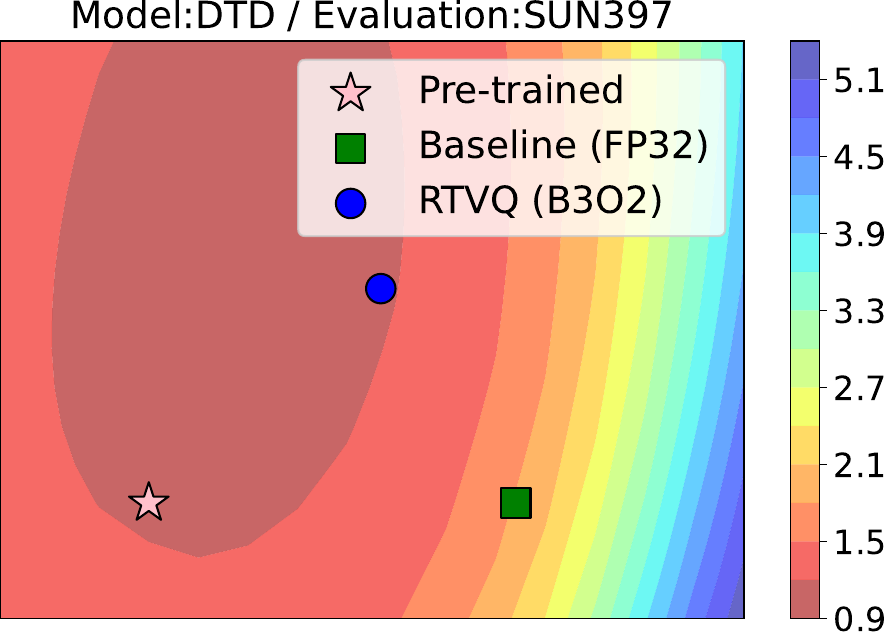}} &
        \subfloat{\includegraphics[width=0.23\linewidth]{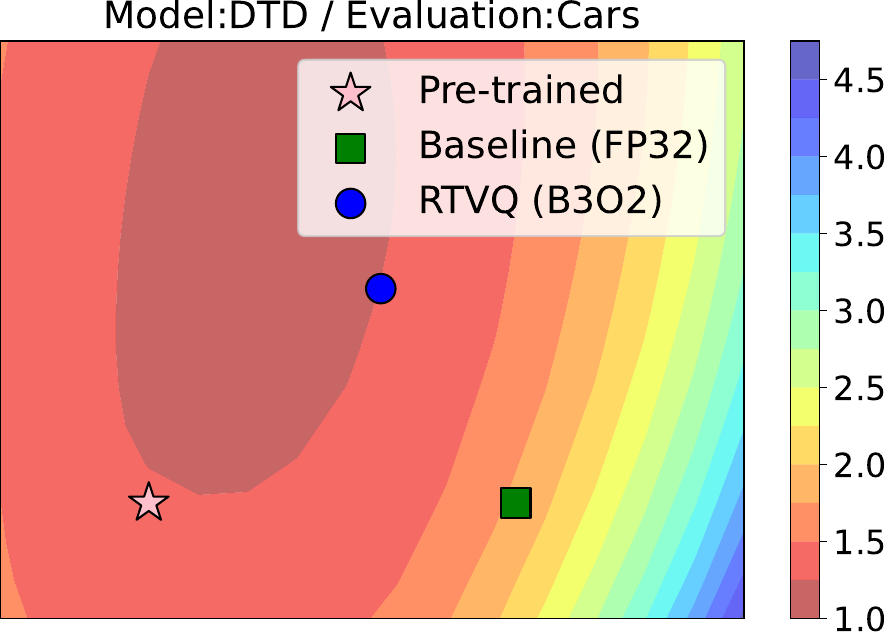}} &
        \subfloat{\includegraphics[width=0.23\linewidth]{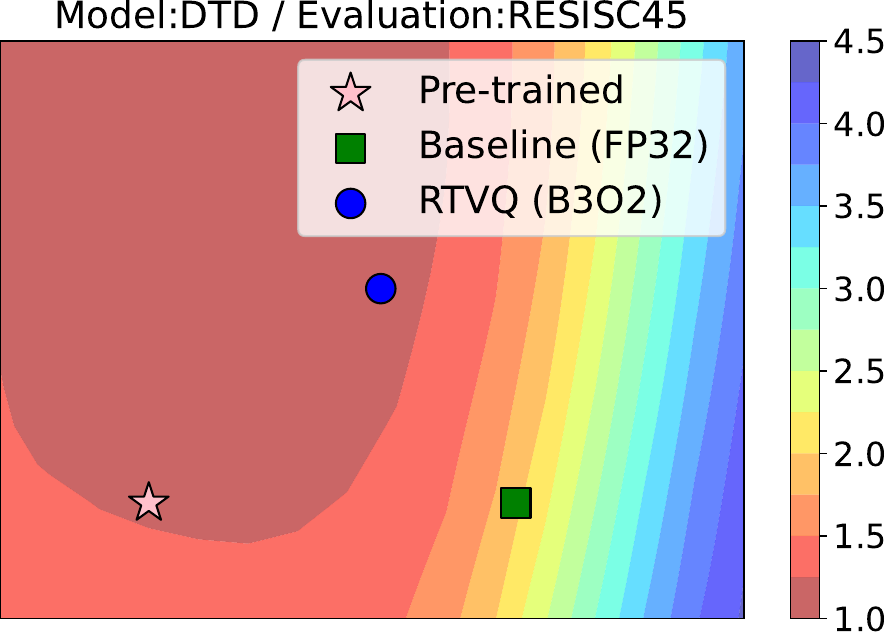}} &
        \subfloat{\includegraphics[width=0.23\linewidth]{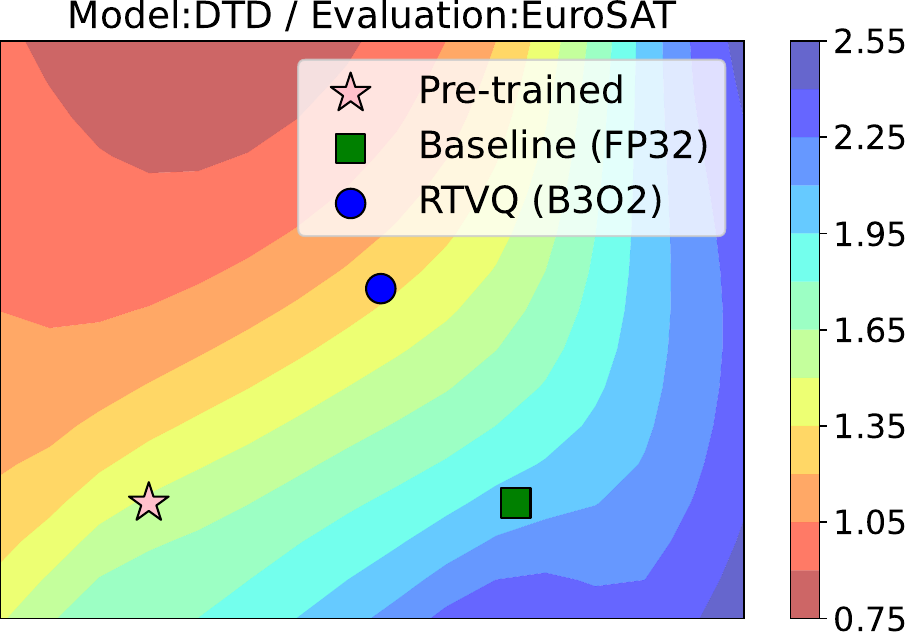}} \\

    \end{tabular}
    \caption{Loss landscape visualization of cross task pairs for RTVQ (B2O3). The results show evaluations on SUN397, Cars, RESISC45, and EuroSAT.}
    \label{fig:rtvq_landscape_control1}
\end{figure*}

\begin{figure*}[t]
    \centering
    \renewcommand{\thesubfigure}{} %
    \setlength{\tabcolsep}{1pt} %
    \renewcommand{\arraystretch}{0.8} %
    \begin{tabular}{cccc} %
    
        \subfloat{\includegraphics[width=0.23\linewidth]{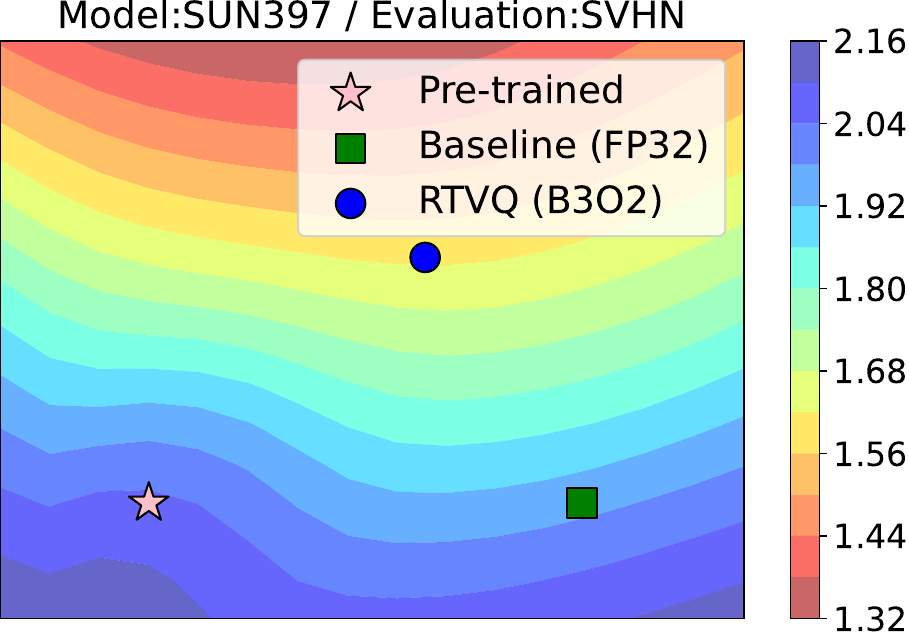}} &
        \subfloat{\includegraphics[width=0.23\linewidth]{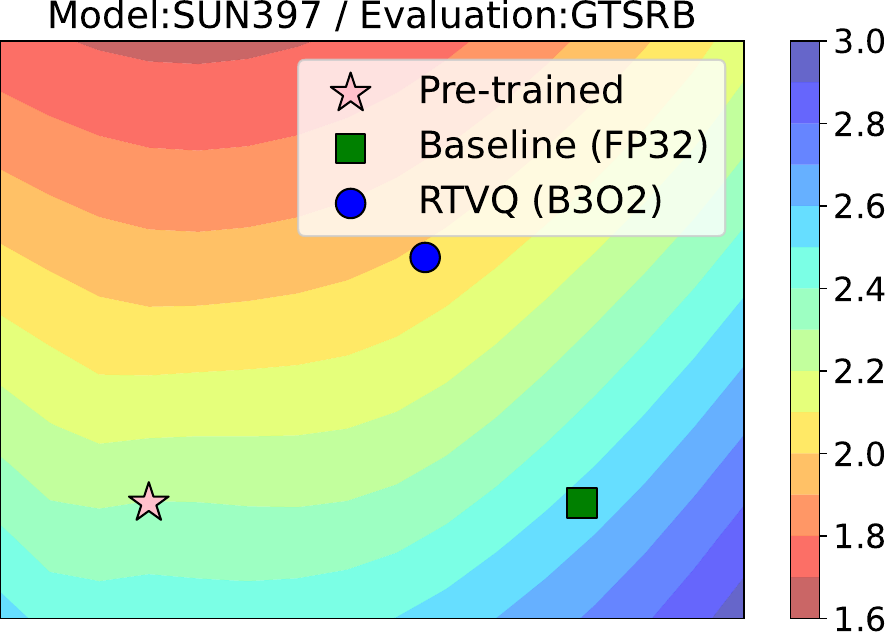}} &
        \subfloat{\includegraphics[width=0.23\linewidth]{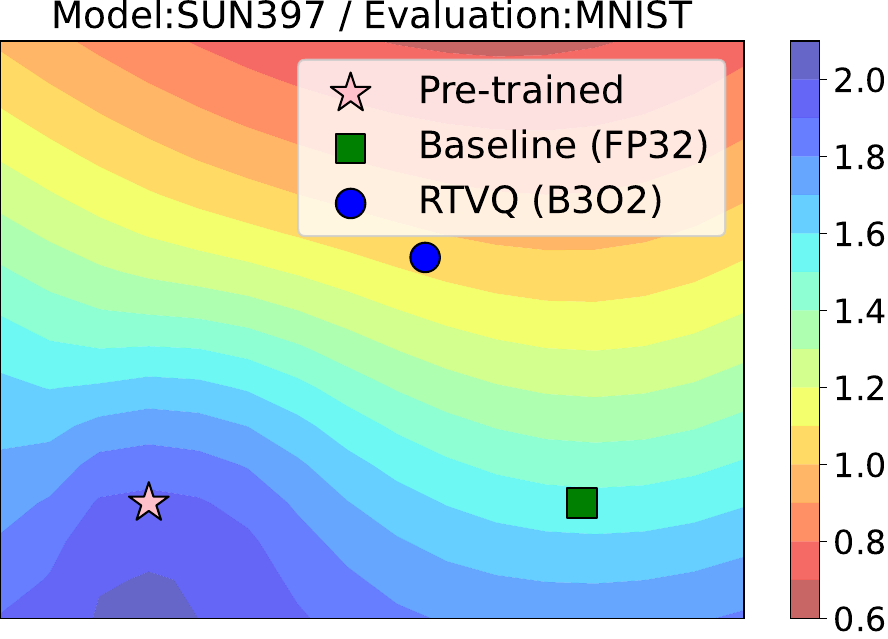}} &
        \subfloat{\includegraphics[width=0.23\linewidth]{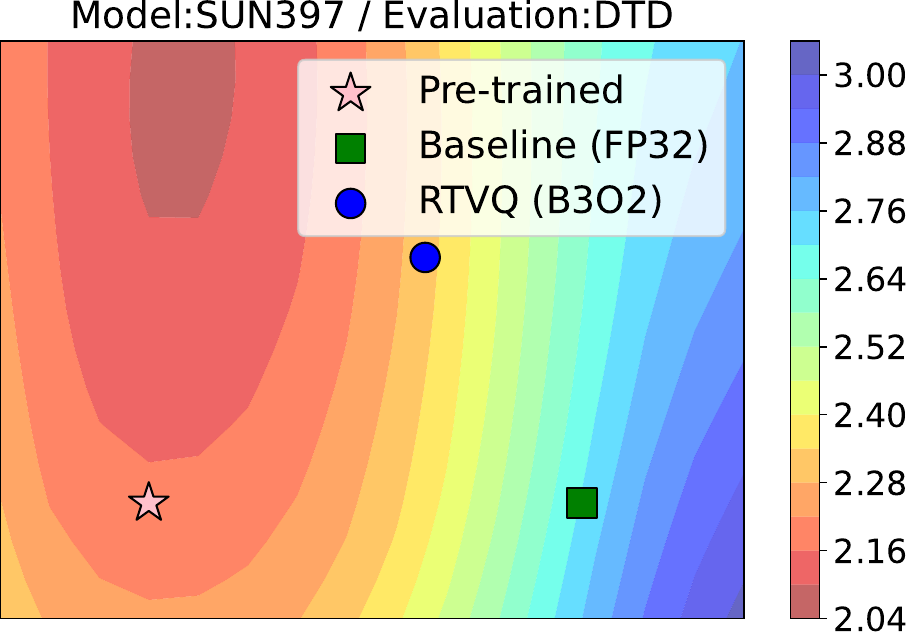}} \\
        
        \subfloat{\includegraphics[width=0.23\linewidth]{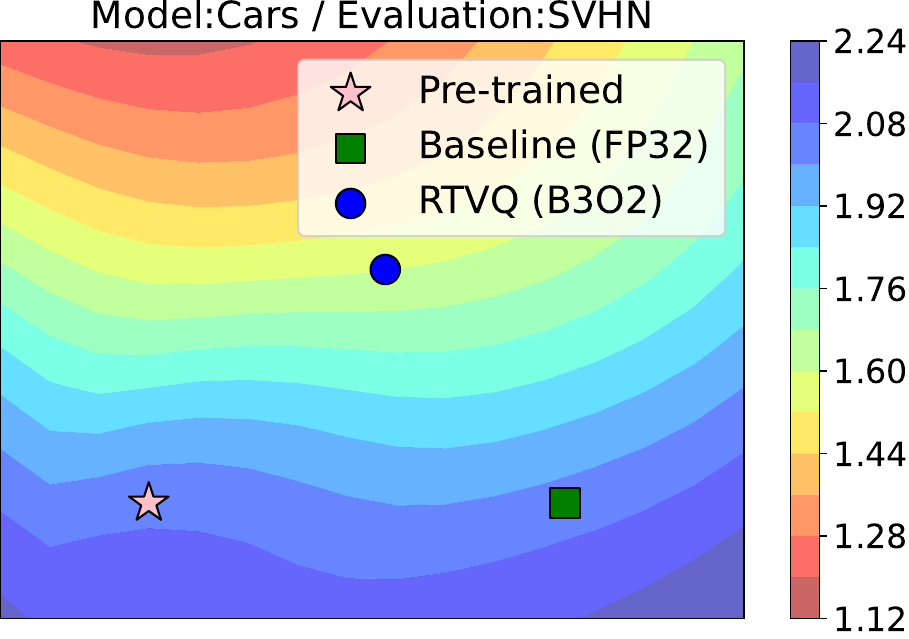}} &
        \subfloat{\includegraphics[width=0.23\linewidth]{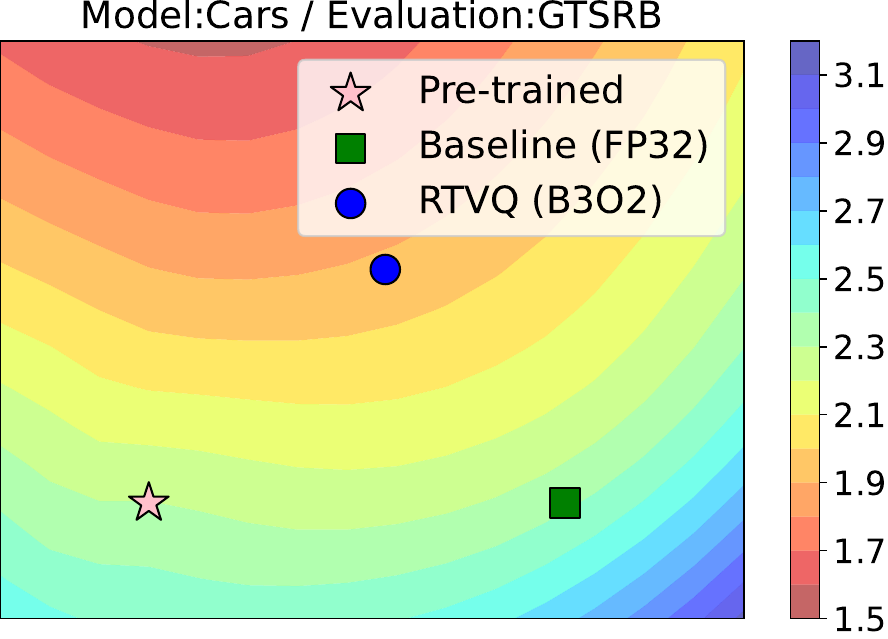}} &
        \subfloat{\includegraphics[width=0.23\linewidth]{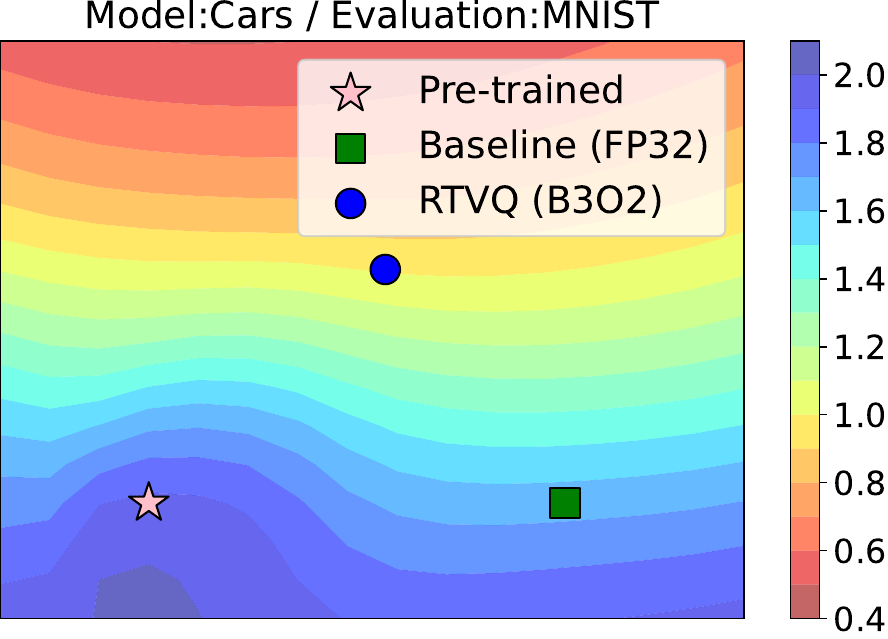}} &
        \subfloat{\includegraphics[width=0.23\linewidth]{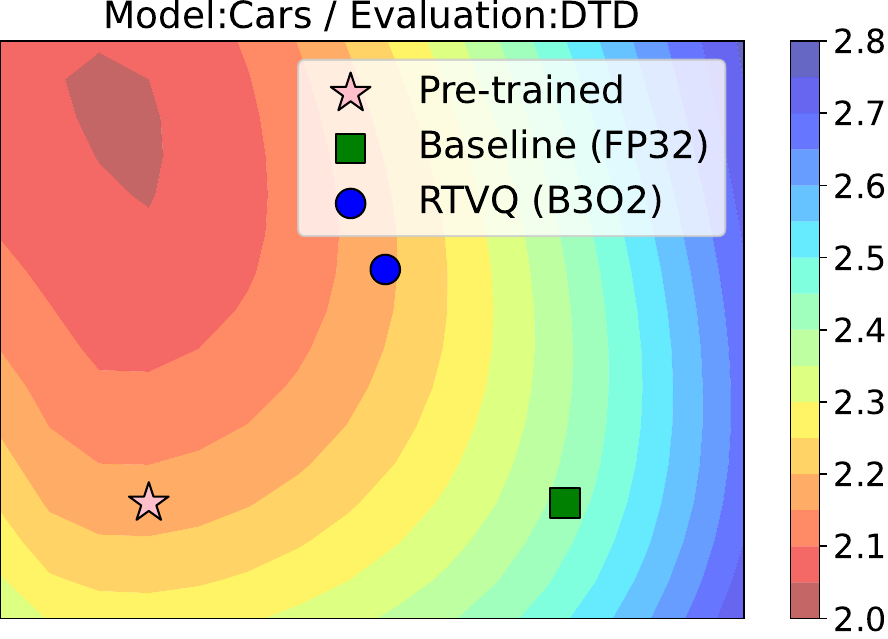}} \\
        
        \subfloat{\includegraphics[width=0.23\linewidth]{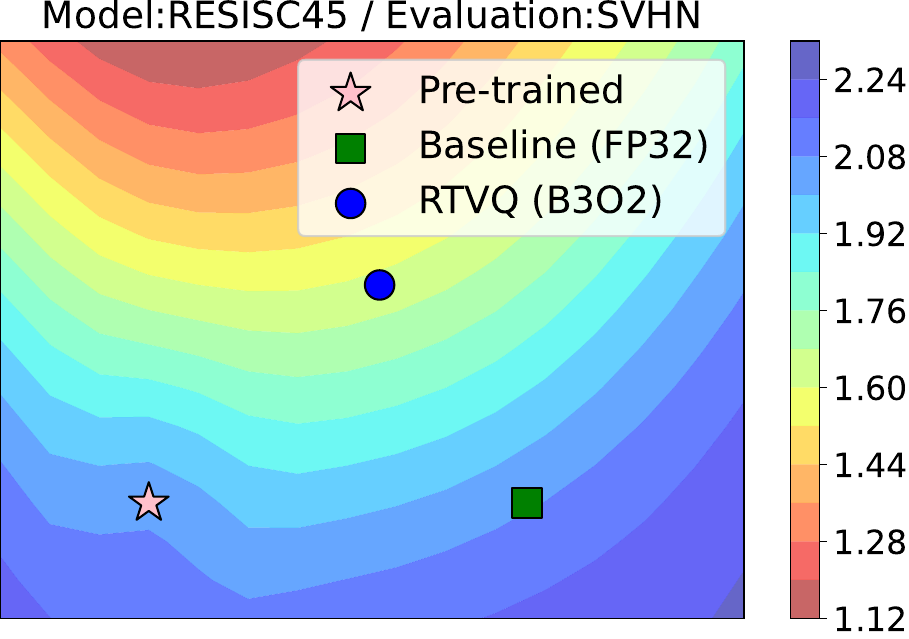}} &
        \subfloat{\includegraphics[width=0.23\linewidth]{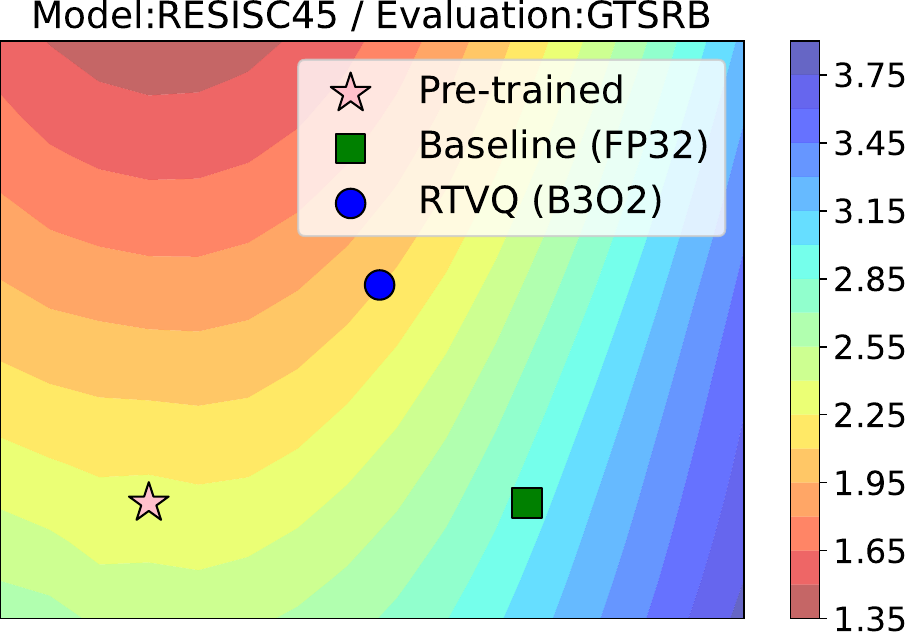}} &
        \subfloat{\includegraphics[width=0.23\linewidth]{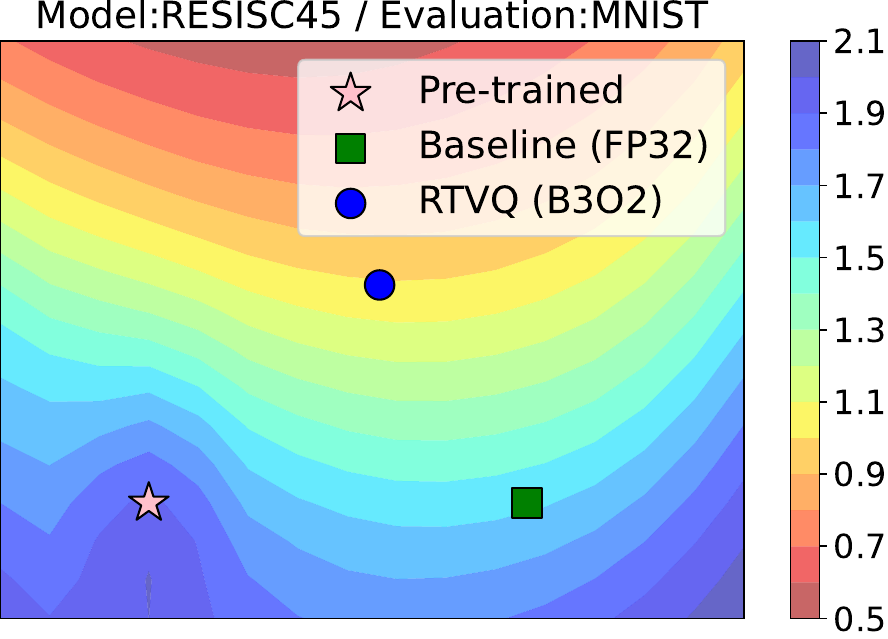}} &
        \subfloat{\includegraphics[width=0.23\linewidth]{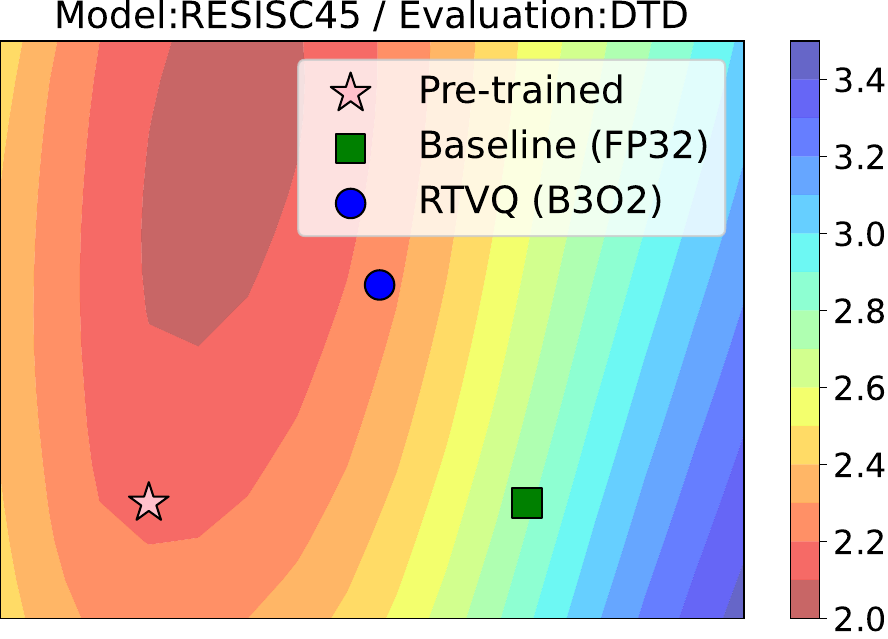}} \\
        
        \subfloat{\includegraphics[width=0.23\linewidth]{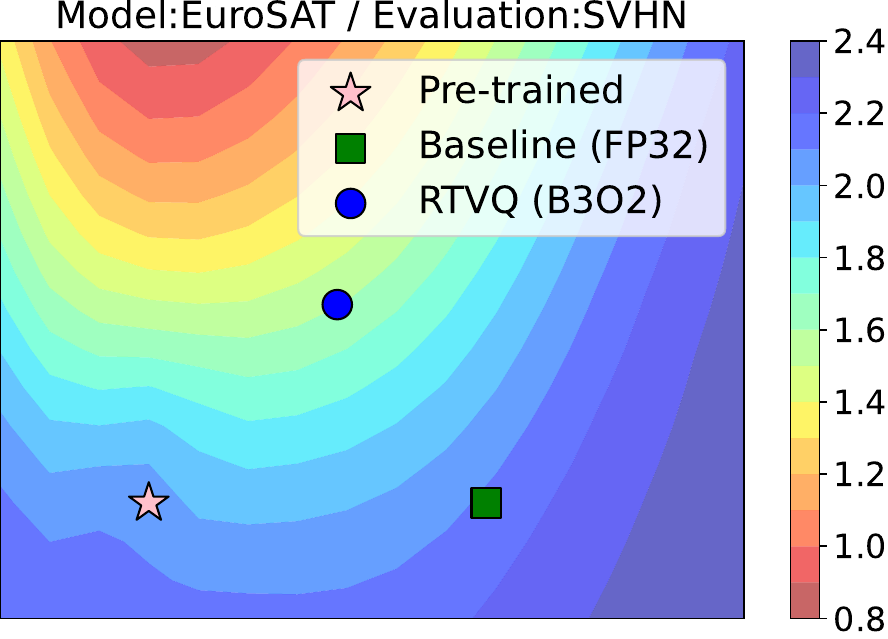}} &
        \subfloat{\includegraphics[width=0.23\linewidth]{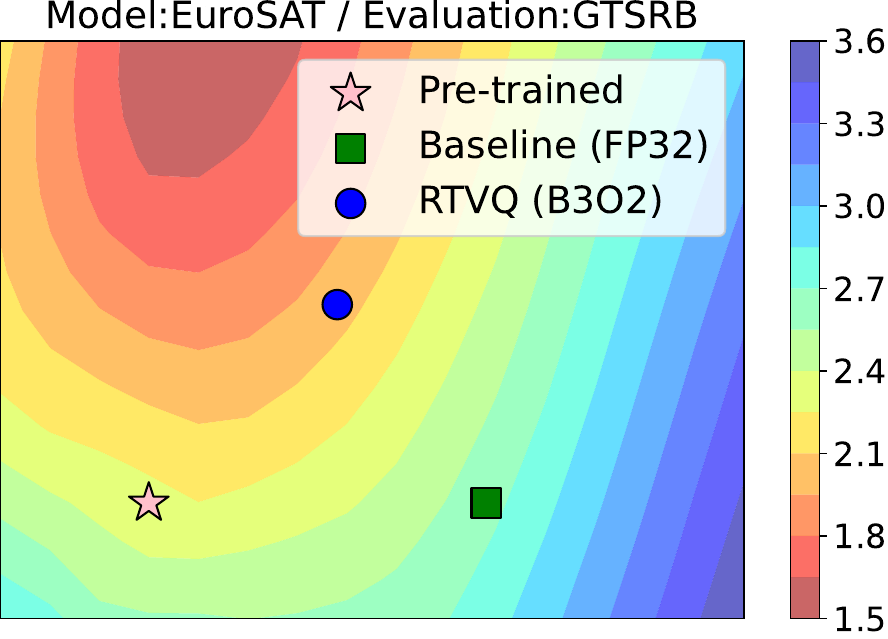}} &
        \subfloat{\includegraphics[width=0.23\linewidth]{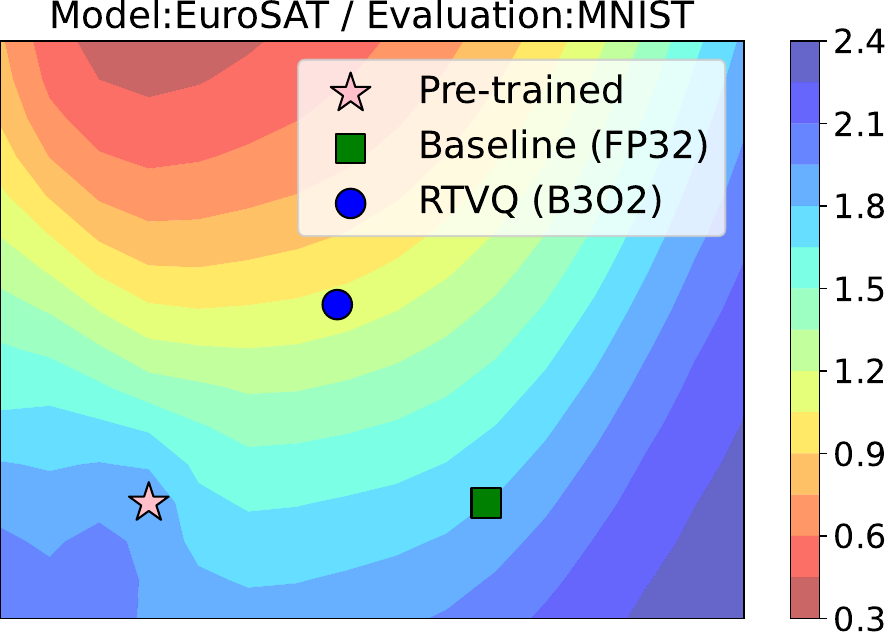}} &
        \subfloat{\includegraphics[width=0.23\linewidth]{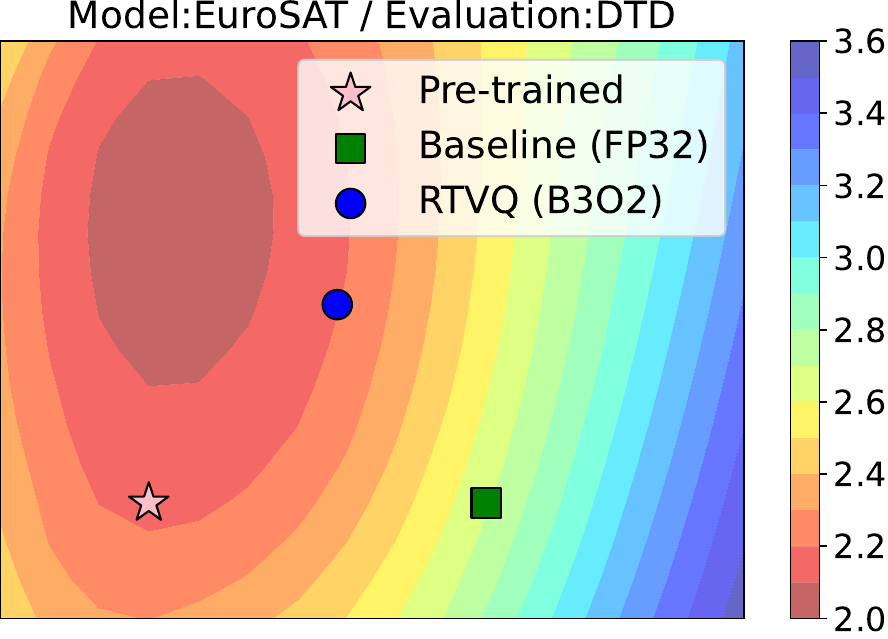}} \\
        
        \subfloat{\includegraphics[width=0.23\linewidth]{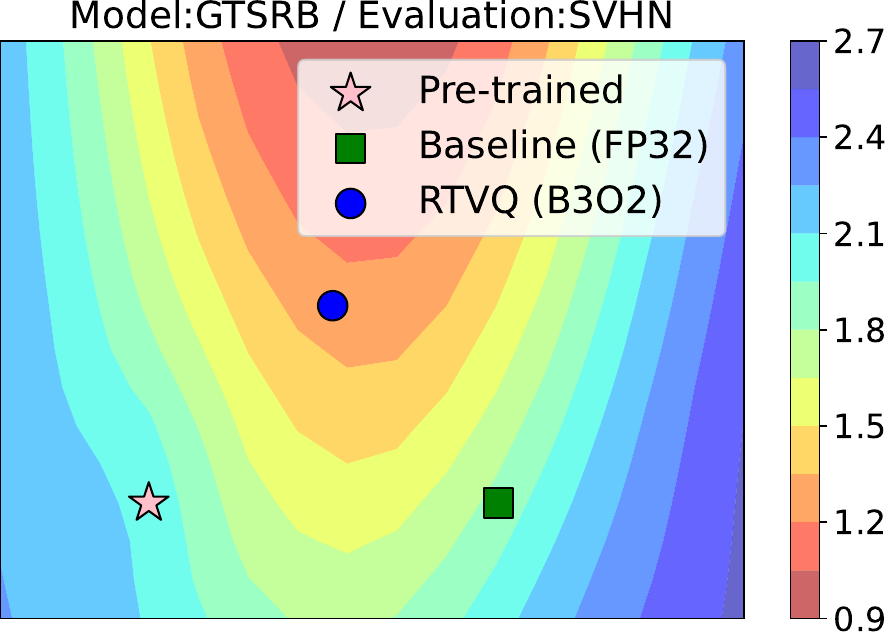}} &
        \subfloat{\includegraphics[width=0.23\linewidth]{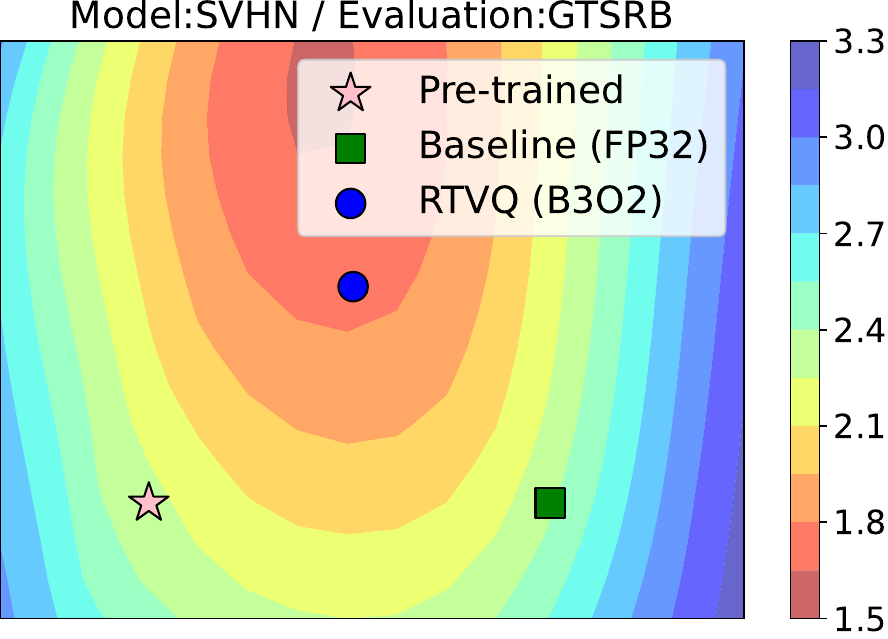}} &
        \subfloat{\includegraphics[width=0.23\linewidth]{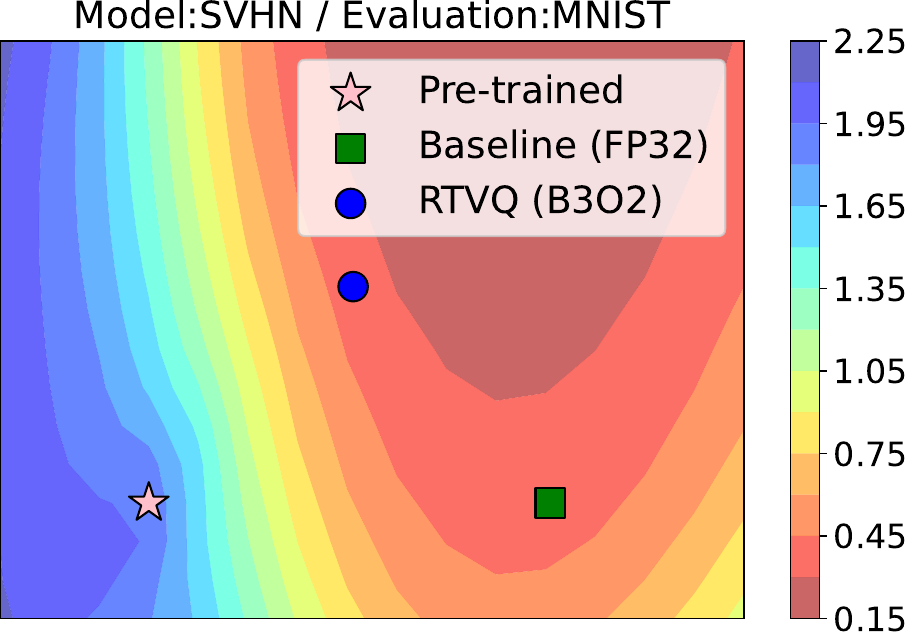}} &
        \subfloat{\includegraphics[width=0.23\linewidth]{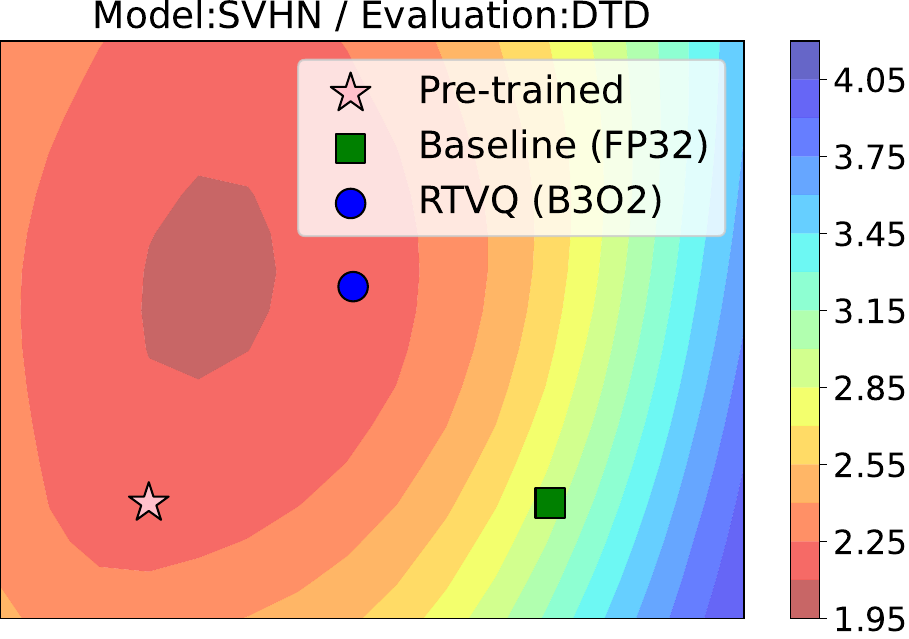}} \\
        
        \subfloat{\includegraphics[width=0.23\linewidth]{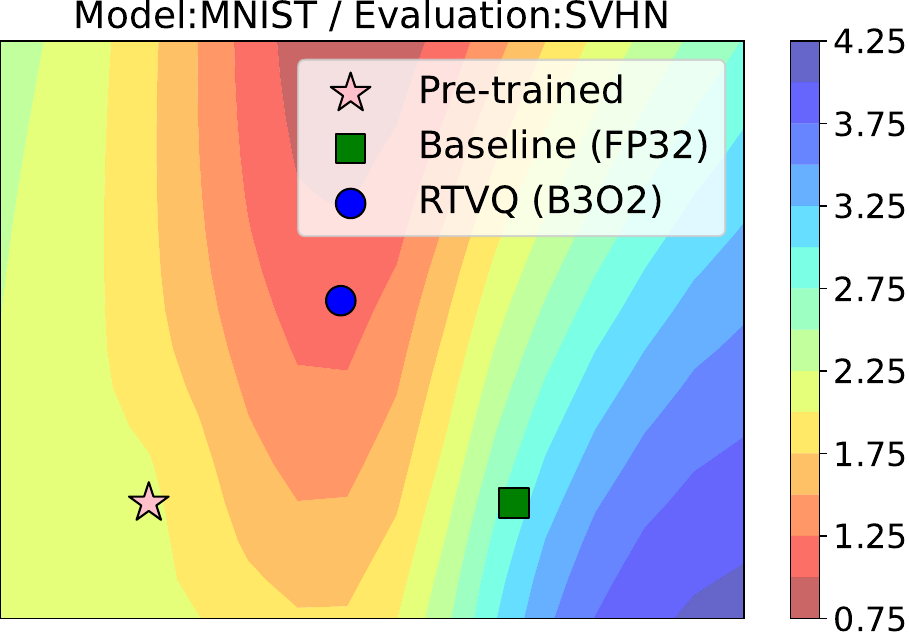}} &
        \subfloat{\includegraphics[width=0.23\linewidth]{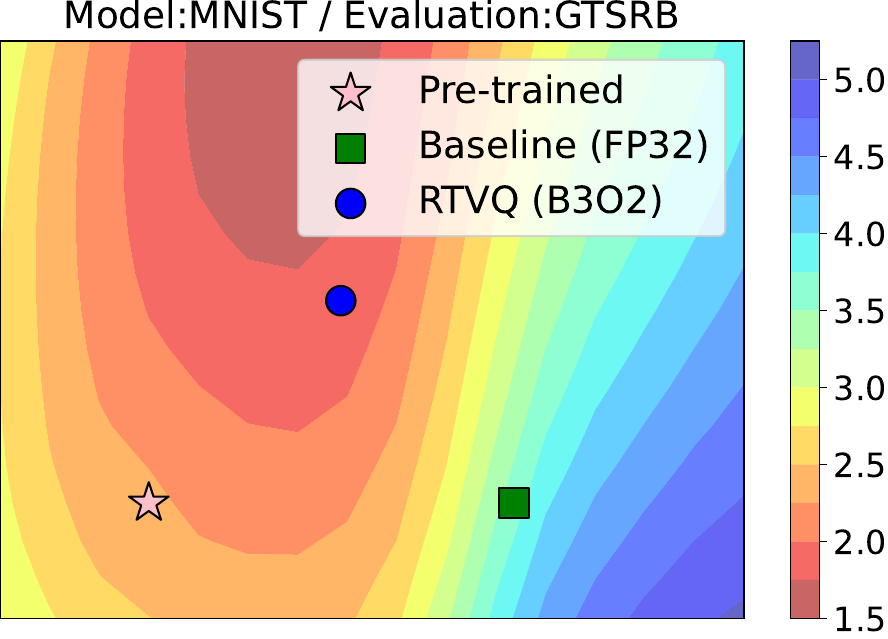}} &
        \subfloat{\includegraphics[width=0.23\linewidth]{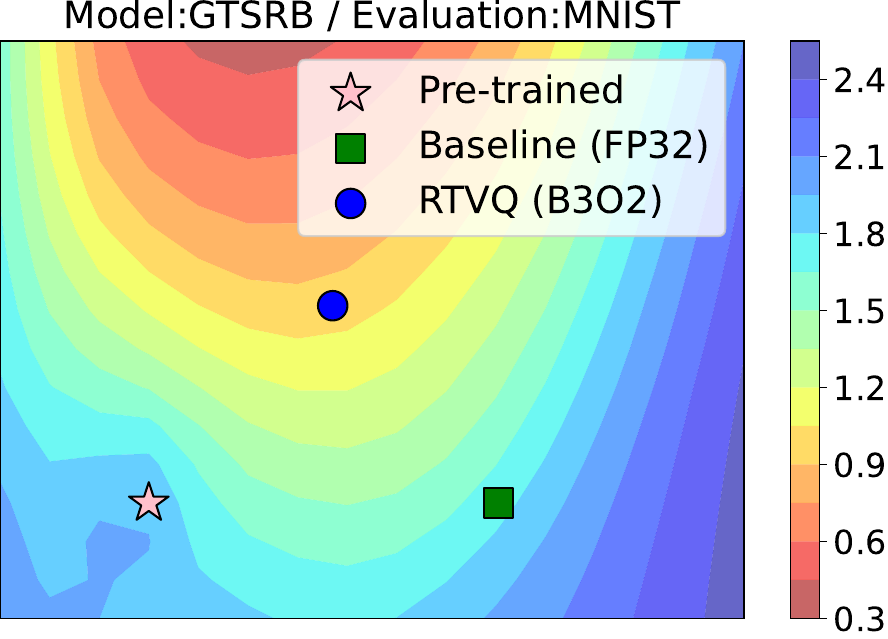}} &
        \subfloat{\includegraphics[width=0.23\linewidth]{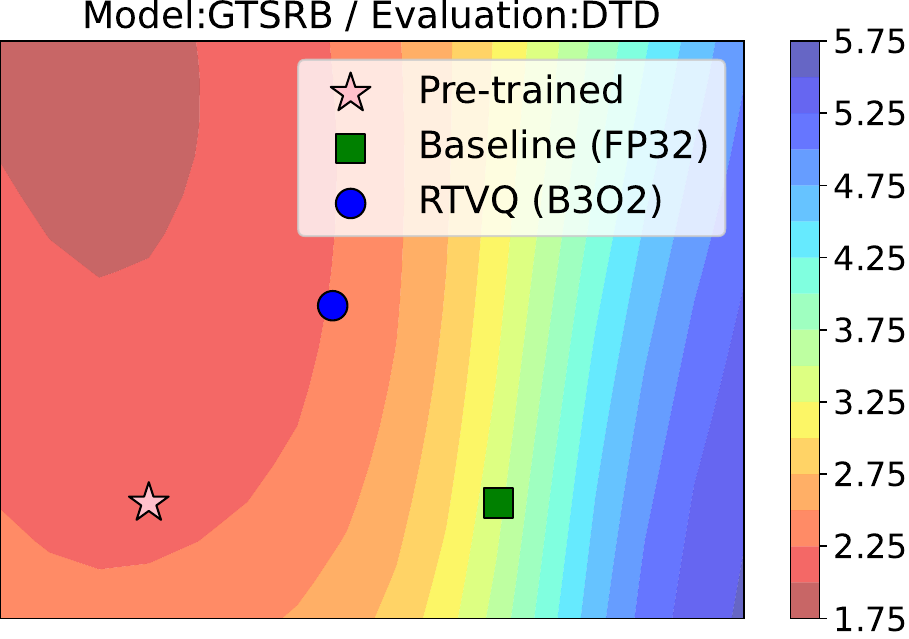}} \\
        
        \subfloat{\includegraphics[width=0.23\linewidth]{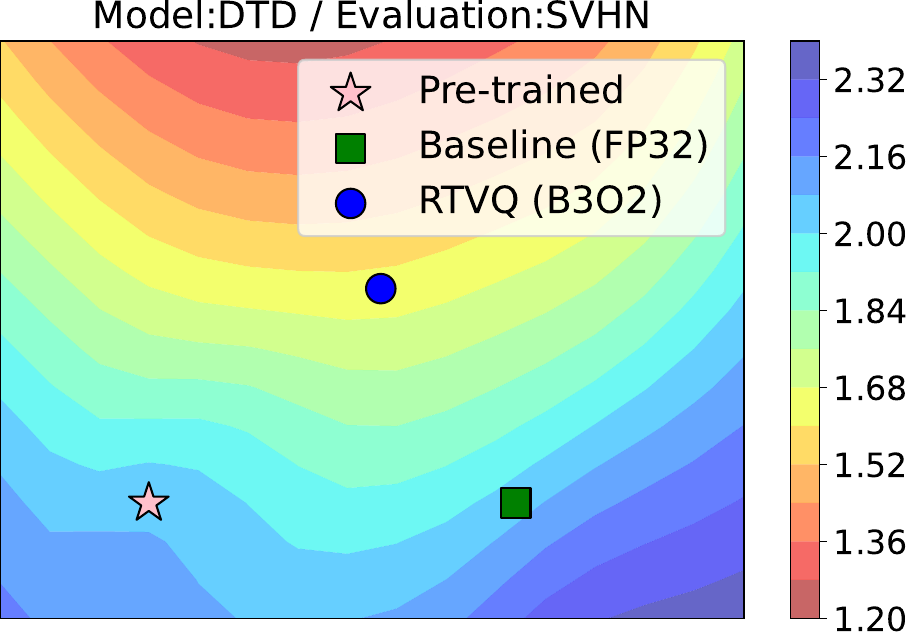}} &
        \subfloat{\includegraphics[width=0.23\linewidth]{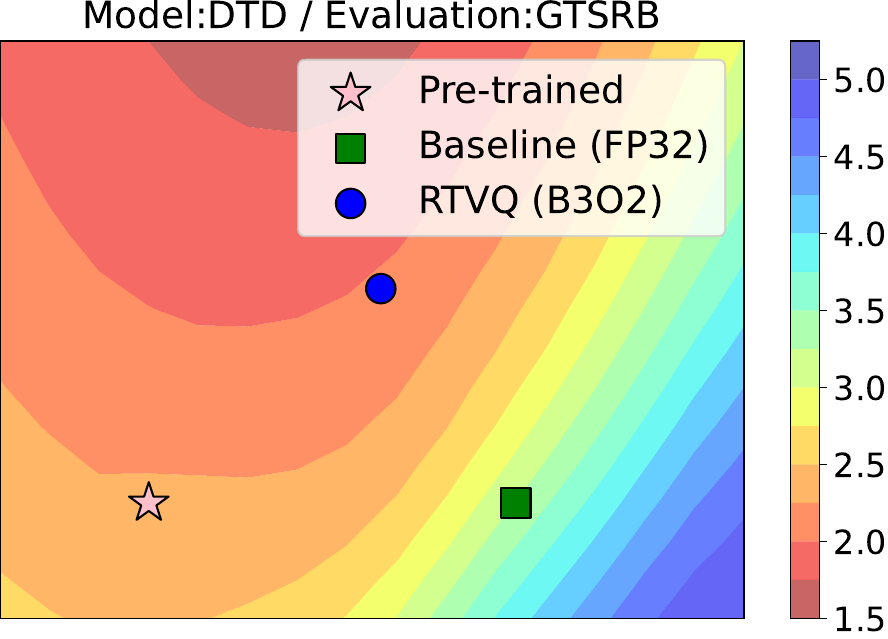}} &
        \subfloat{\includegraphics[width=0.23\linewidth]{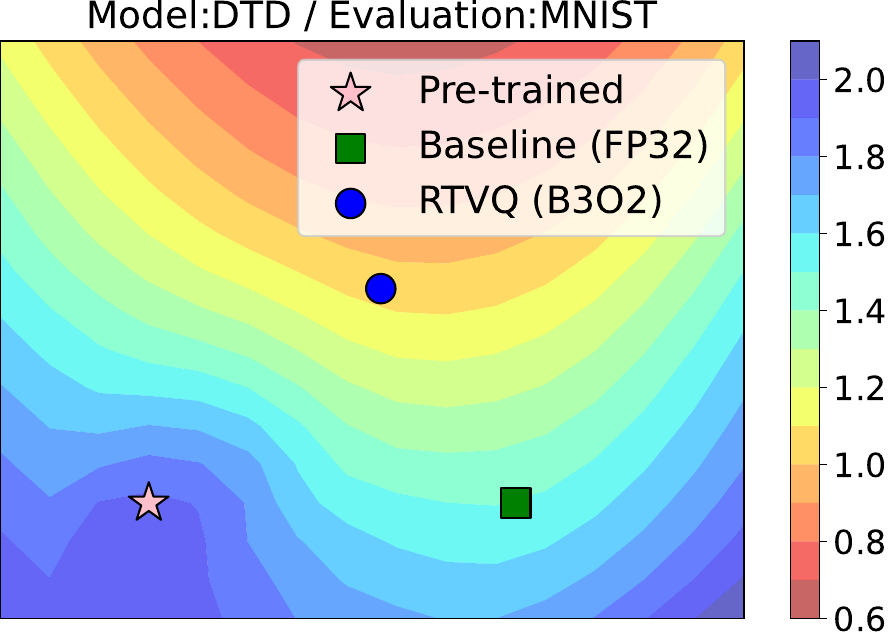}} &
        \subfloat{\includegraphics[width=0.23\linewidth]{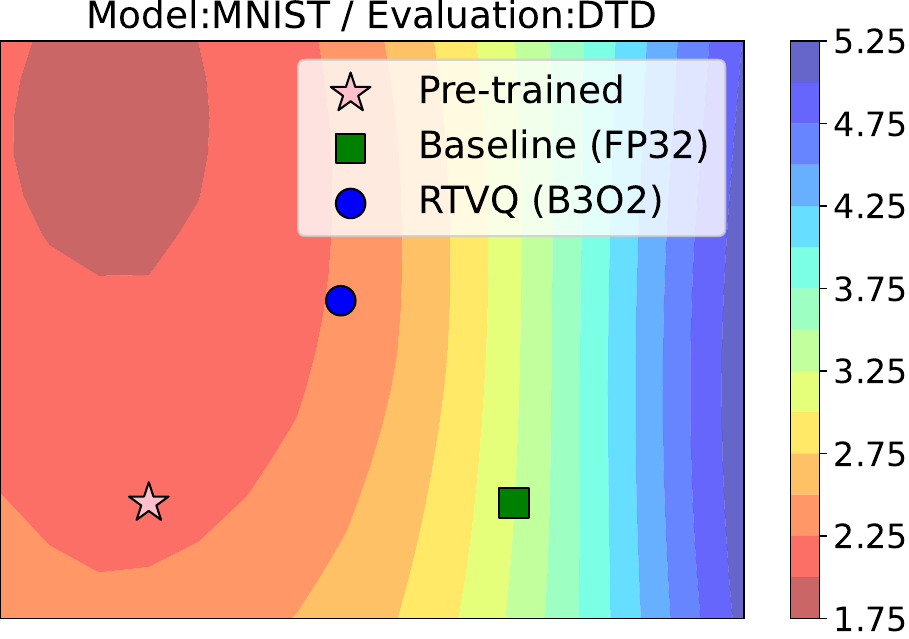}} \\

    \end{tabular}
    \caption{Loss landscape visualization of cross task pairs for RTVQ (B2O3). The results show evaluations on SVHN, GTSRB, MNIST, and DTD.}
    \label{fig:rtvq_landscape_control2}
\end{figure*}

\begin{figure*}[t]
    \centering
    \renewcommand{\thesubfigure}{} %
    \setlength{\tabcolsep}{1pt} %
    \renewcommand{\arraystretch}{0.8} %
    \begin{tabular}{cccc} %
        \subfloat{\includegraphics[width=0.23\linewidth]{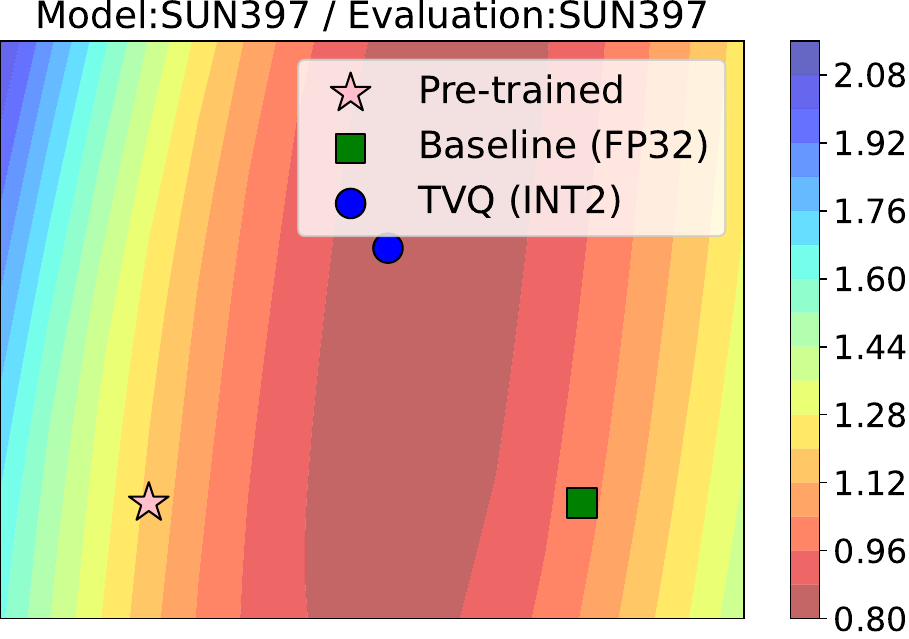}} &
        \subfloat{\includegraphics[width=0.23\linewidth]{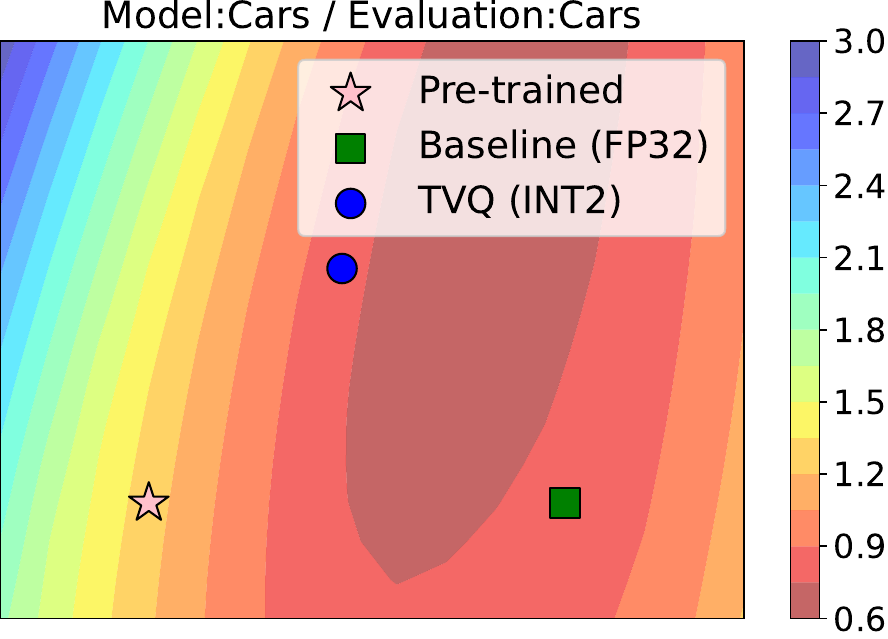}} &
        \subfloat{\includegraphics[width=0.23\linewidth]{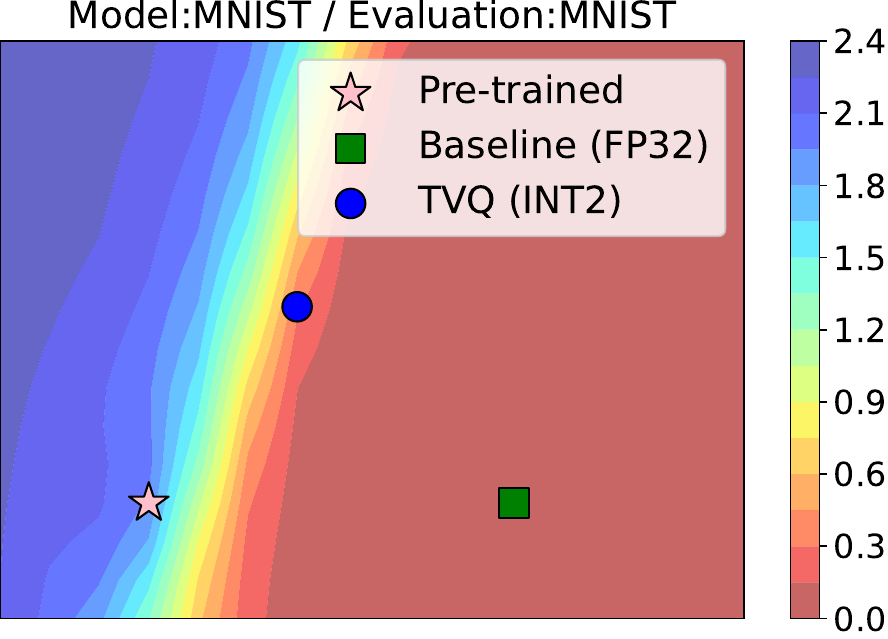}} &
        \subfloat{\includegraphics[width=0.23\linewidth]{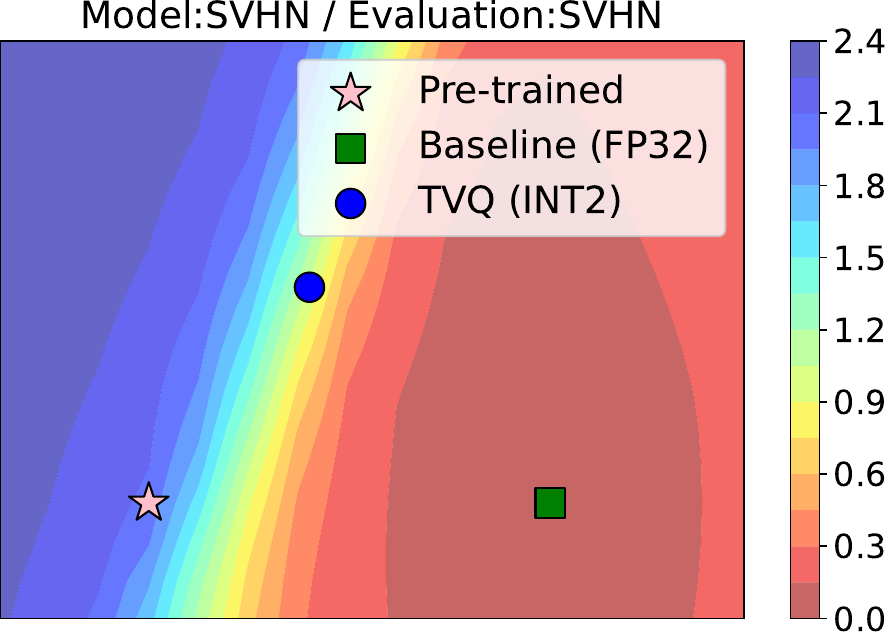}} 
        \\
        \subfloat{\includegraphics[width=0.23\linewidth]{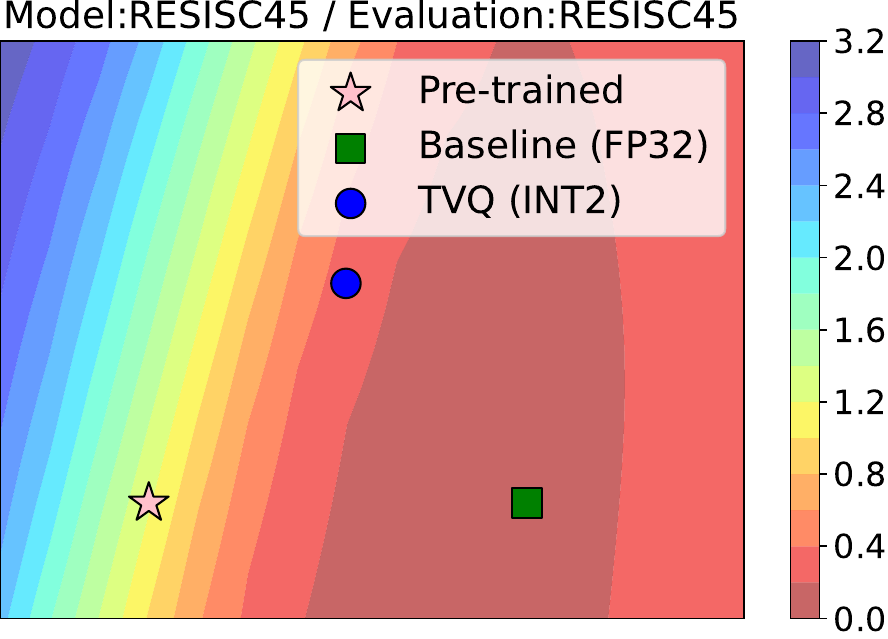}} &
        \subfloat{\includegraphics[width=0.23\linewidth]{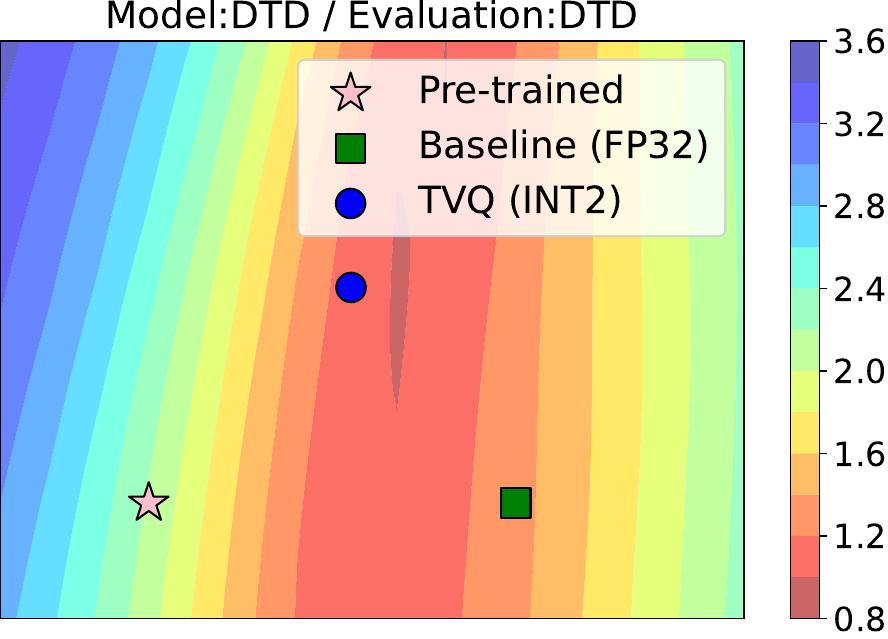}} &
        \subfloat{\includegraphics[width=0.23\linewidth]{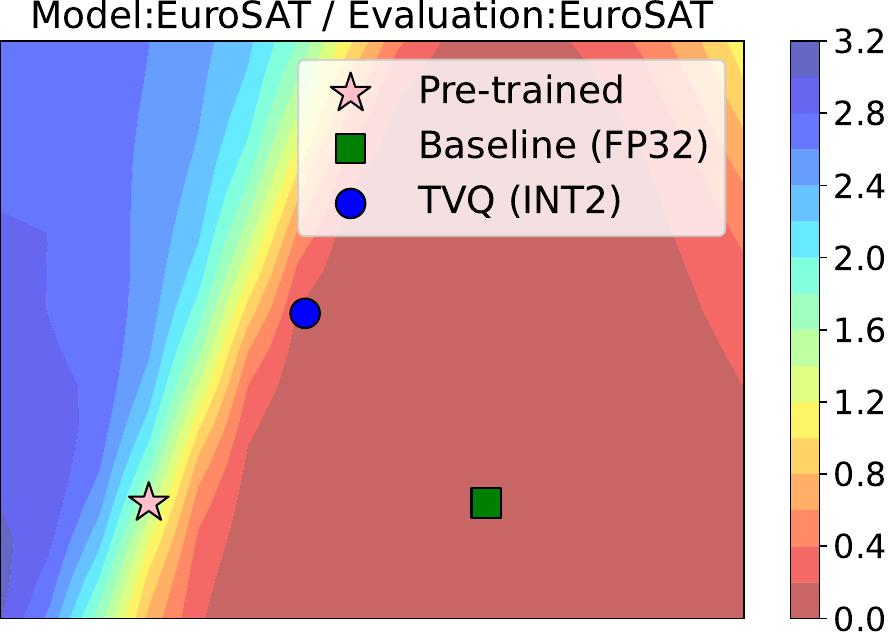}} &
        \subfloat{\includegraphics[width=0.23\linewidth]{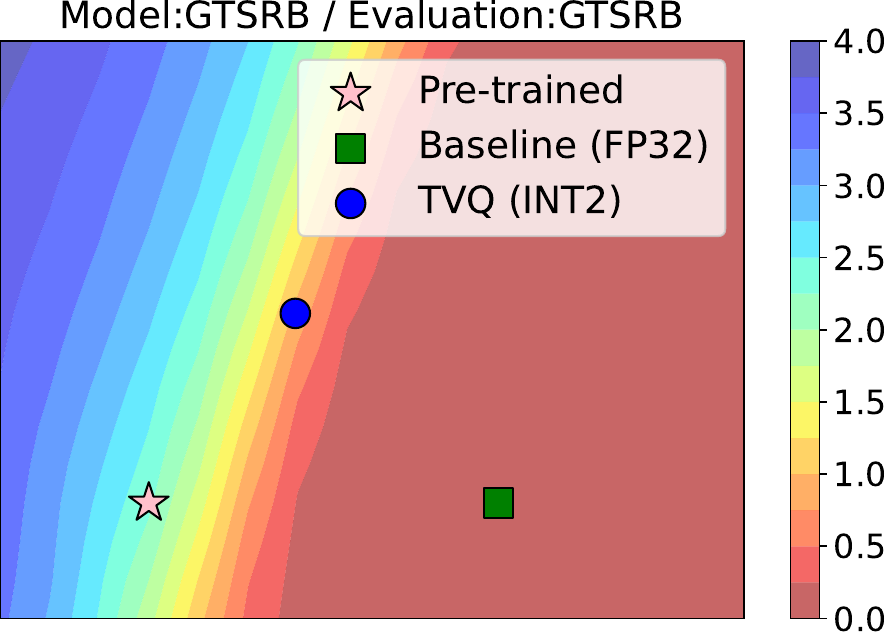}} \\

    \end{tabular}
    \caption{Loss landscape visualization of all target task pairs for 2-bit TVQ.}
    \label{fig:tvq_target_landscape}
\end{figure*}

\begin{figure*}[t]
    \centering
    \renewcommand{\thesubfigure}{} %
    \setlength{\tabcolsep}{1pt} %
    \renewcommand{\arraystretch}{0.8} %
    \begin{tabular}{cccc} %
        \subfloat{\includegraphics[width=0.23\linewidth]{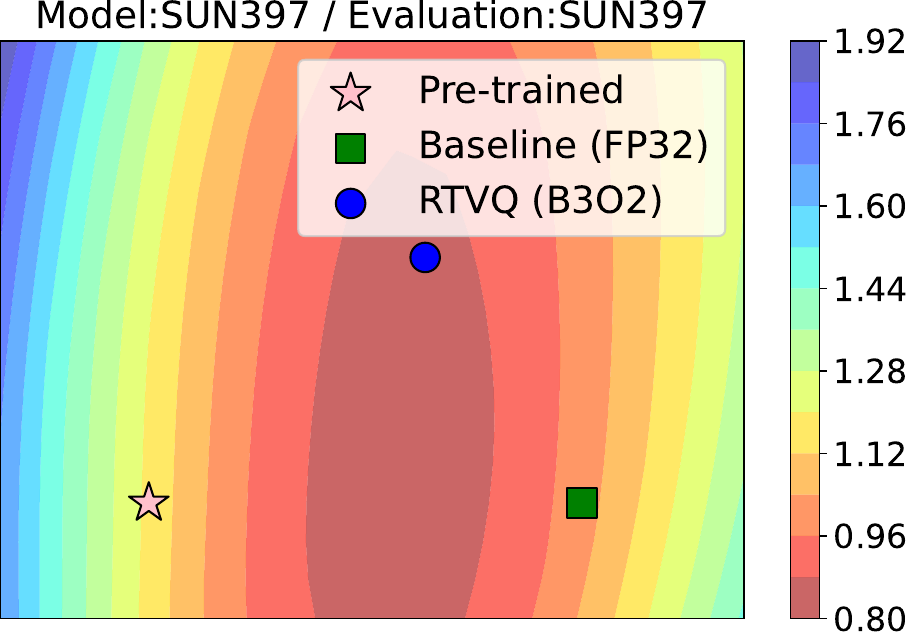}} &
        \subfloat{\includegraphics[width=0.23\linewidth]{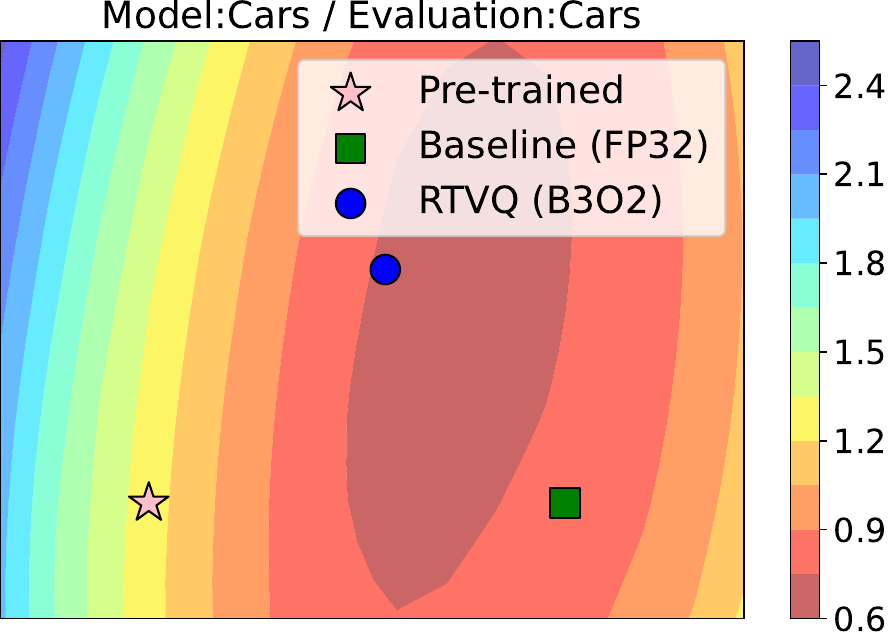}} &
        \subfloat{\includegraphics[width=0.23\linewidth]{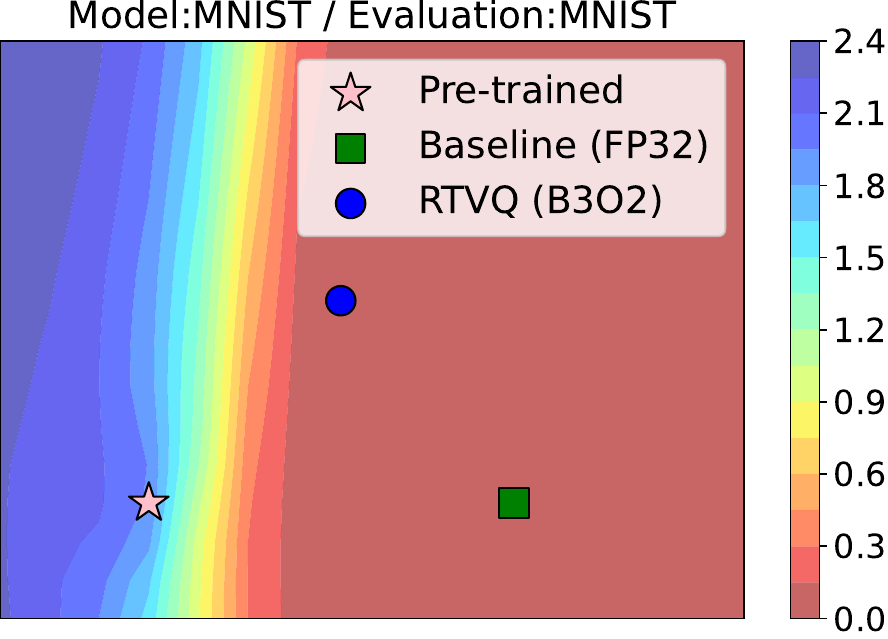}} &
        \subfloat{\includegraphics[width=0.23\linewidth]{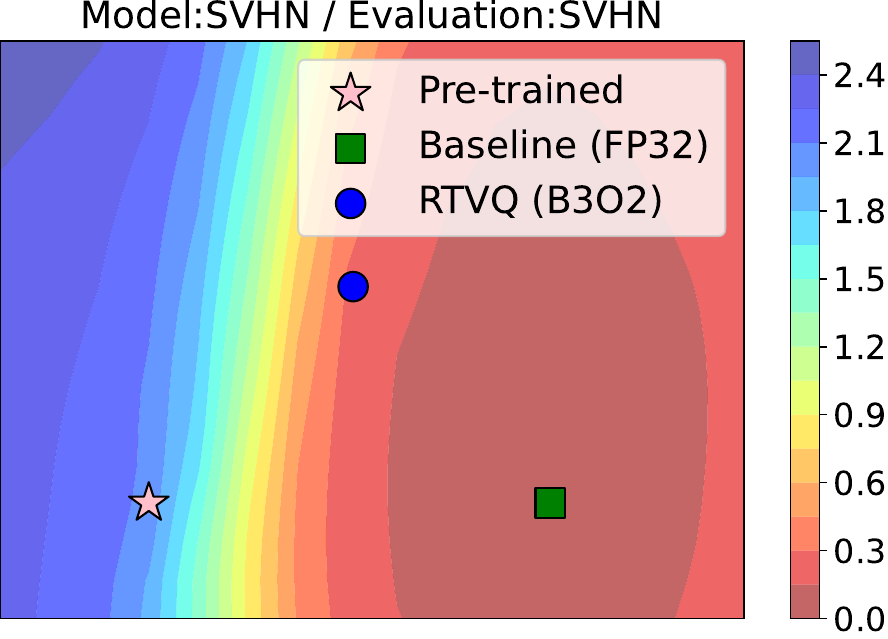}} 
        \\
        \subfloat{\includegraphics[width=0.23\linewidth]{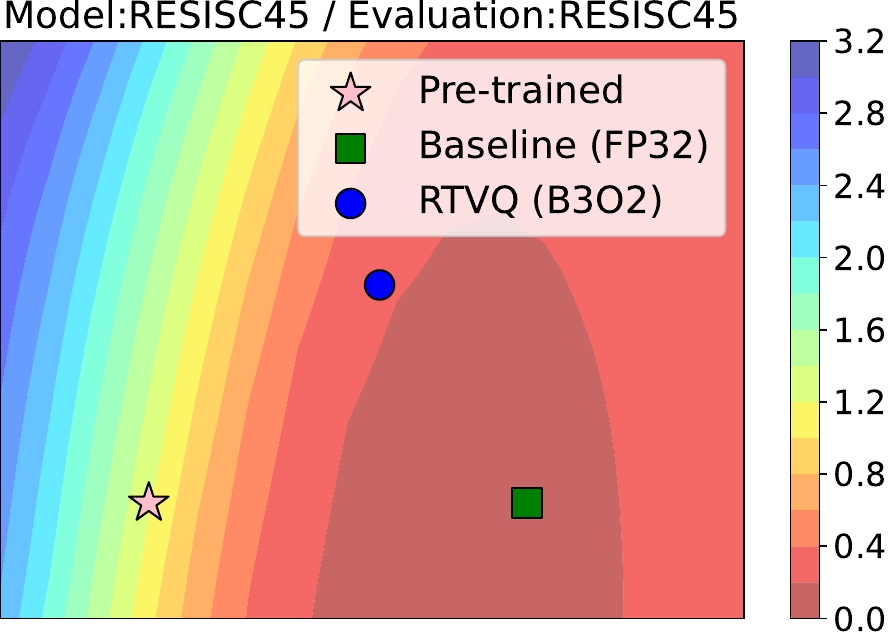}} &
        \subfloat{\includegraphics[width=0.23\linewidth]{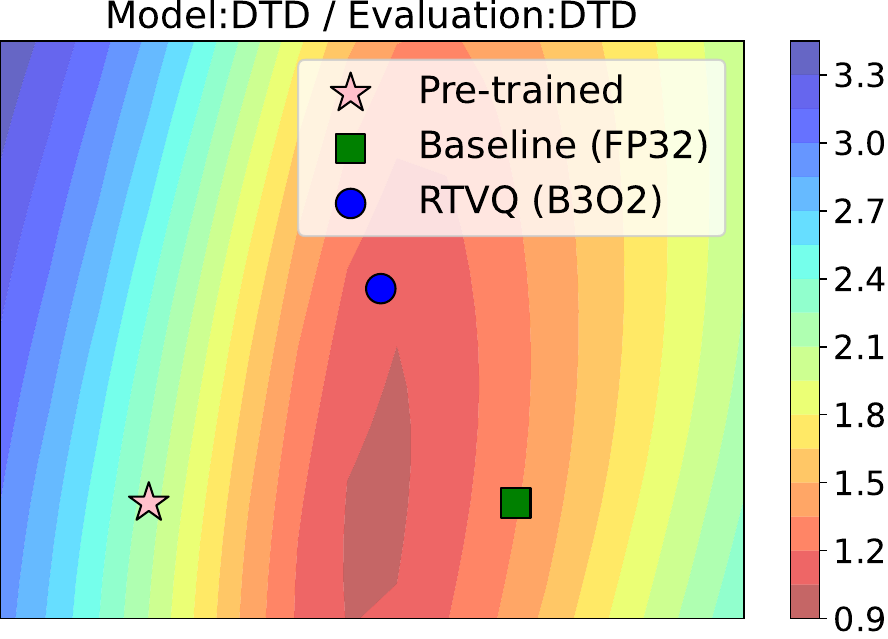}} &
        \subfloat{\includegraphics[width=0.23\linewidth]{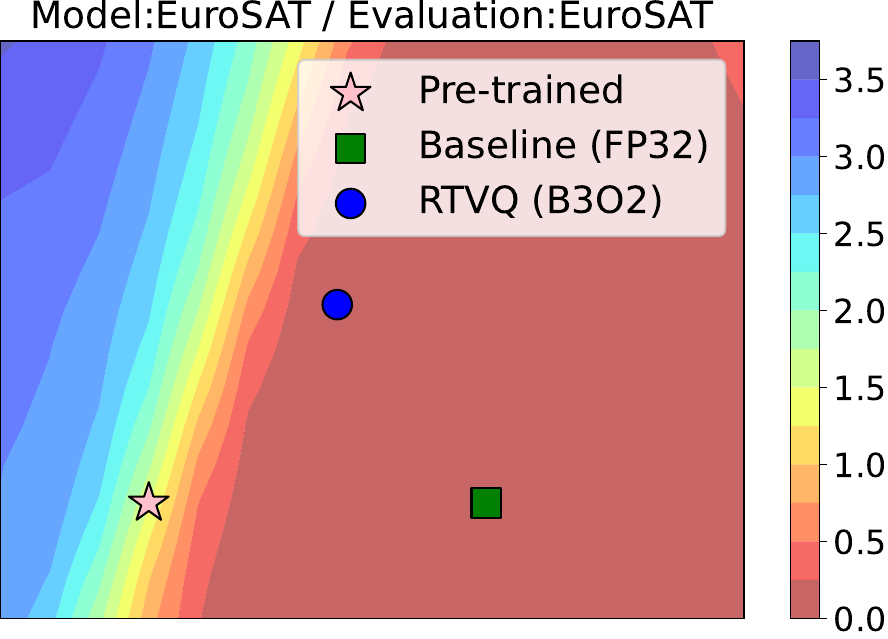}} &
        \subfloat{\includegraphics[width=0.23\linewidth]{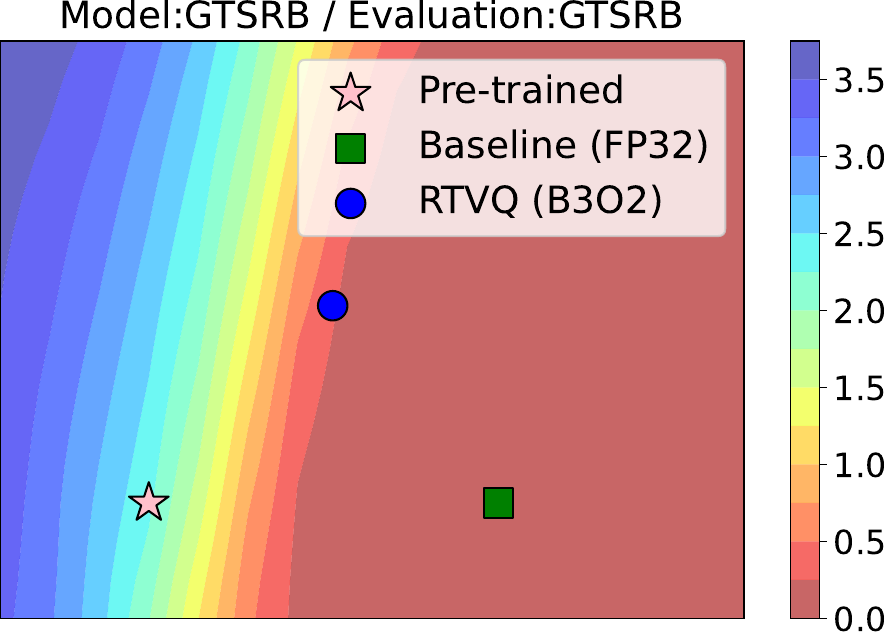}} \\

    \end{tabular}
    \caption{Loss landscape visualization of all target task pairs for RTVQ (B2O3).}
    \label{fig:rtvq_target_landscape}
\end{figure*}

We utilized the visualization method introduced in~\cite{mode_connectivity_fge_2018_NeurIPS}. Using 1,024 test images, we computed the loss values across all 16$\times$16 grid points for each task. For 8 vision tasks, we visualized the loss landscapes of all target task and cross-task pairs using ViT-B/32. The visualization results for cross-tasks can be found in Fig.~\ref{fig:tvq_landscape_control1},~\ref{fig:tvq_landscape_control2},~\ref{fig:rtvq_landscape_control1},~\ref{fig:rtvq_landscape_control2}, while the visualization results for target tasks are presented in Fig.~\ref{fig:tvq_target_landscape},~\ref{fig:rtvq_target_landscape}.

\begin{table*}[t]
    \centering
    \footnotesize 
    \begin{tabular}{l|c|ll|llll|l}
        \toprule
        \multirow{2}{*}{Method} & \multicolumn{1}{c|}{Baseline} & \multicolumn{2}{c|}{FQ} & \multicolumn{4}{c|}{TVQ  (ours)} & \multicolumn{1}{c}{\multirow{2}{*}{RTVQ  (ours)}} \\
        \cmidrule(lr){2-2} \cmidrule(lr){3-4} \cmidrule(lr){5-8} 
        & \multicolumn{1}{c|}{FP32} & \multicolumn{1}{c}{INT8} & \multicolumn{1}{c|}{INT4} & \multicolumn{1}{c}{INT8} & \multicolumn{1}{c}{INT4} & \multicolumn{1}{c}{INT3} & \multicolumn{1}{c|}{INT2} & \\
        \midrule
        Task arithmetic~\cite{TaskArithmetic_2023_ICLR} & 65.4 & 65.2 \perf{-0.2}  & 8.7 \perf{-56.7}  & 65.4 \perf{0.0} & 65.3 \perf{-0.1} & 65.1 \perf{-0.3} & 60.9 \perf{-4.5} & 65.0 \perf{-0.4} \\
        
        Ties merging~\cite{TiesMerging_2023_NeurIPS} & 65.2 & 63.5 \perf{-1.7} & 9.0 \perf{-56.2}  & 65.2 \perf{0.0} & 65.0 \perf{-0.2} & 66.4 \perf{1.2} & 61.3 \perf{-3.9} & 63.2 \perf{-2.0} \\

        LiNeS~\cite{LiNeS_2025_ICLR} & 68.0 & 67.9 \perf{-0.1} & 8.1 \perf{-59.9} & 68.0 \perf{0.0} & 68.0 \perf{0.0} & 67.8 \perf{-0.2} & 62.0 \perf{-6.0} & 67.5 \perf{-0.5} \\
        
        Consensus TA~\cite{Tall_Mask_2024_ICML} & 70.2 & 64.4 \perf{-5.8}  & 8.4 \perf{-61.8} & 70.2 \perf{0.0} & 70.1 \perf{-0.1} & 70.2 \perf{0.0} & 63.2 \perf{-7.0} & 69.8 \perf{-0.4}\\
        
        AdaMerging~\cite{Adamerging_2024_ICLR} & 76.7 & 76.2 \perf{-0.5}  & 8.2 \perf{-68.5} & 76.7 \perf{0.0} & 76.8 \perf{0.1} & 77.2 \perf{0.5} & 74.4 \perf{-2.3}  & 76.1 \perf{-0.6} \\
        
        EMR-Merging~\cite{EMRMerging_2024} & 86.1 & 86.2 \perf{0.1} & 7.7 \perf{-78.4} & 86.3 \perf{0.2} & 88.2 \perf{2.1} & 88.4 \perf{2.3} & 76.2 \perf{-9.9}  & 78.9 \perf{-7.2} \\
                
        \bottomrule
    \end{tabular}
    
    \caption{Comparison of our proposed quantization methods for merging 14 classification tasks using ViT-B/32. Note that our primary objective is to improve storage efficiency for checkpoint saving while minimizing performance degradation relative to the baseline, rather than to maximize performance. The baseline refers to full-precision (FP32) model checkpoints. FQ denotes quantizing fine-tuned checkpoints, TVQ indicates Task Vector Quantization, and RTVQ incorporates our proposed Residual Task Vector Quantization using a 3-bit base vector and a 2-bit offset vector (equivalent to 2.21 bits per task). We report the average accuracy (\%) across all tasks, with the difference relative to FP32 shown in parentheses (\textcolor{red}{red} indicates a performance drop, \textcolor{ForestGreen}{green} a gain).}
    \label{tab:supple_vitb32_14_tasks}
\end{table*}

\begin{table*}[t]
    \centering
    \footnotesize 
    \begin{tabular}{l|c|ll|llll|l}
        \toprule
        \multirow{2}{*}{Method} & \multicolumn{1}{c|}{Baseline} & \multicolumn{2}{c|}{FQ} & \multicolumn{4}{c|}{TVQ  (ours)} & \multicolumn{1}{c}{\multirow{2}{*}{RTVQ  (ours)}} \\
        \cmidrule(lr){2-2} \cmidrule(lr){3-4} \cmidrule(lr){5-8} 
        & \multicolumn{1}{c|}{FP32} & \multicolumn{1}{c}{INT8} & \multicolumn{1}{c|}{INT4} & \multicolumn{1}{c}{INT8} & \multicolumn{1}{c}{INT4} & \multicolumn{1}{c}{INT3} & \multicolumn{1}{c|}{INT2} & \\
        \midrule
        Task arithmetic~\cite{TaskArithmetic_2023_ICLR}& 60.8 & 60.5 \perf{-0.3}& 10.4 \perf{-50.4} & 60.8 \perf{0.0} & 60.8 \perf{0.0}& 61.3 \perf{0.5}& 59.5 \perf{-1.3}& 61.0 \perf{0.2} \\
        
        Ties merging~\cite{TiesMerging_2023_NeurIPS} & 63.1 & 59.0 \perf{-4.1} & 10.4 \perf{-52.7} & 63.1 \perf{0.0} & 62.9 \perf{-0.2} & 64.1 \perf{1.0} & 60.0 \perf{-3.1} & 58.9 \perf{-4.2} \\

        LiNeS~\cite{LiNeS_2025_ICLR}& 63.7 & 63.5 \perf{-0.2}& 10.3 \perf{-53.4}& 63.7 \perf{0.0}& 63.7 \perf{0.0}& 64.1 \perf{0.4}& 60.3 \perf{-3.4}& 63.7 \perf{0.0}\\
        
        Consensus TA~\cite{Tall_Mask_2024_ICML}& 65.0 & 59.3 \perf{-5.7}& 10.5 \perf{-54.5}& 65.0 \perf{0.0}& 65.0 \perf{0.0}& 65.5 \perf{0.5}& 60.7 \perf{-4.3}& 65.7 \perf{0.7}\\
        
        AdaMerging~\cite{Adamerging_2024_ICLR}& 69.6 & 69.3 \perf{-0.3}& \phantom{0}9.7 \perf{-59.9}& 71.3 \perf{1.7}& 71.3 \perf{1.7}& 71.8 \perf{2.2}& 70.2 \perf{0.6}& 71.5 \perf{1.9}\\
        
        EMR-Merging~\cite{EMRMerging_2024}& 86.6 & 84.3 \perf{-2.3}& 10.0 \perf{-76.6}& 86.7 \perf{0.1}& 88.4 \perf{1.8}& 87.1 \perf{0.5}& 72.7 \perf{-13.9}& 75.6 \perf{-11.0}\\
                
        \bottomrule
    \end{tabular}
    
    \caption{Comparison of our proposed quantization methods for merging 20 classification tasks using ViT-B/32. For RTVQ, we use a 3-bit base vector and a 2-bit offset vector (equivalent to 2.15 bits per task)}
    \label{tab:supple_vitb32_20_tasks}
\end{table*}

\begin{table*}[tbh!]
    \footnotesize
    \centering
    \begin{tabular}{llllllllll} 
    \toprule
    \multicolumn{2}{c}{\textbf{}} & \multicolumn{2}{c}{Segmentation} &  & \multicolumn{2}{c}{Depth} &  & Normal \\
    \cmidrule(lr){3-4} \cmidrule(lr){6-7} \cmidrule(lr){9-9}
    & & mIoU $\uparrow$ & Pix Acc $\uparrow$ &  & Abs Err $\downarrow$ & Rel Err $\downarrow$ &  & Mean $\downarrow$ \\
    \midrule
    \textbf{\multirow{8}{*}{Individual}}
    &FP32 & $52.02$ & $74.15$ &  & $41.45$ & $17.28$ &  & $24.24$\\
    & FQ8  & $51.92$ \perf{-0.10} & $73.99$ \perf{-0.16} & & $41.70$ \revperf{0.25} & $17.52$ \revperf{0.24} & & $24.34$ \revperf{0.10} \\
    & FQ4  & $\phantom{0}1.20$ \perf{-50.82} & $15.55$ \perf{-58.60} & & $89.48$ \revperf{48.03} & $36.00$ \revperf{18.72} & & $44.24$ \revperf{20.00} \\
    & INT8  & $52.02$ \perf{0.00} & $74.15$ \perf{0.00} & & $41.45$ \revperf{0.00} & $17.28$ \revperf{0.00} & & $24.24$ \revperf{0.00} \\
    & INT4  & $51.98$ \perf{-0.04} & $74.16$ \perf{0.01} & & $41.44$ \revperf{-0.01} & $17.35$ \revperf{0.07} & & $24.22$ \revperf{-0.02} \\
    & INT3  & $51.78$ \perf{-0.24} & $74.05$ \perf{-0.10} & & $41.57$ \revperf{0.12} & $17.35$ \revperf{0.07} & & $24.49$ \revperf{0.25} \\
    & INT2  & $37.67$ \perf{-14.35} & $56.25$ \perf{-17.90} & & $62.46$ \revperf{21.01} & $24.14$ \revperf{6.86} & & $34.17$ \revperf{9.93} \\
        
    \midrule
    
    \textbf{\multirow{8}{*}{Task Arithmetic ~\cite{TaskArithmetic_2023_ICLR}}}
    & FP32 & $31.60$ & $60.31$ &  & $56.67$ & $24.03$ &  & $30.62$\\
    & FQ8  & $31.68$ \perf{0.08} & $60.36$ \perf{0.05} & & $56.79$ \revperf{0.12} & $24.01$ \revperf{-0.02} & & $30.63$ \revperf{0.01} \\
    & FQ4  & $\phantom{0}6.11$ \perf{-25.49} & $23.06$ \perf{-37.25} & & $79.52$ \revperf{22.85} & $33.69$ \revperf{9.66} & & $43.67$ \revperf{13.05} \\
    & INT8  & $31.61$ \perf{0.01} & $60.32$ \perf{0.01} & & $56.67$ \revperf{0.00} & $24.03$ \revperf{0.00} & & $30.62$ \revperf{0.00} \\
    & INT4  & $31.54$ \perf{-0.06} & $60.32$ \perf{0.01} & & $56.65$ \revperf{-0.02} & $24.04$ \revperf{0.01} & & $30.62$ \revperf{0.00} \\
    & INT3  & $32.11$ \perf{0.51} & $60.81$ \perf{0.50} & & $56.78$ \revperf{0.11} & $23.98$ \revperf{-0.05} & & $30.72$ \revperf{0.10} \\
    & INT2  & $36.36$ \perf{4.76} & $61.32$ \perf{1.01} & & $63.88$ \revperf{7.21} & $26.21$ \revperf{2.18} & & $36.07$ \revperf{5.45} \\
    & RTVQ  & $36.13$ \perf{4.53} & $59.23$ \perf{-1.08} & & $59.23$ \revperf{2.56} & $24.63$ \revperf{0.60} & & $32.60$ \revperf{1.98} \\
        
    \midrule
    
    \textbf{\multirow{8}{*}{Ties-Merging ~\cite{TiesMerging_2023_NeurIPS}}}
    &FP32 & $39.91$ & $62.70$ &  & $61.25$ & $27.28$ &  & $36.17$\\
    & FQ8  & $40.14$ \perf{0.23} & $63.27$ \perf{0.57} & & $61.04$ \revperf{-0.21} & $27.29$ \revperf{0.01} & & $36.19$ \revperf{0.02} \\
    & FQ4  & $\phantom{0}9.52$ \perf{-30.39} & $25.05$ \perf{-37.65} & & $77.46$ \revperf{16.21} & $30.45$ \revperf{3.17} & & $68.94$ \revperf{32.77} \\
    & INT8  & $39.90$ \perf{-0.01} & $62.78$ \perf{0.08} & & $61.21$ \revperf{-0.04} & $27.26$ \revperf{-0.02} & & $36.16$ \revperf{-0.01} \\
    & INT4  & $39.98$ \perf{0.07} & $63.17$ \perf{0.47} & & $61.04$ \revperf{-0.21} & $27.21$ \revperf{-0.07} & & $36.20$ \revperf{0.03} \\
    & INT3  & $39.23$ \perf{-0.68} & $62.02$ \perf{-0.68} & & $61.54$ \revperf{0.29} & $27.26$ \revperf{-0.02} & & $36.37$ \revperf{0.20} \\
    & INT2  & $36.09$ \perf{-3.82} & $59.92$ \perf{-2.78} & & $65.10$ \revperf{3.85} & $26.54$ \revperf{-0.74} & & $37.03$ \revperf{0.86} \\
    & RTVQ  & $37.02$ \perf{-2.89} & $63.57$ \perf{0.87} & & $59.22$ \revperf{-2.03} & $24.55$ \revperf{-2.73} & & $32.64$ \revperf{-3.53} \\
    
    \midrule
    
    \textbf{\multirow{8}{*}{MagMax ~\cite{MagMax_2024_ECCV}}}
    &FP32& $24.73$ & $54.71$ &  & $60.27$ & $23.88$ &  & $30.25$\\
    & FQ8  & $24.80$ \perf{0.07} & $54.75$ \perf{0.04} & & $61.18$ \revperf{0.91} & $24.17$ \revperf{0.29} & & $30.44$ \revperf{0.19} \\
    & FQ4  & $\phantom{0}6.16$ \perf{-18.57} & $24.52$ \perf{-30.19} & & $79.24$ \revperf{18.97} & $31.60$ \revperf{7.72} & & $39.97$ \revperf{9.72} \\
    & INT8  & $24.76$ \perf{0.03} & $54.75$ \perf{0.04} & & $60.25$ \revperf{-0.02} & $23.88$ \revperf{0.00} & & $30.26$ \revperf{0.01} \\
    & INT4  & $25.39$ \perf{0.66} & $55.19$ \perf{0.48} & & $61.33$ \revperf{1.06} & $24.19$ \revperf{0.31} & & $30.04$ \revperf{-0.21} \\
    & INT3  & $23.28$ \perf{-1.45} & $52.88$ \perf{-1.83} & & $62.87$ \revperf{2.60} & $24.59$ \revperf{0.71} & & $29.77$ \revperf{-0.48} \\
    & INT2  & $29.93$ \perf{5.20} & $58.65$ \perf{3.94} & & $64.29$ \revperf{4.02} & $25.59$ \revperf{1.71} & & $32.22$ \revperf{1.97} \\
    & RTVQ  & $29.39$ \perf{4.66} & $58.48$ \perf{3.77} & & $62.57$ \revperf{2.30} & $24.73$ \revperf{0.85} & & $31.06$ \revperf{0.81} \\
    
    \midrule
    
    \textbf{\multirow{8}{*}{Breadcrumbs ~\cite{Breadcrumbs_2024_ECCV}}}
    &FP32 & $34.14$ & $58.51$ &  & $66.05$ & $27.17$ &  & $36.85$\\
    & FQ8  & $34.29$ \perf{0.15} & $58.80$ \perf{0.29} & & $65.97$ \revperf{-0.08} & $27.15$ \revperf{-0.02} & & $36.84$ \revperf{-0.01} \\
    & FQ4  & $19.73$ \perf{-14.41} & $40.63$ \perf{-17.88} & & $75.86$ \revperf{9.81} & $29.45$ \revperf{2.28} & & $40.60$ \revperf{3.75} \\
    & INT8  & $34.19$ \perf{0.05} & $58.57$ \perf{0.06} & & $66.06$ \revperf{0.01} & $27.17$ \revperf{0.00} & & $36.88$ \revperf{0.03} \\
    & INT4  & $34.26$ \perf{0.12} & $58.56$ \perf{0.05} & & $66.07$ \revperf{0.02} & $27.18$ \revperf{0.01} & & $37.00$ \revperf{0.15} \\
    & INT3  & $34.30$ \perf{0.16} & $58.70$ \perf{0.19} & & $66.11$ \revperf{0.06} & $27.15$ \revperf{-0.02} & & $36.86$ \revperf{0.01} \\
    & INT2  & $32.19$ \perf{-1.95} & $54.22$ \perf{-4.29} & & $69.03$ \revperf{2.98} & $28.44$ \revperf{1.27} & & $40.58$ \revperf{3.73} \\
    & RTVQ  & $33.97$ \perf{-0.17} & $57.31$ \perf{-1.20} & & $67.13$ \revperf{1.08} & $27.66$ \revperf{0.49} & & $38.29$ \revperf{1.44} \\
    
    \midrule
    
    \textbf{\multirow{8}{*}{EMR-Merging ~\cite{EMRMerging_2024}}}
    &FP32& $41.50$ & $67.24$ &  & $48.59$ & $19.44$ &  & $26.52$\\
    & FQ8  & $41.73$ \perf{0.23} & $67.27$ \perf{0.03} & & $48.64$ \revperf{0.05} & $19.37$ \revperf{-0.07} & & $26.54$ \revperf{0.02} \\
    & FQ4  & $\phantom{0}1.20$ \perf{-40.30} & $15.55$ \perf{-51.69} & & $80.46$ \revperf{31.87} & $35.39$ \revperf{15.95} & & $45.78$ \revperf{19.26} \\
    & INT8  & $41.69$ \perf{0.19} & $67.37$ \perf{0.13} & & $48.43$ \revperf{-0.16} & $19.35$ \revperf{-0.09} & & $26.50$ \revperf{-0.02} \\
    & INT4  & $44.79$ \perf{3.29} & $69.53$ \perf{2.29} & & $47.13$ \revperf{-1.46} & $18.76$ \revperf{-0.68} & & $26.56$ \revperf{0.04} \\
    & INT3  & $43.00$ \perf{1.50} & $67.66$ \perf{0.42} & & $48.05$ \revperf{-0.54} & $18.87$ \revperf{-0.57} & & $29.21$ \revperf{2.69} \\
    & INT2  & $21.33$ \perf{-20.17} & $40.95$ \perf{-26.29} & & $68.31$ \revperf{19.72} & $25.45$ \revperf{6.01} & & $45.16$ \revperf{18.64} \\
    & RTVQ  & $34.11$ \perf{-7.39} & $57.23$ \perf{-10.01} & & $59.31$ \revperf{10.72} & $22.07$ \revperf{2.63} & & $34.99$ \revperf{8.47} \\
    \bottomrule
    \end{tabular}
    \caption{Comprehensive experimental results of merging ResNet-50 models on three NYUv2 tasks. We compare the proposed quantization methods against the full-precision baseline (FP32). RTVQ quantizes both the base vector and offset vector to 2 bits.}
    \label{tab:supple_dense}

\end{table*}

\definecolor{mygray}{gray}{.95}

\begin{table*}[tbh!]
\footnotesize 

\vspace{-5pt}
\centering
\resizebox{\textwidth}{!}{
\begin{tabular}{l|l | l l l l l l l l |l}
        \toprule
        \multicolumn{2}{c|}{\textbf{Method}} & \multicolumn{1}{c}{\textbf{SUN}} & \multicolumn{1}{c}{\textbf{Cars}} & \multicolumn{1}{c}{\textbf{RES.}} & \multicolumn{1}{c}{\textbf{Euro}} & \multicolumn{1}{c}{\textbf{SVH.}} & \multicolumn{1}{c}{\textbf{GTS.}} & \multicolumn{1}{c}{\textbf{MNI.}} & \multicolumn{1}{c|}{\textbf{DTD}} &\multicolumn{1}{c}{\textbf{Avg.}} \\
            \midrule
            & FP32  & 75.3 & 77.7 & 96.1 & 99.7 & 97.5 & 98.7 & 99.7 & 79.4 & 90.5 \\
            & FQ8  & 74.9 \perf{-0.4} & 77.2 \perf{-0.5} & 96.0 \perf{-0.1} & 99.9 \perf{0.2} & 97.4 \perf{-0.1} & 98.7 \perf{0.0} & 99.7 \perf{0.0} & 79.7 \perf{0.3} & 90.4 \perf{-0.1} \\
            & FQ4  & \phantom{0}0.3 \perf{-75.0} & \phantom{0}0.7 \perf{-77.0} & \phantom{0}3.2 \perf{-92.9} & \phantom{0}7.5 \perf{-92.2} & \phantom{0}8.9 \perf{-88.6} & \phantom{0}2.2 \perf{-96.5} & \phantom{0}8.7 \perf{-91.0} & \phantom{0}2.0 \perf{-77.4} & \phantom{0}4.2 \perf{-86.3} \\
            \multirow{2}{*}{\textbf{Individual}} & INT8  & 75.3 \perf{0.0} & 77.7 \perf{0.0} & 96.1 \perf{0.0} & 99.9 \perf{0.2} & 97.5 \perf{0.0} & 98.7 \perf{0.0} & 99.7 \perf{0.0} & 79.4 \perf{0.0} & 90.5 \perf{0.0} \\
            & INT4  & 75.2 \perf{-0.1} & 77.8 \perf{0.1} & 96.1 \perf{0.0} & 99.9 \perf{0.2} & 97.4 \perf{-0.1} & 98.7 \perf{0.0} & 99.7 \perf{0.0} & 79.4 \perf{0.0} & 90.5 \perf{0.0} \\
            & INT3  & 75.4 \perf{0.1} & 78.9 \perf{1.2} & 96.0 \perf{-0.1} & 99.9 \perf{0.2} & 97.3 \perf{-0.2} & 99.0 \perf{0.3} & 99.7 \perf{0.0} & 79.4 \perf{0.0} & 90.7 \perf{0.2} \\
            & INT2  & 73.8 \perf{-1.5} & 73.3 \perf{-4.4} & 91.2 \perf{-4.9} & 94.8 \perf{-4.9} & 81.4 \perf{-16.1} & 81.8 \perf{-16.9} & 98.5 \perf{-1.2} & 73.2 \perf{-6.2} & 83.5 \perf{-7.0} \\
            \midrule
            & FP32  & 55.2 & 54.9 & 66.7 & 78.9 & 80.2 & 69.7 & 97.3 & 50.4 & 69.2 \\
            & FQ8  & 53.7 \perf{-1.5} & 52.7 \perf{-2.2} & 66.2 \perf{-0.5} & 75.5 \perf{-3.4} & 79.7 \perf{-0.5} & 69.0 \perf{-0.7} & 97.2 \perf{-0.1} & 50.6 \perf{0.2} & 68.1 \perf{-1.1} \\
            & FQ4  & \phantom{0}0.3 \perf{-54.9} & \phantom{0}0.6 \perf{-54.3} & \phantom{0}1.7 \perf{-65.0} & \phantom{0}7.0 \perf{-71.9} & \phantom{0}9.1 \perf{-71.1} & \phantom{0}3.8 \perf{-65.9} & \phantom{0}8.9 \perf{-88.4} & \phantom{0}2.3 \perf{-48.1} & \phantom{0}4.2 \perf{-65.0} \\
            \multirow{2}{*}{\textbf{Task Arithmetic ~\cite{TaskArithmetic_2023_ICLR}}}& INT8  & 55.2 \perf{0.0} & 55.0 \perf{0.1} & 66.7 \perf{0.0} & 77.4 \perf{-1.5} & 80.2 \perf{0.0} & 69.7 \perf{0.0} & 97.3 \perf{0.0} & 50.1 \perf{-0.3} & 69.0 \perf{-0.2} \\
            & INT4  & 55.6 \perf{0.4} & 55.0 \perf{0.1} & 66.9 \perf{0.2} & 77.7 \perf{-1.2} & 80.2 \perf{0.0} & 69.6 \perf{-0.1} & 97.3 \perf{0.0} & 50.5 \perf{0.1} & 69.1 \perf{-0.1} \\
            & INT3  & 60.8 \perf{5.6} & 59.0 \perf{4.1} & 71.6 \perf{4.9} & 79.5 \perf{0.6} & 79.2 \perf{-1.0} & 69.1 \perf{-0.6} & 97.1 \perf{-0.2} & 53.5 \perf{3.1} & 71.2 \perf{2.0} \\
            & INT2  & 66.1 \perf{10.9} & 62.4 \perf{7.5} & 69.8 \perf{3.1} & 66.6 \perf{-12.3} & 54.6 \perf{-25.6} & 43.2 \perf{-26.5} & 83.0 \perf{-14.3} & 50.8 \perf{0.4} & 62.1 \perf{-7.1} \\
            & RTVQ  & 58.1 \perf{2.9} & 56.8 \perf{1.9} & 70.4 \perf{3.7} & 81.8 \perf{2.9} & 77.6 \perf{-2.6} & 67.6 \perf{-2.1} & 96.8 \perf{-0.5} & 52.9 \perf{2.5} & 70.2 \perf{1.0} \\
            \midrule
            & FP32  & 65.0  & 64.4  & 74.8  & 77.4  & 81.2  & 69.3  & 96.5  & 54.5  & 72.9 \\
            & FQ8  & 56.1 \perf{-8.9} & 52.6 \perf{-11.8} & 63.8 \perf{-11.0} & 67.9 \perf{-9.5} & 73.6 \perf{-7.6} & 62.3 \perf{-7.0} & 94.7 \perf{-1.8} & 47.1 \perf{-7.4} & 64.8 \perf{-8.1} \\
            & FQ4  & \phantom{0}0.2 \perf{-64.8} & \phantom{0}0.5 \perf{-63.9} & \phantom{0}1.9 \perf{-72.9} & \phantom{0}9.0 \perf{-68.4} & \phantom{0}7.5 \perf{-73.7} & \phantom{0}2.2 \perf{-67.1} & \phantom{0}9.6 \perf{-86.9} & \phantom{0}2.4 \perf{-52.1} & \phantom{0}4.2 \perf{-68.7} \\
            \multirow{2}{*}{\textbf{Ties-Merging ~\cite{TiesMerging_2023_NeurIPS}}}& INT8  & 65.0 \perf{0.0} & 64.2 \perf{-0.2} & 74.6 \perf{-0.2} & 76.7 \perf{-0.7} & 81.2 \perf{0.0} & 69.4 \perf{0.1} & 96.6 \perf{0.1} & 54.2 \perf{-0.3} & 72.7 \perf{-0.2} \\
            & INT4  & 63.8 \perf{-1.2} & 63.5 \perf{-0.9} & 74.6 \perf{-0.2} & 77.5 \perf{0.1} & 78.9 \perf{-2.3} & 69.1 \perf{-0.2} & 95.7 \perf{-0.8} & 52.9 \perf{-1.6} & 72.0 \perf{-0.9} \\
            & INT3  & 64.5 \perf{-0.5} & 64.1 \perf{-0.3} & 71.9 \perf{-2.9} & 74.0 \perf{-3.4} & 88.6 \perf{7.4} & 74.0 \perf{4.7} & 98.2 \perf{1.7} & 53.4 \perf{-1.1} & 73.6 \perf{0.7} \\
            & INT2  & 66.3 \perf{1.3} & 62.9 \perf{-1.5} & 69.7 \perf{-5.1} & 66.4 \perf{-11.0} & 56.4 \perf{-24.8} & 43.7 \perf{-25.6} & 84.8 \perf{-11.7} & 50.6 \perf{-3.9} & 62.6 \perf{-10.3} \\
            & RTVQ  & 61.9 \perf{-3.1} & 64.1 \perf{-0.3} & 77.3 \perf{2.5} & 75.9 \perf{-1.5} & 81.8 \perf{0.6} & 68.6 \perf{-0.7} & 96.4 \perf{-0.1} & 55.9 \perf{1.4} & 72.7 \perf{-0.2} \\
            \midrule
            & FP32  & 63.7 & 63.9 & 75.1 & 86.1 & 79.4 & 72.2 & 96.2 & 56.5 & 74.1 \\
            & FQ8  & 63.6 \perf{-0.1} & 62.7 \perf{-1.2} & 75.1 \perf{0.0} & 85.6 \perf{-0.5} & 79.4 \perf{0.0} & 72.6 \perf{0.4} & 96.0 \perf{-0.2} & 56.3 \perf{-0.2} & 73.9 \perf{-0.2} \\
            & FQ4  & \phantom{0}0.2 \perf{-63.5} & \phantom{0}0.3 \perf{-63.6} & \phantom{0}2.5 \perf{-72.6} & \phantom{0}8.9 \perf{-77.2} & \phantom{0}7.7 \perf{-71.7} & \phantom{0}5.9 \perf{-66.3} & \phantom{0}6.9 \perf{-89.3} & \phantom{0}1.7 \perf{-54.8} & \phantom{0}4.3 \perf{-69.8} \\
             \multirow{2}{*}{\textbf{LiNes ~\cite{LiNeS_2025_ICLR}}}& INT8  & 63.7 \perf{0.0} & 63.9 \perf{0.0} & 75.1 \perf{0.0} & 86.2 \perf{0.1} & 79.4 \perf{0.0} & 72.2 \perf{0.0} & 96.2 \perf{0.0} & 56.5 \perf{0.0} & 74.2 \perf{0.1} \\
            & INT4  & 63.9 \perf{0.2} & 64.2 \perf{0.3} & 75.5 \perf{0.4} & 86.1 \perf{0.0} & 79.3 \perf{-0.1} & 72.1 \perf{-0.1} & 96.1 \perf{-0.1} & 56.8 \perf{0.3} & 74.2 \perf{0.1} \\
            & INT3  & 66.6 \perf{2.9} & 65.9 \perf{2.0} & 77.7 \perf{2.6} & 85.7 \perf{-0.4} & 77.2 \perf{-2.2} & 70.6 \perf{-1.6} & 95.7 \perf{-0.5} & 58.0 \perf{1.5} & 74.7 \perf{0.6} \\
            & INT2  & 66.4 \perf{2.7} & 63.1 \perf{-0.8} & 71.0 \perf{-4.1} & 64.9 \perf{-21.2} & 47.5 \perf{-31.9} & 42.7 \perf{-29.5} & 78.8 \perf{-17.4} & 51.2 \perf{-5.3} & 60.7 \perf{-13.4} \\
            & RTVQ  & 65.4 \perf{1.7} & 64.3 \perf{0.4} & 77.6 \perf{2.5} & 86.7 \perf{0.6} & 75.8 \perf{-3.6} & 69.5 \perf{-2.7} & 95.3 \perf{-0.9} & 58.7 \perf{2.2} & 74.2 \perf{0.1} \\
            \midrule
            & FP32 & 64.8 & 63.1 & 72.2 & 82.6 & 84.4 & 77.2 & 97.2 & 57.8 & 74.9 \\
            & FQ8  & 61.0 \perf{-3.8} & 52.7 \perf{-10.4} & 68.9 \perf{-3.3} & 77.3 \perf{-5.3} & 82.5 \perf{-1.9} & 73.1 \perf{-4.1} & 93.6 \perf{-3.6} & 55.3 \perf{-2.5} & 70.6 \perf{-4.3} \\
            & FQ4  & \phantom{0}0.4 \perf{-64.4} & \phantom{0}0.5 \perf{-62.6} & \phantom{0}1.8 \perf{-70.4} & \phantom{0}6.8 \perf{-75.8} & \phantom{0}8.0 \perf{-76.4} & \phantom{0}1.9 \perf{-75.3} & \phantom{0}8.2 \perf{-89.0} & \phantom{0}2.1 \perf{-55.7} & \phantom{0}3.7 \perf{-71.2} \\
             \multirow{2}{*}{\textbf{Consensus TA ~\cite{Tall_Mask_2024_ICML}}}& INT8  & 64.8 \perf{0.0} & 63.1 \perf{0.0} & 72.2 \perf{0.0} & 82.5 \perf{-0.1} & 84.4 \perf{0.0} & 77.2 \perf{0.0} & 97.3 \perf{0.1} & 57.7 \perf{-0.1} & 74.9 \perf{0.0} \\
            & INT4  & 64.9 \perf{0.1} & 63.1 \perf{0.0} & 72.7 \perf{0.5} & 82.8 \perf{0.2} & 84.2 \perf{-0.2} & 76.7 \perf{-0.5} & 97.2 \perf{0.0} & 57.8 \perf{0.0} & 74.9 \perf{0.0} \\
            & INT3  & 66.5 \perf{1.7} & 64.6 \perf{1.5} & 75.0 \perf{2.8} & 81.8 \perf{-0.8} & 81.4 \perf{-3.0} & 72.5 \perf{-4.7} & 96.6 \perf{-0.6} & 59.8 \perf{2.0} & 74.8 \perf{-0.1} \\
            & INT2  & 66.5 \perf{1.7} & 61.2 \perf{-1.9} & 69.5 \perf{-2.7} & 61.9 \perf{-20.7} & 46.4 \perf{-38.0} & 37.8 \perf{-39.4} & 71.0 \perf{-26.2} & 53.4 \perf{-4.4} & 58.5 \perf{-16.4} \\
            & RTVQ  & 67.7 \perf{2.9} & 63.9 \perf{0.8} & 76.6 \perf{4.4} & 79.9 \perf{-2.7} & 73.1 \perf{-11.3} & 64.7 \perf{-12.5} & 94.3 \perf{-2.9} & 61.2 \perf{3.4} & 72.7 \perf{-2.2} \\
            \midrule
            & FP32 & 64.7 & 70.2 & 83.6 & 94.2 & 85.7 & 94.2 & 97.7 & 63.7 & 81.8 \\
            & FQ8  & 63.5 \perf{-1.2} & 69.8 \perf{-0.4} & 83.9 \perf{0.3} & 93.1 \perf{-1.1} & 87.2 \perf{1.5} & 94.4 \perf{0.2} & 97.9 \perf{0.2} & 63.1 \perf{-0.6} & 81.6 \perf{-0.2} \\
            & FQ4  & \phantom{0}0.3 \perf{-64.4} & \phantom{0}0.5 \perf{-69.7} & \phantom{0}2.7 \perf{-80.9} & \phantom{0}9.0 \perf{-85.2} & \phantom{0}8.8 \perf{-76.9} & \phantom{0}4.0 \perf{-90.2} & \phantom{0}8.9 \perf{-88.8} & \phantom{0}2.1 \perf{-61.6} & \phantom{0}4.5 \perf{-77.3} \\
            \multirow{2}{*}{\textbf{AdaMerging ~\cite{Adamerging_2024_ICLR}}}& INT8  & 64.4 \perf{-0.3} & 70.5 \perf{0.3} & 83.7 \perf{0.1} & 92.7 \perf{-1.5} & 86.8 \perf{1.1} & 94.3 \perf{0.1} & 97.5 \perf{-0.2} & 62.6 \perf{-1.1} & 81.6 \perf{-0.2} \\
            & INT4  & 64.1 \perf{-0.6} & 70.4 \perf{0.2} & 83.2 \perf{-0.4} & 93.3 \perf{-0.9} & 86.7 \perf{1.0} & 94.2 \perf{0.0} & 97.7 \perf{0.0} & 62.1 \perf{-1.6} & 81.5 \perf{-0.3} \\
            & INT3  & 64.3 \perf{-0.4} & 71.3 \perf{1.1} & 83.8 \perf{0.2} & 93.0 \perf{-1.2} & 87.9 \perf{2.2} & 94.3 \perf{0.1} & 97.8 \perf{0.1} & 63.5 \perf{-0.2} & 82.0 \perf{0.2} \\
            & INT2  & 63.4 \perf{-1.3} & 64.8 \perf{-5.4} & 84.0 \perf{0.4} & 90.7 \perf{-3.5} & 88.5 \perf{2.8} & 71.2 \perf{-23.0} & 98.5 \perf{0.8} & 63.7 \perf{0.0} & 78.1 \perf{-3.7} \\
            & RTVQ  & 66.3 \perf{1.6} & 71.5 \perf{1.3} & 84.6 \perf{1.0} & 92.7 \perf{-1.5} & 88.2 \perf{2.5} & 93.5 \perf{-0.7} & 98.1 \perf{0.4} & 67.8 \perf{4.1} & 82.8 \perf{1.0} \\
            \midrule
            & FP32 & 71.8 & 72.8 & 93.5 & 99.4 & 96.9 & 98.1 & 99.6 & 74.4 & 88.3 \\
            & FQ8  & 73.6 \perf{1.8} & 75.9 \perf{3.1} & 94.0 \perf{0.5} & 99.0 \perf{-0.4} & 96.6 \perf{-0.3} & 97.2 \perf{-0.9} & 99.6 \perf{0.0} & 73.4 \perf{-1.0} & 88.7 \perf{0.4} \\
            & FQ4  & \phantom{0}0.3 \perf{-71.5} & \phantom{0}0.6 \perf{-72.2} & \phantom{0}2.8 \perf{-90.7} & \phantom{0}5.8 \perf{-93.6} & \phantom{0}9.5 \perf{-87.4} & \phantom{0}1.5 \perf{-96.6} & \phantom{0}8.0 \perf{-91.6} & \phantom{0}2.7 \perf{-71.7} & \phantom{0}3.9 \perf{-84.4} \\
            \multirow{2}{*}{\textbf{EMR-Merging ~\cite{EMRMerging_2024}}}& INT8  & 72.0 \perf{0.2} & 72.9 \perf{0.1} & 93.7 \perf{0.2} & 99.4 \perf{0.0} & 96.9 \perf{0.0} & 98.2 \perf{0.1} & 99.6 \perf{0.0} & 74.5 \perf{0.1} & 88.4 \perf{0.1} \\
            & INT4  & 74.1 \perf{2.3} & 76.4 \perf{3.6} & 94.9 \perf{1.4} & 99.7 \perf{0.3} & 97.3 \perf{0.4} & 98.6 \perf{0.5} & 99.6 \perf{0.0} & 77.4 \perf{3.0} & 89.8 \perf{1.5} \\
            & INT3  & 74.9 \perf{3.1} & 78.0 \perf{5.2} & 95.2 \perf{1.7} & 99.6 \perf{0.2} & 97.2 \perf{0.3} & 98.4 \perf{0.3} & 99.7 \perf{0.1} & 77.1 \perf{2.7} & 90.0 \perf{1.7} \\
            & INT2  & 71.5 \perf{-0.3} & 70.5 \perf{-2.3} & 84.9 \perf{-8.6} & 89.0 \perf{-10.4} & 72.4 \perf{-24.5} & 68.0 \perf{-30.1} & 96.3 \perf{-3.3} & 64.9 \perf{-9.5} & 77.2 \perf{-11.1} \\
            & RTVQ  & 71.4 \perf{-0.4} & 71.4 \perf{-1.4} & 88.6 \perf{-4.9} & 94.2 \perf{-5.2} & 89.7 \perf{-7.2} & 84.3 \perf{-13.8} & 99.0 \perf{-0.6} & 66.6 \perf{-7.8} & 83.2 \perf{-5.1} \\
            
        \bottomrule
\end{tabular}
}
\caption{Comprehensive task-level results for merging 8 classification tasks using ViT-B/32. For RTVQ, we use a 3-bit base vector and a 2-bit offset vector (equivalent to 2.375 bits per task).}
\label{tab:supple_vitb32_tasklevel}

\vspace{-5pt}
\end{table*}

\definecolor{mygray}{gray}{.95}

\begin{table*}[tbh!]
\footnotesize 

\vspace{-5pt}
\centering
\resizebox{\textwidth}{!}{
\begin{tabular}{l|l | l l l l l l l l |l}
        \toprule
        \multicolumn{2}{c|}{\textbf{Method}} & \multicolumn{1}{c}{\textbf{SUN}} & \multicolumn{1}{c}{\textbf{Cars}} & \multicolumn{1}{c}{\textbf{RES.}} & \multicolumn{1}{c}{\textbf{Euro}} & \multicolumn{1}{c}{\textbf{SVH.}} & \multicolumn{1}{c}{\textbf{GTS.}} & \multicolumn{1}{c}{\textbf{MNI.}} & \multicolumn{1}{c|}{\textbf{DTD}} &\multicolumn{1}{c}{\textbf{Avg.}} \\
            \midrule
            & FP32   & 82.3  & 92.4 & 97.4 & 100  & 98.1 & 99.2 & 99.7 & 84.1 & 94.2 \\
            & FT8   & 82.3 \perf{0.0} & 92.2 \perf{-0.2} & 97.3 \perf{-0.1} & 99.9 \perf{-0.1} & 98.1 \perf{0.0} & 99.2 \perf{0.0} & 99.7 \perf{0.0} & 84.7 \perf{0.6} & 94.2 \perf{0.0} \\
            & FT4   & \phantom{0}0.2 \perf{-82.1} & \phantom{0}0.5 \perf{-91.9} & \phantom{0}2.7 \perf{-94.7} & \phantom{0}4.0 \perf{-96.0} & \phantom{0}8.0 \perf{-90.1} & \phantom{0}2.4 \perf{-96.8} & 14.8 \perf{-84.9} & \phantom{0}1.8 \perf{-82.3} & \phantom{0}4.3 \perf{-89.9} \\
            \multirow{2}{*}{\textbf{Individual}} & INT8  & 82.3 \perf{0.0} & 92.4 \perf{0.0} & 97.4 \perf{0.0} & 99.9 \perf{-0.1} & 98.1 \perf{0.0} & 99.2 \perf{0.0} & 99.7 \perf{0.0} & 84.1 \perf{0.0} & 94.1 \perf{-0.1} \\
            & INT4  & 82.3 \perf{0.0} & 92.4 \perf{0.0} & 97.4 \perf{0.0} & 99.9 \perf{-0.1} & 98.1 \perf{0.0} & 99.2 \perf{0.0} & 99.7 \perf{0.0} & 84.3 \perf{0.2} & 94.2 \perf{0.0} \\
            & INT3  & 82.3 \perf{0.0} & 92.5 \perf{0.1} & 97.4 \perf{0.0} & 99.9 \perf{-0.1} & 98.1 \perf{0.0} & 99.2 \perf{0.0} & 99.7 \perf{0.0} & 84.3 \perf{0.2} & 94.2 \perf{0.0} \\
            & INT2  & 80.5 \perf{-1.8} & 90.9 \perf{-1.5} & 96.3 \perf{-1.1} & 99.3 \perf{-0.7} & 87.5 \perf{-10.6} & 93.7 \perf{-5.5} & 99.3 \perf{-0.4} & 80.0 \perf{-4.1} & 90.9 \perf{-3.3} \\
            \midrule
            & FP32  & 74.1 & 82.1 & 86.7 & 92.6 & 87.9 & 86.8 & 98.9 & 65.6 & 84.3 \\
            & FQ8  & 73.8 \perf{-0.3} & 81.6 \perf{-0.5} & 86.3 \perf{-0.4} & 92.9 \perf{0.3} & 87.6 \perf{-0.3} & 86.1 \perf{-0.7} & 98.9 \perf{0.0} & 65.6 \perf{0.0} & 84.1 \perf{-0.2} \\
            & FQ4  & \phantom{0}0.2 \perf{-73.9} & \phantom{0}0.5 \perf{-81.6} & \phantom{0}1.5 \perf{-85.2} & \phantom{0}9.6 \perf{-83.0} & \phantom{0}9.7 \perf{-78.2} & \phantom{0}2.1 \perf{-84.7} & \phantom{0}9.8 \perf{-89.1} & \phantom{0}2.4 \perf{-63.2} & \phantom{0}4.5 \perf{-79.8} \\
            \multirow{2}{*}{\textbf{Task Arithmetic ~\cite{TaskArithmetic_2023_ICLR}}}& INT8  & 74.1 \perf{0.0} & 82.1 \perf{0.0} & 86.7 \perf{0.0} & 92.6 \perf{0.0} & 87.9 \perf{0.0} & 86.8 \perf{0.0} & 98.9 \perf{0.0} & 65.6 \perf{0.0} & 84.3 \perf{0.0} \\
            & INT4  & 74.1 \perf{0.0} & 82.2 \perf{0.1} & 86.7 \perf{0.0} & 92.7 \perf{0.1} & 87.8 \perf{-0.1} & 86.7 \perf{-0.1} & 98.9 \perf{0.0} & 65.9 \perf{0.3} & 84.4 \perf{0.1} \\
            & INT3  & 74.5 \perf{0.4} & 83.3 \perf{1.2} & 87.7 \perf{1.0} & 93.5 \perf{0.9} & 87.0 \perf{-0.9} & 86.9 \perf{0.1} & 98.9 \perf{0.0} & 66.9 \perf{1.3} & 84.8 \perf{0.5} \\
            & INT2  & 72.8 \perf{-1.3} & 82.8 \perf{0.7} & 84.0 \perf{-2.7} & 86.1 \perf{-6.5} & 72.9 \perf{-15.0} & 65.5 \perf{-21.3} & 94.2 \perf{-4.7} & 64.7 \perf{-0.9} & 77.9 \perf{-6.4} \\
            & RTVQ  & 75.0 \perf{0.9} & 83.0 \perf{0.9} & 87.8 \perf{1.1} & 93.0 \perf{0.4} & 86.0 \perf{-1.9} & 86.6 \perf{-0.2} & 98.8 \perf{-0.1} & 68.0 \perf{2.4} & 84.8 \perf{0.5} \\
            \midrule
            & FP32  & 75.0 & 84.5 & 88.0 & 94.3 & 85.7 & 82.1 & 98.7 & 67.7 & 84.5 \\
            & FQ8  & 71.2 \perf{-3.8} & 77.0 \perf{-7.5} & 79.5 \perf{-8.5} & 86.0 \perf{-8.3} & 80.8 \perf{-4.9} & 72.8 \perf{-9.3} & 97.9 \perf{-0.8} & 61.5 \perf{-6.2} & 78.3 \perf{-6.2} \\
            & FQ4  & \phantom{0}0.2 \perf{-74.8} & \phantom{0}0.5 \perf{-84.0} & \phantom{0}4.5 \perf{-83.5} & \phantom{0}5.1 \perf{-89.2} & \phantom{0}7.8 \perf{-77.9} & \phantom{0}3.0 \perf{-79.1} & 10.3 \perf{-88.4} & \phantom{0}1.8 \perf{-65.9} & \phantom{0}4.2 \perf{-80.3} \\
            \multirow{2}{*}{\textbf{Ties-Merging ~\cite{TiesMerging_2023_NeurIPS}}}& INT8  & 75.0 \perf{0.0} & 84.5 \perf{0.0} & 88.0 \perf{0.0} & 94.3 \perf{0.0} & 85.7 \perf{0.0} & 82.1 \perf{0.0} & 98.7 \perf{0.0} & 67.8 \perf{0.1} & 84.5 \perf{0.0} \\
            & INT4  & 75.0 \perf{0.0} & 84.4 \perf{-0.1} & 88.2 \perf{0.2} & 94.3 \perf{0.0} & 85.9 \perf{0.2} & 82.9 \perf{0.8} & 98.6 \perf{-0.1} & 67.8 \perf{0.1} & 84.6 \perf{0.1} \\
            & INT3  & 76.4 \perf{1.4} & 85.7 \perf{1.2} & 88.0 \perf{0.0} & 93.6 \perf{-0.7} & 89.0 \perf{3.3} & 82.6 \perf{0.5} & 99.1 \perf{0.4} & 67.7 \perf{0.0} & 85.3 \perf{0.8} \\
            & INT2  & 73.3 \perf{-1.7} & 83.1 \perf{-1.4} & 84.4 \perf{-3.6} & 85.9 \perf{-8.4} & 73.8 \perf{-11.9} & 64.2 \perf{-17.9} & 94.8 \perf{-3.9} & 64.5 \perf{-3.2} & 78.0 \perf{-6.5} \\
            & RTVQ  & 74.9 \perf{-0.1} & 83.6 \perf{-0.9} & 86.1 \perf{-1.9} & 91.5 \perf{-2.8} & 79.7 \perf{-6.0} & 72.6 \perf{-9.5} & 97.6 \perf{-1.1} & 67.1 \perf{-0.6} & 81.6 \perf{-2.9} \\
            \midrule
            & FP32  & 74.5 & 85.4 & 88.8 & 95.4 & 90.8 & 90.8 & 99.3 & 70.4 & 86.9 \\
            & FQ8  & 73.9 \perf{-0.6} & 84.2 \perf{-1.2} & 88.2 \perf{-0.6} & 95.4 \perf{0.0} & 90.3 \perf{-0.5} & 89.8 \perf{-1.0} & 99.3 \perf{0.0} & 69.7 \perf{-0.7} & 86.4 \perf{-0.5} \\
            & FQ4  & \phantom{0}0.3 \perf{-74.2} & \phantom{0}0.5 \perf{-84.9} & \phantom{0}3.6 \perf{-85.2} & 13.7 \perf{-81.7} & \phantom{0}8.8 \perf{-82.0} & \phantom{0}4.3 \perf{-86.5} & \phantom{0}9.8 \perf{-89.5} & \phantom{0}2.2 \perf{-68.2} & \phantom{0}5.4 \perf{-81.5} \\
            \multirow{2}{*}{\textbf{LiNes ~\cite{LiNeS_2025_ICLR}}}& INT8  & 74.5 \perf{0.0} & 85.4 \perf{0.0} & 88.8 \perf{0.0} & 95.4 \perf{0.0} & 90.8 \perf{0.0} & 90.8 \perf{0.0} & 99.3 \perf{0.0} & 70.4 \perf{0.0} & 86.9 \perf{0.0} \\
            & INT4  & 74.5 \perf{0.0} & 85.4 \perf{0.0} & 88.8 \perf{0.0} & 95.4 \perf{0.0} & 90.7 \perf{-0.1} & 90.8 \perf{0.0} & 99.3 \perf{0.0} & 70.6 \perf{0.2} & 86.9 \perf{0.0} \\
            & INT3  & 75.4 \perf{0.9} & 86.5 \perf{1.1} & 90.4 \perf{1.6} & 95.8 \perf{0.4} & 90.2 \perf{-0.6} & 91.6 \perf{0.8} & 99.3 \perf{0.0} & 72.4 \perf{2.0} & 87.7 \perf{0.8} \\
            & INT2  & 75.4 \perf{0.9} & 85.8 \perf{0.4} & 88.7 \perf{-0.1} & 91.4 \perf{-4.0} & 74.6 \perf{-16.2} & 72.7 \perf{-18.1} & 95.7 \perf{-3.6} & 69.7 \perf{-0.7} & 81.8 \perf{-5.1} \\
            & RTVQ  & 75.3 \perf{0.8} & 86.4 \perf{1.0} & 90.2 \perf{1.4} & 95.4 \perf{0.0} & 89.3 \perf{-1.5} & 91.3 \perf{0.5} & 99.3 \perf{0.0} & 74.1 \perf{3.7} & 87.7 \perf{0.8} \\
            \midrule
            & FP32  & 74.9 & 83.0 & 88.1 & 95.4 & 91.3 & 91.5 & 99.1 & 69.6 & 86.6 \\
            & FQ8  & 73.0 \perf{-1.9} & 78.7 \perf{-4.3} & 85.5 \perf{-2.6} & 94.7 \perf{-0.7} & 89.4 \perf{-1.9} & 89.2 \perf{-2.3} & 98.8 \perf{-0.3} & 67.5 \perf{-2.1} & 84.6 \perf{-2.0} \\
            & FQ4  & \phantom{0}0.1 \perf{-74.8} & \phantom{0}0.5 \perf{-82.5} & \phantom{0}3.2 \perf{-84.9} & \phantom{0}9.2 \perf{-86.2} & \phantom{0}6.7 \perf{-84.6} & \phantom{0}2.1 \perf{-89.4} & \phantom{0}9.8 \perf{-89.3} & \phantom{0}1.6 \perf{-68.0} & \phantom{0}4.2 \perf{-82.4} \\
            \multirow{2}{*}{\textbf{Consensus TA ~\cite{Tall_Mask_2024_ICML}}} & INT8  & 74.9 \perf{0.0} & 83.0 \perf{0.0} & 88.1 \perf{0.0} & 95.4 \perf{0.0} & 91.3 \perf{0.0} & 91.5 \perf{0.0} & 99.1 \perf{0.0} & 69.5 \perf{-0.1} & 86.6 \perf{0.0} \\
            & INT4  & 75.0 \perf{0.1} & 82.8 \perf{-0.2} & 88.2 \perf{0.1} & 95.5 \perf{0.1} & 91.2 \perf{-0.1} & 91.5 \perf{0.0} & 99.1 \perf{0.0} & 69.8 \perf{0.2} & 86.6 \perf{0.0} \\
            & INT3  & 75.4 \perf{0.5} & 83.9 \perf{0.9} & 89.5 \perf{1.4} & 95.7 \perf{0.3} & 89.9 \perf{-1.4} & 91.7 \perf{0.2} & 99.0 \perf{-0.1} & 71.7 \perf{2.1} & 87.1 \perf{0.5} \\
            & INT2  & 74.3 \perf{-0.6} & 83.8 \perf{0.8} & 86.9 \perf{-1.2} & 85.3 \perf{-10.1} & 70.7 \perf{-20.6} & 69.0 \perf{-22.5} & 92.0 \perf{-7.1} & 70.1 \perf{0.5} & 79.0 \perf{-7.6} \\
            & RTVQ  & 76.3 \perf{1.4} & 84.8 \perf{1.8} & 90.5 \perf{2.4} & 95.4 \perf{0.0} & 87.1 \perf{-4.2} & 90.9 \perf{-0.6} & 98.6 \perf{-0.5} & 75.0 \perf{5.4} & 87.3 \perf{0.7} \\
            \midrule
            & FP32  & 77.0 & 90.6 & 91.2 & 96.6 & 93.5 & 97.8 & 99.1 & 80.2 & 90.8 \\
            & FQ8  & 76.9 \perf{-0.1} & 90.5 \perf{-0.1} & 91.5 \perf{0.3} & 96.2 \perf{-0.4} & 93.6 \perf{0.1} & 97.9 \perf{0.1} & 99.0 \perf{-0.1} & 80.5 \perf{0.3} & 90.8 \perf{0.0} \\
            & FQ4  & \phantom{0}0.3 \perf{-76.7} & \phantom{0}0.5 \perf{-90.1} & \phantom{0}2.1 \perf{-89.1} & 11.7 \perf{-84.9} & \phantom{0}9.7 \perf{-83.8} & \phantom{0}2.1 \perf{-95.7} & \phantom{0}9.8 \perf{-89.3} & \phantom{0}2.1 \perf{-78.1} & \phantom{0}4.8 \perf{-86.0} \\
            \multirow{2}{*}{\textbf{AdaMerging} ~\cite{Adamerging_2024_ICLR}}& INT8  & 77.3 \perf{0.3} & 90.7 \perf{0.1} & 91.5 \perf{0.3} & 96.6 \perf{0.0} & 93.9 \perf{0.4} & 98.1 \perf{0.3} & 99.0 \perf{-0.1} & 80.2 \perf{0.0} & 90.9 \perf{0.1} \\
            & INT4  & 77.2 \perf{0.2} & 90.7 \perf{0.1} & 91.9 \perf{0.7} & 96.1 \perf{-0.5} & 93.8 \perf{0.3} & 98.0 \perf{0.2} & 99.1 \perf{0.0} & 80.3 \perf{0.1} & 90.9 \perf{0.1} \\
            & INT3  & 77.3 \perf{0.3} & 90.7 \perf{0.1} & 91.8 \perf{0.6} & 96.1 \perf{-0.5} & 93.7 \perf{0.2} & 98.3 \perf{0.5} & 99.1 \perf{0.0} & 80.7 \perf{0.5} & 91.0 \perf{0.2} \\
            & INT2  & 77.7 \perf{0.7} & 88.8 \perf{-1.8} & 92.3 \perf{1.1} & 96.4 \perf{-0.2} & 88.6 \perf{-4.9} & 94.6 \perf{-3.2} & 99.0 \perf{-0.1} & 78.1 \perf{-2.1} & 89.4 \perf{-1.4} \\
            & RTVQ  & 76.8 \perf{-0.2} & 90.7 \perf{0.1} & 91.7 \perf{0.5} & 96.6 \perf{0.0} & 92.9 \perf{-0.6} & 97.9 \perf{0.1} & 99.2 \perf{0.1} & 81.2 \perf{1.0} & 90.9 \perf{0.1} \\
            \midrule
            & FP32  & 81.1 & 90.7 & 96.8 & 99.7 & 97.9 & 99.1 & 99.7 & 82.7 & 93.5 \\
            & FQ8  & 80.3 \perf{-0.8} & 90.8 \perf{0.1} & 96.4 \perf{-0.4} & 99.7 \perf{0.0} & 97.3 \perf{-0.6} & 98.2 \perf{-0.9} & 99.6 \perf{-0.1} & 79.8 \perf{-2.9} & 92.8 \perf{-0.7} \\
            & FQ4  & \phantom{0}0.2 \perf{-80.9} & \phantom{0}0.6 \perf{-90.1} & \phantom{0}2.5 \perf{-94.3} & \phantom{0}5.1 \perf{-94.6} & \phantom{0}8.2 \perf{-89.7} & \phantom{0}2.7 \perf{-96.4} & 13.9 \perf{-85.8} & \phantom{0}2.3 \perf{-80.4} & \phantom{0}4.4 \perf{-89.1} \\
             \multirow{2}{*}{\textbf{EMR-Merging ~\cite{EMRMerging_2024}}}& INT8  & 81.2 \perf{0.1} & 90.8 \perf{0.1} & 96.8 \perf{0.0} & 99.7 \perf{0.0} & 98.0 \perf{0.1} & 99.1 \perf{0.0} & 99.7 \perf{0.0} & 82.8 \perf{0.1} & 93.5 \perf{0.0} \\
            & INT4  & 81.7 \perf{0.6} & 91.5 \perf{0.8} & 97.3 \perf{0.5} & 99.8 \perf{0.1} & 98.1 \perf{0.2} & 99.2 \perf{0.1} & 99.8 \perf{0.1} & 83.5 \perf{0.8} & 93.9 \perf{0.4} \\
            & INT3  & 82.2 \perf{1.1} & 91.9 \perf{1.2} & 97.3 \perf{0.5} & 99.8 \perf{0.1} & 98.1 \perf{0.2} & 99.2 \perf{0.1} & 99.7 \perf{0.0} & 83.0 \perf{0.3} & 93.9 \perf{0.4} \\
            & INT2  & 78.3 \perf{-2.8} & 89.1 \perf{-1.6} & 94.2 \perf{-2.6} & 97.8 \perf{-1.9} & 83.0 \perf{-14.9} & 84.7 \perf{-14.4} & 98.4 \perf{-1.3} & 75.4 \perf{-7.3} & 87.6 \perf{-5.9} \\
            & RTVQ  & 79.1 \perf{-2.0} & 89.0 \perf{-1.7} & 94.5 \perf{-2.3} & 97.9 \perf{-1.8} & 91.9 \perf{-6.0} & 93.8 \perf{-5.3} & 99.3 \perf{-0.4} & 76.8 \perf{-5.9} & 90.3 \perf{-3.2} \\
            
        \bottomrule
\end{tabular}
}
\caption{Comprehensive task-level results for merging 8 classification tasks using ViT-L/14. For RTVQ, we use a 3-bit base vector and a 2-bit offset vector (equivalent to 2.375 bits per task).}
\label{tab:supple_vitl14_tasklevel}

\vspace{-5pt}
\end{table*}

\clearpage

\end{document}